\documentclass[10pt,journal,compsoc]{IEEEtran}
\ifCLASSOPTIONcompsoc
  \usepackage[nocompress]{cite}
\else
  \usepackage{cite}
\fi
\usepackage{graphicx}
\usepackage[pagebackref,breaklinks,colorlinks]{hyperref}
\ifCLASSINFOpdf
\else
\fi
\usepackage{amsmath}
\usepackage{amssymb}
\usepackage{booktabs}
\usepackage{colortbl}
\usepackage[table]{xcolor}
\usepackage{color}

\usepackage{bm}
\usepackage{nicefrac} 

\usepackage{xspace}
\providecommand{\eg}{\textit{e.g.}\@\xspace}
\providecommand{\ie}{\textit{i.e.}\@\xspace}

\newlength\savewidth

\usepackage{enumitem}
\newcommand*\mean[1]{\bar{#1}}

\usepackage{algorithmic}

\usepackage{mathtools}
\DeclarePairedDelimiter{\norm}{\lVert}{\rVert}
\newcommand{\defeq}{\vcentcolon=}

\usepackage{orcidlink}

\begin{document}
%
\title{3DFaceShop: Explicitly Controllable 3D-Aware Portrait Generation
}
%
%
%
%
\author{Junshu Tang\orcidlink{0000-0002-6549-5257}, Bo Zhang\orcidlink{0000-0002-9795-4673}, Binxin Yang\orcidlink{0000-0003-4110-1986}, Ting Zhang\orcidlink{0000-0002-3952-2522}, Dong Chen\orcidlink{0000-0003-0588-9331}, Lizhuang Ma\orcidlink{0000-0003-0588-9331}, Fang Wen
\IEEEcompsocitemizethanks{\IEEEcompsocthanksitem The work is done when J. Tang and B. Yang are research interns at Microsoft Research Asia.
\IEEEcompsocthanksitem J. Tang and L. Ma are with the Department of Computer Science and Engineering, Shanghai Jiao Tong University, Shanghai, 200240, China. E-mail: tangjs@sjtu.edu.cn, ma-lz@cs.sjtu.edu.cn.\protect
\IEEEcompsocthanksitem B. Yang is with the Department of Electronic Engineering and Information Science, University of Science and Technology of China, Hefei, 230026, China, and also with Zhangjiang Laboratory, Shanghai, China. E-mail: tennyson@mail.ustc.edu.cn.
\IEEEcompsocthanksitem B. Zhang, T. Zhang, D. Chen, and F. Wen are with Microsoft Research Asia, Beijing, 100080, China. E-mail: \{zhanbo, tinzhan, doch, fangwen\}@microsoft.com. 
\IEEEcompsocthanksitem{B. Zhang and L. Ma are the corresponding authors.}
\IEEEcompsocthanksitem{Additional video results and code will be available on the \href{https://junshutang.github.io/control/index.html}{project webpage}. }
}
}

\IEEEtitleabstractindextext{%
\begin{abstract}

In contrast to the traditional avatar creation pipeline which is a costly process, contemporary generative approaches directly learn the data distribution from photographs. While plenty of works extend unconditional generative models and achieve some levels of controllability, it is still challenging to ensure multi-view consistency, especially in large poses. In this work, we propose a network that generates 3D-aware portraits while being controllable according to semantic parameters regarding pose, identity, expression and illumination. Our network uses neural scene representation to model 3D-aware portraits, whose generation is guided by a parametric face model that supports explicit control. While the latent disentanglement can be further enhanced by contrasting images with partially different attributes, there still exists noticeable inconsistency in non-face areas, \eg, hair and background, when animating expressions. We solve this by proposing a volume blending strategy in which we form a composite output by blending dynamic and static areas, with two parts segmented from the jointly learned semantic field. Our method outperforms prior arts in extensive experiments, producing realistic portraits with vivid expression in natural lighting when viewed from free viewpoints. It also demonstrates generalization ability to real images as well as out-of-domain data, showing great promise in real applications. 

\end{abstract}

}

\maketitle

\IEEEdisplaynontitleabstractindextext

%
\IEEEpeerreviewmaketitle

\section{Introduction}

\IEEEPARstart{C}an you imagine synthesizing a collection of photorealistic portraits that are animatable in 3D or waking up a photo as if the character is talking in front of us? Rendering controllable portraits is of significant importance to a variety of fields like film industry, video games, extended reality or immersive telecommunication. Traditional graphics pipeline~\cite{carranza2003free,casas20144d,li2017learning,sanyal2019learning,ichim2015dynamic} involves specialized 3D model creation with texture decoration which is then illuminated with realistic lighting and rendered 
using a physics-based renderer. While easily controllable, it is challenging to produce a myriad of avatars with photo-realistic quality.

In recent years we have seen a surge of neural rendering approaches~\cite{tewari2022advances,kato2020differentiable} that generate highly photo-realistic faces in a data-driven manner. In particular, generative adversarial networks (GANs)~\cite{liu2021generative,aggarwal2021generative,stylegan,stylegan2} hold the state of the arts, capable to synthesize high-resolution novel faces that are indistinguishable from the real ones. A prominent property of these off-the-shelf GANs is that they are often equipped with a semantic and disentangled latent space, hence one can animate the face or control specific facial attributes by traversing the latent space in the direction that correlates with the manipulated attribute~\cite{stylerig, pie, discofacegan, gan-control,patashnik2021styleclip,harkonen2020ganspace,shen2021closed,patashnik2021styleclip}. Moreover, since the whole rendering is fully differentiable, real photos can be inversely projected to the latent space~\cite{xia2022gan,tov2021designing,richardson2021encoding} and undergo the same semantic editing process. 

Due to the rise of augmented reality applications, there is an increasing demand for 3D face rendering such that faces can be rendered in different view angles while maintaining geometry consistency. While aforementioned 2D GANs~\cite{gan-control,pan20202d,siarohin2019first,harkonen2020ganspace} allow explicit head pose control to some extent, they fail to guarantee appearance consistency, leading to inconsistent identity or facial attributes when viewed from vastly different angles. To solve this, there are a few attempts~\cite{stylerig,discofacegan} that leverage parametric 3D face models~\cite{3dmm,egger20203d} to guide the generative process, thus faces can be explicitly controlled according to a set of semantic parameters relating to head pose, face shape, expression as well as illumination in the way like the computer animation workflow~\cite{parent2012computer}. Notwithstanding the improved disentanglement and controllability, the outputs yielded by 2D GANs are not truly 3D-aware in that one may still observe noticeable appearance and shape variations under distinct views of the same subject.

Meanwhile, the neural scene representations~\cite{tewari2022advances,nerf,deepsdf,mescheder2019occupancy,niemeyer2020differentiable,xie2022neural} emerge to become an expressive 3D-structure aware representation for general scenes, objects or persons. Among them, the seminar work, neural radiance field (NeRF)~\cite{nerf}, characterizes the complex scene as a continuous volume with radiance and  density at each location, which can be rendered to 2D observations via differentiable volumetric rendering~\cite{max1995optical}. Using a partial set of 2D images as supervision, NeRF-based methods can faithfully reconstruct the scene that can be rendered at  free viewpoint with photo-realistic quality. Thereafter, the research community turns to 3D-aware generative models~\cite{pigan,stylenerf,eg3d,stylenerf,giraffe,zhou2021cips,or2022stylesdf,zhao2022generative,rebain2022lolnerf} that learn to produce the 3D neural representation from 2D imagery. Now 
state-of-the-art approaches can generate compelling 3D portraits, yet the results are not as vivid as real persons since they are neither animatable nor controllable.   

In this work, we propose to generate 3D portraits that can be explicitly controlled by allowing the user edits upon a group of semantic parameters. To this end, we try to make the best of both the explicit parametric model and the neural face representation: the 3D Morphable Model (3DMM)~\cite{3dmm,egger20203d} provides the desired controllability in terms of face shape, expression and lighting; whereas the neural representation ensures multi-view consistency and offers photorealism. Specifically, the generation network is built upon the tri-plane representation proposed by~\cite{eg3d} which can be efficiently rendered, and the generation of such proxy representation is conditioned on the control space of 3DMM. The rendered images are expected to lie on the real face distribution as guaranteed by the adversarial loss~\cite{gan}, with the appearance resembling the rendered face mesh. 
However, the 3DMM guidance does not necessarily lead to disentangled control, since the latent sub-vector that accounts for a specific facial attribute may interfere with other attributes unexpectedly. Hence, we further improve the disentanglement by forming contrastive image pairs that differ in partial latent segments and enforcing the consistency for the attributes that share identical latent codes. As such, different latent sub-vectors bring independent effects to the final output and changing a segment of latent codes will not alter uncorrelated face properties. 

Nonetheless, it is still challenging to ensure ideal disentanglement, especially for non-face regions like hair, clothes and background, when controlling the facial expression. In fact, the expression animation is desired to mainly affect the facial area while bringing little influence on other parts. Given this, we propose a volume blending scheme which retains the radiance field of static area, \ie, the non-face region, in the final blended output. To accomplish this, we need to precisely locate the points belonging to the face area, hence we let the generation network simultaneously learn the semantic field using the online parsing results as supervision. In this way, we can determine the semantics of continuous space and accordingly segment the portrait in 3D. During inference, the output image is rendered from the blended radiance field with edits only applied to the localized face area. As demonstrated in the experiments, such a volume blending strategy effectively benefits the disentanglement, bringing much improved temporal consistency during the facial animation.

We compare our approach with multiple state-of-the-art methods, including controllable 2D GANs and talking head works, in which our approach performs favorably in terms of perceptual quality, control diversity, and disentanglement capability. Our method can generate lifelike 3D portraits in $512\times 512$ resolution and exhibits significant advantages for disentangled control in large poses which is particularly challenging for prior arts. We further show that our controllable portrait generation network can be personalized on real persons or even out-of-domain images, \eg, cartoons, and brings the static character to the 3D world with natural lighting and vivid expressions. Moreover, the learned semantic field offers accurate portrait matting and thus can enable background replacement with remarkable quality. We summarize our contributions as follows:
\begin{itemize}[leftmargin=*]
    \item We propose a controllable 3D portrait generation network that produces neural face representation conditioned on a set of semantic control parameters, with the disentangled control achieved by the guidance of parametric face models. Our approach allows easy and disentangled control for the pose, identity, expression, illumination as well as background appearance.
    
    \item To further improve the disentanglement, we propose a volume blending scheme, which blends the dynamic and static radiance fields of animated faces, with two parts separated by the jointly learned semantic field. 
    
    \item Our method outperforms the state of the arts on controllable 3D portrait generation in terms of both quantitative measure and qualitative evaluation. The network also performs well on real images and out-of-domain images like cartoon faces, showing great potential for various extended reality applications.   
\end{itemize}

\section{Related Work}
Photo-realistic face image synthesis and animation have been the longstanding focus of computer graphics as well as computer vision. Over the past years, rapid progress has been achieved with a large volume of efforts dedicated~\cite{gan,stylegan,stylegan2,graf,giraffe,eg3d,pie,stylerig,discofacegan,pirenderer,HeadNeRF}. Here we present a brief overview. For a more comprehensive review, please refer to the recent surveys~\cite{nickabadi2022comprehensive,zollhofer2018state,egger20203d,zhang2020deep,lu2017recent}.

\subsection{Controllable Face Image Editing}
Controlling and editing the appearance of the images, especially for face images, is an important feature demanded in many real-world applications.
Early works~\cite{lu2018attribute, di2017face,wang2018attribute} heavily depend on additional manual annotations for training a specific face animation model for the specific attribute.
Later, a great many efforts~\cite{pumarola2018ganimation,tang2020fine,pie,stylerig,discofacegan,gan-control,tang2021eggan,wang2021HFGI,zhang2022styleswin} aim at learning a disentangled and meaningful latent space for face images.
It is desired that different dimensions in the latent space characterize different facial attributes so that editing certain latent dimensions enables some kind of facial animation.
However, it is not guaranteed that factors of interested attributes are disentangled and often facial attributes are mingled in the latent space~\cite{locatello2019challenging,dai2019diagnosing}, which undermines the quality of face editing.

In order to enable more precise control over the generated images, 
many works~\cite{gan-control, pie,discofacegan,stylerig,piao2021inverting, geng20193d, nguyen2019hologan, piao2019semi, xu2020deep,pirenderer} have been proposed to incorporate 3D priors from parametric face models such as
3D Morphable Models (3DMMs), into GAN-based generative models. For example, 
DiscoFaceGAN~\cite{discofacegan}
proposes an imitative-contrastive paradigm that enforces the generative network to mimic
the rendering process of 3DMM. After training, it can enable precise control of the desired face properties such as pose, expression, illumination and so forth.
StyleRig~\cite{stylerig} describes a similar method to control StyleGAN via a 3DMM.
GAN-control~\cite{gan-control} also enhances GANs with an explicitly disentangled latent space and can edit the image by setting exact attributes such as age, pose, expression, etc.

These methods are still based on 2D image generators, and they are thus subject to severe 3D inconsistency issues due to the non-physical image
rendering process of 2D GANs, especially under large expression and pose variations.
Recent methods~\cite{sun2022fenerf, sun2022ide, gafni2021dynamic,athar2022rignerf,HeadNeRF,sun2022controllable} utilize neural representation to produce view-consistent and editable portrait images. FENerf~\cite{sun2022fenerf} and IDE3D~\cite{sun2022ide} utilize rendered semantic mask to edit 3D volume via GAN inversion, but they cannot produce continues animation results.
Other methods~\cite{gafni2021dynamic,athar2022rignerf,HeadNeRF,sun2022controllable}
incorporate 3DMM knowledge into volumetric rendering using neural scene representations to achieve consistent face editing.
For example,
HeadNeRF~\cite{HeadNeRF} integrates the neural radiance field to the parametric representation of the human head and is trained using annotated multi-view datasets.
Additionally, the concurrent works
cGOF~\cite{sun2022controllable} and GNARF~\cite{bergman2022generative} also
leverage the parametric face model and propose a conditional generative occupancy field for face expression animation. In contrast,
we focus on generative modeling that creates high-resolution and photo-realistic portraits including vivid expressions and illuminations.

\subsection{3D Morphable Models}
3D morphable models were first proposed in~\cite{3dmm} as a general and statistical representation model for face shape and appearance, which are typically learned from 3D scans of multiple people~\cite{blanz2003reanimating, booth20173d,FLAME, li2017learning,booth2018large}. In this way, faces are parameterized to a low-dimensional face space consisting of identity, expression, and illumination, which can be used to reconstruct 3D face mesh and is widely used for 3D face representation~\cite{cao2013facewarehouse,paysan20093d,tran2018nonlinear, tran2019towards}. Such a parameterized space also allows explicit control of the face synthesis with semantically interpretable parameters~\cite{egger20203d,cao2016real}. Rather than using time-consuming optimization approaches, recent works resort to deep neural networks for the 3D face model fitting~\cite{garrido2016reconstruction,genova2018unsupervised, richardson2017learning, sanyal2019learning, tewari2019fml}. Since the faces rendered by 3DMM often lack delicate details~\cite{kim2018deep,thies2019deferred}, recent efforts~\cite{stylerig,discofacegan,pie} leverage deep generative models for more photo-realistic synthesis. Along this line, this work proposes an explicitly controllable 3D-aware generative model by  combining the best of both worlds. Our method bridges the emerging 3D scene neural representation with the controllability of 3DMM, and  achieves controllable 3D portrait generation with much improved multi-view consistency, compared with prior works. 
\subsection{Neural Scene Representations}
In order to generate high-quality multi-view consistent images, neural scene representation using differentiable rendering~\cite{szabo2019unsupervised, shi2021lifting, wu2016learning, zhu2018visual, gadelha20173d,  nguyen2019hologan, nguyen2020blockgan,xie2021style} that can be optimized on a training set of only 2D multi-view images has gained popularity in the past few years.
Implicit representations~\cite{deepsdf, mescheder2019occupancy, nerf,sitzmann2020implicit} in particular neural radiance field (NeRF)~\cite{stylenerf,pigan, graf, giraffe,gram,gramhd} have been widely used in many areas such as 3D modeling~\cite{wang2021neus, yariv2021volume} and face/body digitization~\cite{gafni2021dynamic, guo2021ad, peng2021neural, raj2021pixel, su2021nerf, wang2021learning,ouyang2022real}.
They characterize the 3D scene with a continuous function via a light MLP, which is memory-efficient.
On the other hand,
explicit representations~\cite{lombardi2019neural, sitzmann2019deepvoxels} such as discrete voxels allow fast evaluation but usually suffer from huge memory consumption. 
Based on the complementary benefits of fully explicit and implicit representations, several methods~\cite{devries2021unconstrained, martel2021acorn, eg3d } have explored the hybrid explicit-implicit models.
Among them, EG3D~\cite{eg3d} shows high 3D GAN image quality by designing an efficient tri-plane hybrid 3D representation for unconditional 3D face synthesis.
In this work, we leverage this hybrid representation and further design a controllable framework to support precise 3D control over the generated faces such as facial expression, head pose, lighting, etc.

\section{Preliminaries: 3D-aware GANs}
\begin{figure*}
    \centering
    \includegraphics[width=\linewidth]{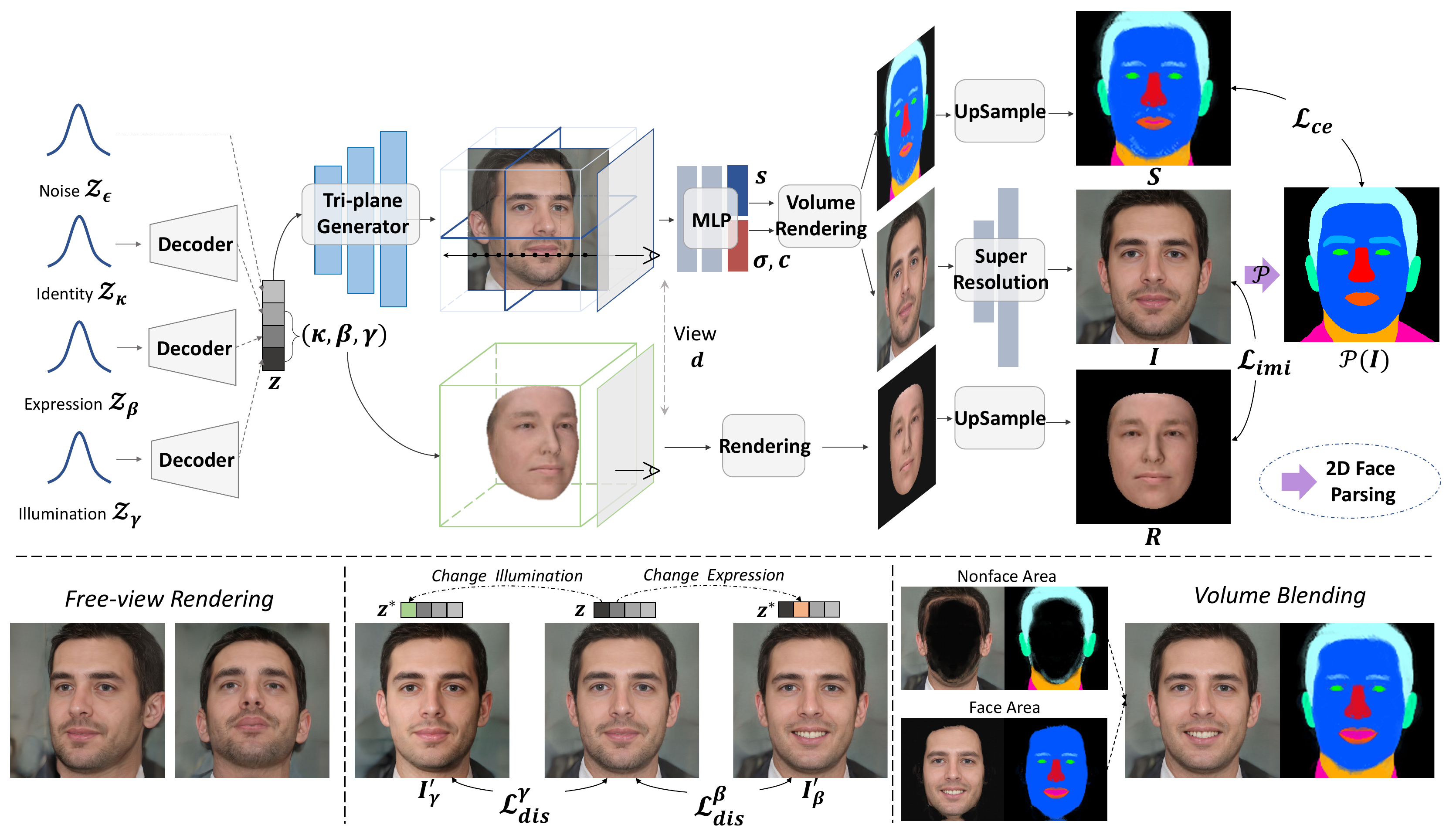}
    \caption{\textbf{Architecture overview.} We first sample the 3DMM control parameters $(\bm{\kappa},\bm{\beta},\bm{\gamma})$ by sampling from three separate decoders of pretrained VAE-GANs (Sec.~\ref{sec:vae}). We train a generative adversarial network conditioned on these control parameters, which generates the tri-plane representation that can be further rendered into 3D-aware portrait images via volumetric rendering (Sec.~\ref{sec:gan}).  Meanwhile, we render the 3DMM face mesh using the same parameters and use it to guide the generation, thus enabling the generator semantic controllability (Sec.~\ref{sec:imitative}). Once such imitative learning converges, we further improve the control disentanglement by enforcing the consistency of contrastive image pairs (Sec.~\ref{sec:dis}). During the radiance field generation, we simultaneously learn the portrait semantics in 3D space. With the accurately learned face parsing, we derive the final output which is a composition of the animated face region with the static non-face region(Sec.~\ref{sec:blend}). }
    \label{fig:framework}
\end{figure*}

While contemporary generative adversarial networks (GANs) are capable to generate photo-realistic faces, the synthesized images are not multi-view consistent even though plenty of advanced disentangled control methods have been proposed.
To ensure geometry consistency  from different views, 3D-aware GANs learn to model the distribution of underlying 3D face geometry from a collection of images. The crux of these approaches is to choose an intermediate 3D representation, which can be further rendered into 2D images from different viewpoints in a differentiable manner. One of the most expressive 3D representations is the neural radiance field (NeRF)~\cite{nerf}, which is a continuous volumetric representation. NeRF models the density $\sigma \in \mathbb{R}^{+}$ and view-dependent color $\bm{c} \in \mathbb{R}^3$ for each position $\bm{p}\in \mathbb{R}^3$ given the viewing direction $\bm{d} \in \mathbb{S}^2$. Hence, on the high level, the 3D-aware generator $\mathcal{G}$ parameterized by $\theta$ maps the latent code $\bm{z}$ in the latent space $\mathcal{Z}$ to the manifold of neural radiance fields:
\begin{equation}
    \bm{c}(\bm{p},\bm{d}), \sigma(\bm{p}) = \mathcal{G}_\theta(\bm{z}, \xi(\bm{p}), \xi(\bm{d})).
\end{equation}
In practice, to better model the high-frequency details, the inputs $\bm{p}$ and $\bm{d}$ are represented with sinusoidal positional encoding $\xi(x)=\left(x,...,\sin(2^k\pi x),\cos(2^k\pi x),...\right)$~\cite{pigan}. 

The scene represented by the above radiance field can be rendered into 2D images via volumetric rendering. Each rendered pixel in the image corresponds to a camera ray $\bm{r}(t) = \bm{o} + t\bm{d}$ shooting from the camera origin $\bm{o}$ and advancing in the direction $\bm{d}$. The ray traverses the radiance field, and the accumulated color along the ray is calculated by the following volume rendering integral~\cite{nerf},
\begin{equation}
    \mathcal{C}(\bm{r},\bm{z})= \int_{0}^{\infty} T(t)\sigma\big(\bm{r}(t)\big)\bm{c}\big(\bm{r}(t),\bm{d}\big)dt,
    \label{eq:render}
\end{equation}
where $T(t)$ denotes the accumulated transmittance along the ray:
\begin{equation}
    T(t) = \exp\left(-\int_{0}^{t} \sigma(\bm{r}(s))ds\right).
\end{equation}
In~\cite{nerf}, a hierarchical sampling scheme is proposed to discretize this integral process by sampling multiple points along the ray.  Parameterized with implicit multi-layer perceptrons (MLPs), NeRFs are quite expressive and can model the scene with photo-realistic quality.

\section{Proposed Approach}
In this work, we propose a 3D-aware GAN that allows explicit semantic control with respect to pose, identity, expression and illumination. To this end, we leverage a 3D-aware generator to ensure view consistency, and the generation is guided by a 3D face prior that admits semantic and interpretable control. As illustrated in Figure~\ref{fig:framework}, the 3D-aware GAN samples from the latent space are formed by a set of control parameters, and the generated images are enforced to imitate the rendered face from the parametric face model. To ensure disentangled control, we compare generated image pairs in a contrastive manner --- parameters associated with a certain attribute should not alter other attributes during generation. Nonetheless, inconsistency still occurs during semantic control, causing disturbing flickers in non-facial areas like hair, clothes and background. We remedy this issue by simultaneously predicting the semantic field and explicitly blending the radiance field accordingly so that only the facial area is manipulated with the non-face regions intact during the face control.

\subsection{Semantic Control Space}
\label{sec:vae}

We adopt the 3D Morphable Model (3DMM)~\cite{3dmm} to parameterize the face attributes and use it to  guide the 3D-aware generation. In 3DMM, faces can be modeled with a set of semantic parameters $(\bm{\alpha}, \bm{\beta}, \bm{\delta}, \bm{\gamma}, \bm{R}, \bm{t}) \in \mathbb{R}^{257}$. Specifically, $\bm{\alpha}\in \mathbb{R}^{80}$ describes the geometry of facial shape, $\bm{\beta}\in \mathbb{R}^{64}$ models the expression, $\bm{\delta}\in \mathbb{R}^{80}$ defines the albedo, $\bm{\gamma}$ characterizes the scene illumination, whereas $\bm{R}\in \text{SO}(3)$ and $\bm{t}\in \mathbb{R}^3$ denote the head rotation and translation respectively. 

The 3DMM yields a triangular mesh of 53$k$ vertices. The face shape $\bm{S}$ and the appearance $\bm{A}$ can be modeled as an affine model:
\begin{align}
    \bm{S} &= \mean{\bm{S}} + \bm{\tilde{S}}_{id}\bm{\alpha} + \bm{\tilde{S}}_{exp}\bm{\beta}, \\
    \bm{A} &= \mean{\bm{A}} + \bm{\tilde{A}}\bm{\delta},
\end{align}
where $\mean{\bm{S}}$ and $\mean{\bm{A}}$ are the average shape and appearance, whereas $\bm{\tilde{S}}_{id}$, $\bm{\tilde{S}}_{exp}$ and $\bm{\tilde{A}}$ are two-dimensional bases that account for the variation space of face shape $\bm{\alpha}$, expression $\bm{\beta}$ and albedo $\bm{\delta}$ respectively. We choose $\bm{\tilde{S}}_{id}$ and $\bm{\tilde{A}}$ from the Basel Face Model (BFM)~\cite{paysan20093d} which is computed from 200 face scans, and the expression bases $\bm{\tilde{S}}_{exp}$ are built from FaceWarehouse~\cite{cao2013facewarehouse}. Jointly considering the face shape and albedo, we let $\bm{\kappa} = [\bm{\alpha},\bm{\delta}]\in \mathbb{R}^{160}$ and use it to describe all the identity-related attributes. As for the lighting, we assume the faces to be Lambertian and approximate the scene illumination with Spherical Harmonics (SH). Specifically, for the vertex $\bm{p}_i$ with the surface normal $\bm{n}_i$ and skin color $\bm{A}_i$, its final color is computed as $\bm{A}_i \sum_{b=1}^{9} \bm{\gamma}' \bm{H}'(\bm{n}_i)$, where $\bm{H}: \mathbb{R}^3 \to \mathbb{R}$ is the SH basis function and $\bm{\gamma}' \in \mathbb{R}^3$ is the corresponding SH coefficient. Given the camera pose $[\bm{R}, \bm{t}]$, the face mesh can be rendered through Nvdiffrast~\cite{laine2020modular}. 

In order to generate controllable 3D portraits by manipulating the attribute parameters $(\bm{\kappa}, \bm{\beta}, \bm{\gamma})$, we need to learn a disentangled $\mathcal{W}$ space which is mapped from the attribute space $\mathcal{Z}$, where the $\mathcal{Z}$ space is easy to sample from. 
To this end, we first learn the latent space that accounts for the variations of each attribute. Specifically, we train separate variational auto-encoders (VAE) for identity $\bm{\kappa}$, expression $\bm{\beta}$ and illumination $\bm{\gamma}$ respectively: the encoder $E_i$ of $i$th VAE maps the real 3DMM coefficients to latent code which can be used to faithfully reconstruct the original coefficients via decoder $D_i$. We assume Gaussian prior for the latent space, so the VAE is optimized as,
\begin{align}
    \mathcal{L}_{\textnormal{VAE}_i} = &\mathbb{E}_{z\sim E_i(\bm{z}|\bm{x})}\norm{\bm{x} - D_i(\bm{x}|\bm{z})}_2  \\ &\quad+ \mu \mathrm{KL}\big(E_i(\bm{z}|\bm{x}) \| \mathcal{N}(\bm{0},\bm{I}) \big)  \\ &\quad+ \mathcal{L}_{\textnormal{VAE},\textnormal{GAN}}(\bm{x}),
\end{align}
where the first term $\ell_2$ reconstructs the input from the latent code and the second term penalizes the Kullback-Leibler (KL) divergence between the latent distribution and the normal distribution. Training VAE with $\ell_2$ reconstruction objective essentially assumes Gaussian distribution for $p(\bm{x}|\bm{z})$ which is prone to produce averaged reconstruction, or mean face. Hence, we additionally introduce an adversarial loss~\cite{vaegan}, $\mathcal{L}_{\textnormal{VAE},\textnormal{GAN}}$, to better distinguish the reconstructed attribute coefficients from real sampled ones. Experimental results 
show that this allows us to sample more diverse attributes. Besides FFHQ dataset~\cite{stylegan}, the VAE training also includes a talking face dataset~\cite{ravdess} containing exaggerated expressions, and thus we obtain a latent space with more diverse expressions. 

Once the VAEs are trained, we can derive plausible 3DMM coefficients $(\bm{\kappa}, \bm{\beta}, \bm{\gamma})$ by sampling from the VAE latent spaces.
These coefficients, along with a random Gaussian noise $\bm{\epsilon}\in \mathcal{Z}_{\epsilon}$ accounting for the remaining variations (\eg, background), define the full control space of faces. 

\subsection{3DMM Conditioned 3D-Aware GAN}
\label{sec:gan}
We expect the 3D-aware GAN that conditions on $\bm{z} \defeq (\bm{\kappa}, \bm{\beta}, \bm{\gamma},\bm{\epsilon})$ can generate faces that accurately resemble the identity, expression, and illumination of the 3DMM renderings. Hence, as illustrated in Figure~\ref{fig:framework}, we train a 3D-aware GAN that starts from the sampling space $\mathcal{Z}=(\mathcal{Z}_{\kappa},\mathcal{Z}_{\beta},\mathcal{Z}_{\gamma},\mathcal{Z}_{\epsilon})$ and generates 3D faces that demonstrate the desired properties. Formally, the radiance field is generated by,
\begin{equation}
    \bm{c}(\bm{p},\bm{d}), \sigma(\bm{p}) = \mathcal{G}_\theta(\bm{z}, \xi(\bm{p}), \xi(\bm{d})).
\end{equation}

In this work, instead of using MLPs to directly regress the continuous radiance field, we adopt the tri-plane representation recently proposed by~\cite{eg3d} which factorizes the 3D space using three orthogonal feature planes, denoted as $\bm{F}_{xy},\bm{F}_{xz},\bm{F}_{yz}\in \mathcal{R}^{H\times W \times C}$. Since much of the scene information is memorized explicitly while the decoder computed in the ray tracing is lightweight, it is computationally efficient to render the tri-plane representation while not compromising the expressivity.
Specifically, for the 3D point $\bm{p}$, we can project it onto the feature planes and obtain the aggregated feature by summing the retrieved feature from each feature plane. Such position-wise features are thereafter decoded to color and density as required by the volumetric rendering with a shallow MLP decoder, \ie,
\begin{equation}
    \bm{c}(\bm{p},\bm{d}),\sigma(\bm{p}) = \mathcal{G}_{\theta}^{\textnormal{MLP}}\big(\bm{F}_{xy}(\bm{p}_{xy})+\bm{F}_{xz}(\bm{p}_{xz})+\bm{F}_{yz}(\bm{p}_{yz})\big).
\end{equation}

Another significant advantage of using tri-plane representation is that one can directly take advantage of a powerful 2D CNN-based generator, \eg, StyleGAN\cite{stylegan, stylegan2}, to generate each feature plane. Hence the 3D-aware generator could enjoy many effective training strategies built for 2D GANs. Therefore, we map the control parameters to $\bm{w}\in \mathbb{R}^{18\times 512}$ code in the $\mathcal{W}$ space which is known to be disentangled, and the non-linear mapping $f_\theta: Z\to \mathcal{W}$ is implemented with eight fully connected layers. The different layers of the generator are modulated by the $\bm{w}$ code and output the tri-plane feature maps,
\begin{equation}
    \{\bm{F}_{xy}, \bm{F}_{xz}, \bm{F}_{yz}\} = \mathcal{G}_{\theta}^{\textnormal{CNN}}(\bm{w}, \bm{d}).
\end{equation}
Here, the tri-plane features are pose-dependent in that some facial attributes, \eg, expression, often correlate with the head pose in the face image capturing. 


\subsubsection{Generative Tri-plane Training Details}
While such an  explicit-implicit representation greatly reduces the memory footprint, it is still challenging to scale the method for high-resolution generation. One workaround is to rely on an image super-resolution network to enhance the fine details, yet this will sacrifice the view consistency and cause annoying flickers when changing the viewpoint. To alleviate this, following~\cite{eg3d} we introduce a dual discriminator $\mathcal{D}_{\theta}$ which does not only discriminate  the realism of the generation outputs but also examines the consistency between the 3D-aware low-resolution outputs and the resolution-enhanced counterpart. In this work, our network first generates coarse 3D-aware portraits at $128\times 128$ resolution and ultimately yields $512\times 512$ outputs by the super-resolution module.

At training time, the real images are sampled from the data distribution $p_{\textnormal{data}}$. To render generated images, we randomly sample the camera position from a unit sphere with the camera pointing to the sphere center~\cite{pigan}. The sampled yaw and pitch distributions $p_{\textnormal{cam}}$ are pre-computed from the training dataset. The tri-plane generation network is trained until an adversarially trained discriminator cannot distinguish the rendered generated images from the real ones. Both the generator and the discriminator are trained using the hinge loss~\cite{stylegan2}:
\begin{equation}
    \begin{aligned}
    \mathcal{L}_{\textnormal{GAN}}^{\mathcal{D}}=&-\mathbb{E}_{\bm{x}\sim p_{\textnormal{data}}}\big[\min(0, -1+\mathcal{D}_{\theta}(\bm{x}))\big]\\&-\mathbb{E}_{\bm{z}\sim p_{\bm{z}}, \bm{d}\sim p_{\textnormal{cam}}}\big[\min\big(0, -1-\mathcal{D}_{\theta}(\mathcal{C}(\mathcal{G}_{\theta}(\bm{z}, \bm{d})))\big)\big] \\ &+ \lambda_{R1} \mathcal{L}_{\textnormal{GP}},
    \\
    \mathcal{L}_{\textnormal{GAN}}^{\mathcal{G}}=&-\mathbb{E}_{\bm{z}\sim p_{\bm{z}}, \bm{d}\sim p_{\textnormal{cam}}}\big[\mathcal{D}_{\theta}(\mathcal{C}(\mathcal{G}_{\theta}(\bm{z}, \bm{d})))\big],
    \end{aligned}
\end{equation}
where, $\mathcal{C}$ denotes the volumetric renderer given by Equation~\ref{eq:render} and $\mathcal{L}_{\textnormal{GP}} = \norm{\nabla_{\theta}[\mathcal{D}_{\theta}(\bm{x})]}^{2}$ is the $R_1$ gradient penalty loss~\cite{mescheder2018training} that helps stabilize the adversarial training.

\subsubsection{Simultaneously Learned Semantic Field}
\label{sec:sf}
On top of obtaining the image renderings, we also train a semantic radiance field so that we can parse the portrait in 3D and better enforce the disentanglement. Specifically, we propose a multi-head decoder which consists of an apparent head and a semantic head:  the appearance head outputs the RGB feature and volume density, whereas the semantic head translates the feature to the semantic prediction at each point.  Both heads are two-layer MLPs with the first layer shared. Let $\bm{s}(\bm{p})$ denote the probability of $K$ semantic classes for point $\bm{p}$, \ie, $\bm{s}(\bm{p})=\mathcal{G}_{\theta}^{\textnormal{MLP}_s}(F(\bm{p})))$, the semantic field is rendered in a similar form as Equation~\ref{eq:render} except that the semantic field $\bm{s}$ does not depend on the viewing direction, which is,
\begin{equation}
    \centering
    \mathcal{S}(\bm{r})= \int_{0}^{\infty} T(t)\sigma\big(\bm{r}(t)\big)\bm{s}\big(\bm{r}(t)\big)dt,
\end{equation}
where $\mathcal{S}(\bm{r})\in\mathbb{R}^{HW\times K}$ is the rendered semantic mask. Likewise, we train this semantic field using image-level supervision. Specifically, we leverage a pretrained face parsing model $\mathcal{P}$~\cite{bisenet} to online extract the semantic mask for the generated image $\bm{I}$ and use this result as the 2D supervision. We minimize the categorical cross-entropy between the rendered semantic mask and the ground truth:
\begin{equation}
    \begin{aligned}
    \mathcal{L}_{\textnormal{ce}} &= -\frac{1}{H\times W}\sum_{i=1}^{H\times W}\sum_{k=1}^K\mathcal{P}(\bm{I})_{i,k}\log(\bm{S}_{i,k}),
    \end{aligned}
\end{equation}
where $i$ indexes the $H\times W$ image and $k$ is the semantic label index. In the following, we will show how the accurately learned semantic field facilitates disentangled control.

\subsection{Learning Controllable and Disentangled Radiance Field}
Given the camera pose, our method can generate the image $\bm{I}$ along with the semantic mask $\bm{S}$ conditioned on the control parameters $(\bm{\kappa}, \bm{\beta}, \bm{\gamma},\bm{\epsilon})$.
We expect the generated faces can accurately resemble the identity, expression, and illumination of the 3DMM face $\bm{R}$ rendered from the same parameters. We thus enable explicit controllability by feeding different semantic parameters, which is achieved in the first imitative learning stage.  Moreover, the control is desired to be disentangled --- editing one specific attribute should not cause changes in other attributes. We enforce such independent attribute editing in the disentanglement learning stage. We subsequently describe the two training stages.

\subsubsection{Imitative Learning}
\label{sec:imitative}
In this stage, we encourage the similarity of $\bm{I}$ and $\bm{R}$. Since the 3DMM model only renders the face area without hair and background, the training is actually guided using the blended face rendering $
\bm{R}' = \bm{R}+\bm{I}\odot (\bm{1}-\bm{B})$, where $\bm{B}$ is the face mask. The texture difference is penalized with $\ell_2$ loss, \ie, $\mathcal{L}_{\textnormal{tex}} = \norm{\bm{I} - \bm{R}'}_{2}$, and the identity similarity is encouraged by enforcing the identity loss , \ie, $\mathcal{L}_\textnormal{id} = 1-<\mathcal{F}(\bm{I}), \mathcal{F}(\bm{R}')>$, where $\mathcal{F}$ is the embedding extracted from a pretrained face recognition model~\cite{serengil2021lightface} and $<\cdot>$ denotes feature cosine similarity. 


To further encourage the expression similarity, we introduce the 3D landmark loss. We denote the 68 3D landmarks of the guided 3DMM image as $\bm{l}^{R}\in \mathbb{R}^{68\times 3}$. For the generated portrait $\bm{I}$, we adopt the differentiable face reconstruction method~\cite{deep3dface} and reconstruct a new face mesh with reconstructed landmarks $\bm{l}^{I} \in \mathbb{R}^{68\times 3}$. To better capture the subtle expressions, we use the re-weighted landmark loss: 
\begin{equation}
    \mathcal{L}_{\textnormal{lmk}} = \sum_i^{68} w_i \norm{\bm{l}_i^I - \bm{l}_i^R}_2,
\end{equation}
where $w_i$ denotes the weight for different landmarks. By default, we set $w=1$ but apply $w=100$ for landmarks relating to eyebrows and mouth.

Similarly, to accurately constrain the illumination, we enforce the similarity between the reconstructed illumination coefficients $\bm{\dot{\gamma}}$ and the input illumination condition $\bm{\gamma}$:
\begin{equation}
    \mathcal{L}_{\textnormal{ill}} = \norm{\bm{\gamma} - \bm{\dot{\gamma}}}_2.
\end{equation}

Therefore, imitative learning optimizes the following loss:
\begin{equation}
    \begin{aligned}
    \mathcal{L}_{\textnormal{imi}} = &\lambda_{\textnormal{tex}}\mathcal{L}_{\textnormal{tex}} + \lambda_{\textnormal{id}}\mathcal{L}_{\textnormal{id}} +  \lambda_{\textnormal{lmk}}\mathcal{L}_{\textnormal{lmk}} + \lambda_{\textnormal{ill}}\mathcal{L}_{\textnormal{ill}} + \lambda_{\textnormal{ce}}\mathcal{L}_{\textnormal{ce}}, 
    \end{aligned}
    \label{eq:imi}
\end{equation}
where $\lambda_{(\cdot)}$ denotes the hyper-parameter that balances the terms. 

\begin{figure}
\centering
\setlength\tabcolsep{0pt}
\renewcommand{\arraystretch}{0.0}
\begin{tabular}{cc}
    \includegraphics[width=0.5\linewidth]{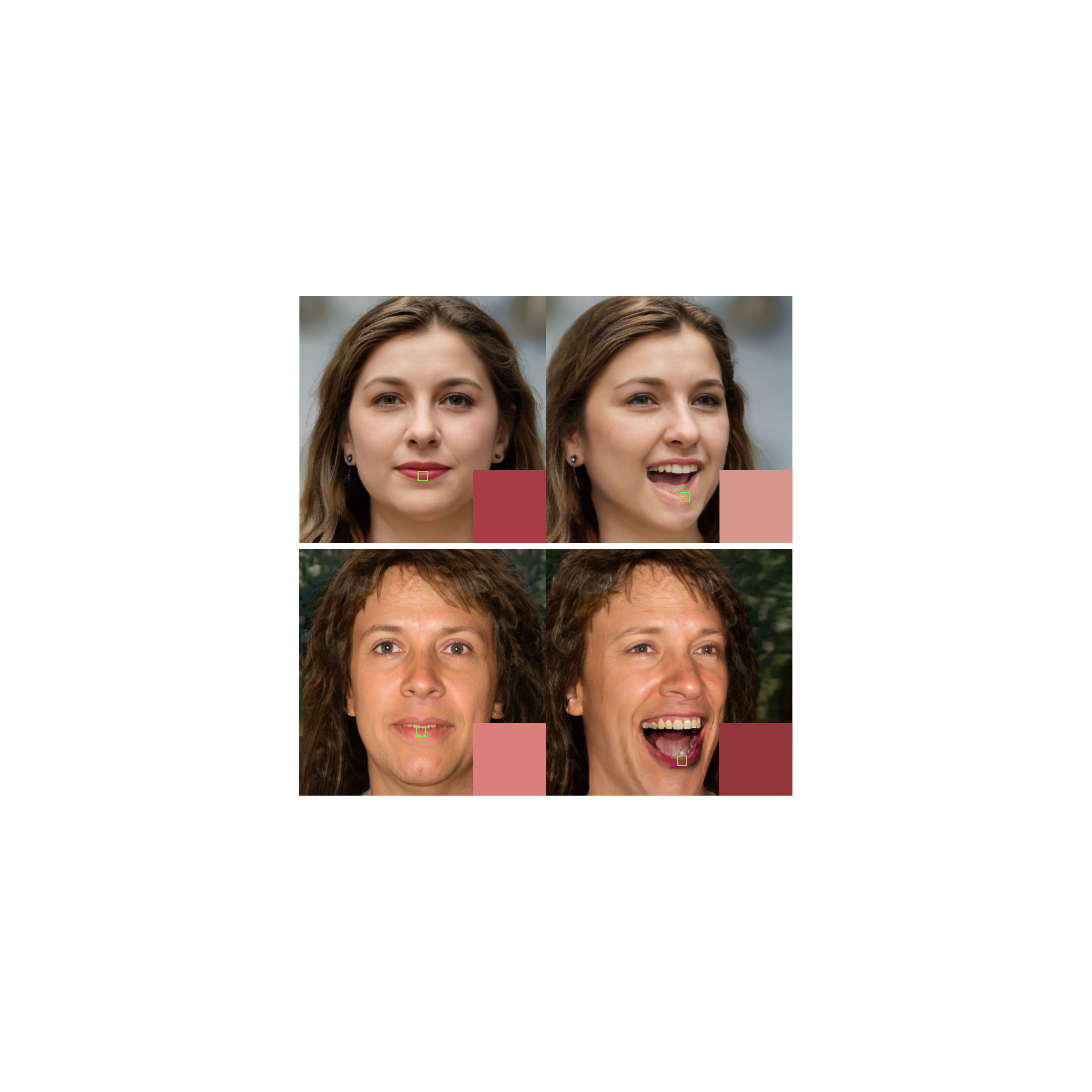} &
    \includegraphics[width=0.5\linewidth]{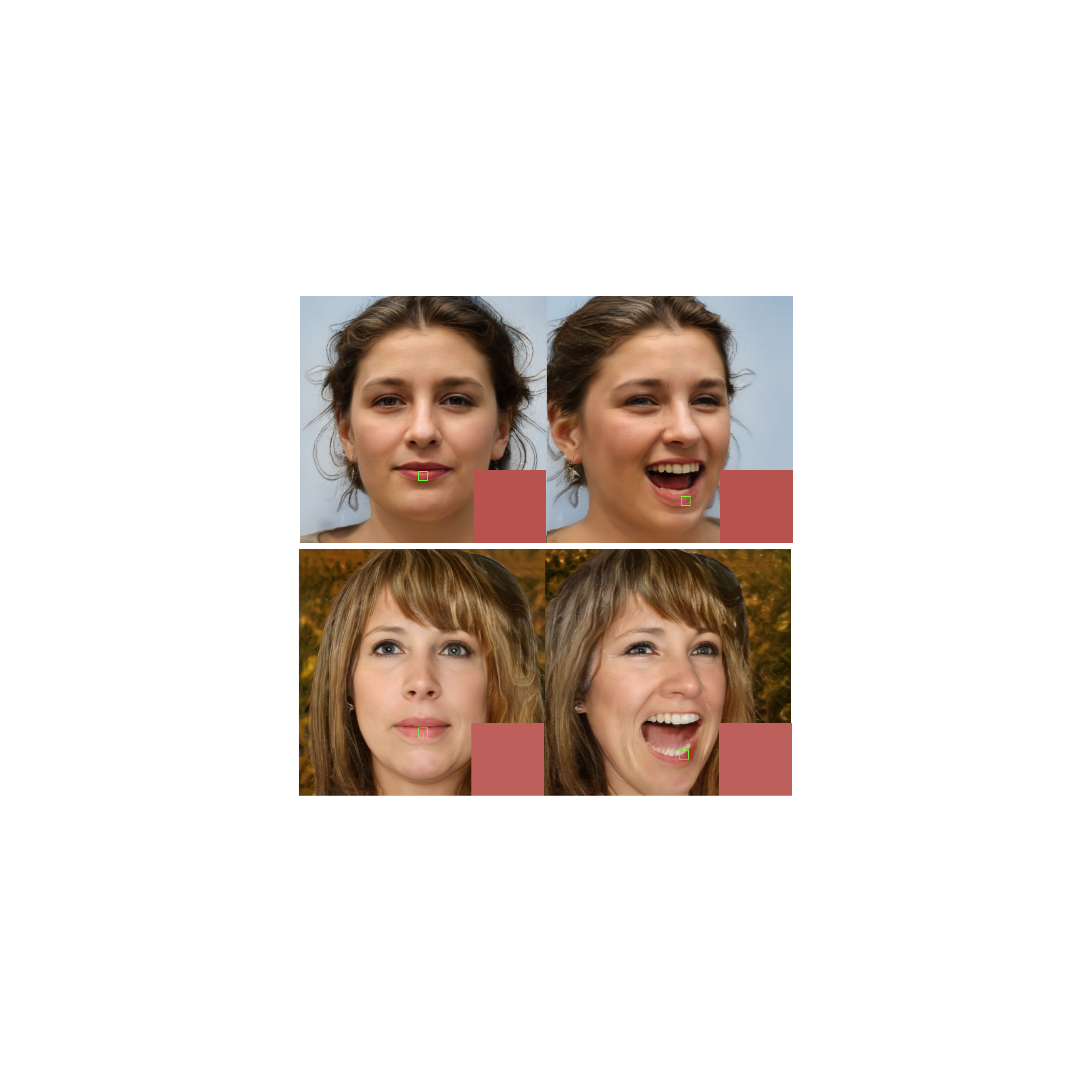}   \\ 
    (a) Ours w/o $\mathcal{L}_{\textnormal{lip}}$ & (b) Ours 
\end{tabular}
\caption{\textbf{The effect of lip style loss.} We visualize results of the same identity using the model trained with (left) or without (right) lip loss, where we highlight the lip color. Training without the lip loss leads to obvious lip appearance change as shown in (a).}
\label{fig:lip}
\end{figure}

\subsubsection{Disentanglement Learning} 
\label{sec:dis}
To learn a more disentangled latent space, the network is further regularized such that the image pairs that differ in one type of control parameter will remain consistent in other properties. Next, we modify one attribute each time and introduce the loss functions when contrasting the identity, expression, and illumination respectively.

When contrasting the image pairs $\bm{I}$ and $\bm{I}^{\bm{\kappa}}$ that differ in the identity parameter $\bm{\kappa}$,  inconsistency is likely to happen in non-face areas like hair, clothes, and backgrounds. To address this issue, we render the semantic mask $\bm{S}$ and convert it to a binary mask $\bm{\tilde{S}_{\textnormal{bg}}}$ with ones indicating the non-face parts. We then enforce the consistency for non-face parts by imposing the identity-aware disentanglement loss :
\begin{equation}
    \begin{aligned}
    &\mathcal{L}^{\kappa}_{\textnormal{dis}} = \norm{ \bm{I}^{\kappa} \odot \bm{\tilde{S}}^{\kappa}_{bg} - \bm{I} \odot \bm{\tilde{S}}_{bg} }_2. 
    \end{aligned}
\end{equation}

Similarly, the image pairs $\bm{I}$ and $\bm{I}^{\gamma}$ with different illumination coefficients $\bm{\gamma}$ should keep the same identity and expression. Hence, the illumination disentanglement is enforced through:
\begin{equation}
    \begin{aligned}
    \mathcal{L}^{\gamma}_{\textnormal{dis}} & = \mathcal{L}^{\gamma}_{\textnormal{id}}(\bm{I}^{\gamma}, \bm{I}) + \mathcal{L}^{\gamma}_{\textnormal{lmk}}(\bm{I}^{\gamma}, \bm{I}).
    \end{aligned}
\end{equation}

Disentangled expression control, on the other hand, is the most challenging. We introduce multiple losses to ensure consistency in various aspects. For the contrastive image pair $\bm{I}$ and $\bm{I}^{\beta}$ with conditioned on different $\bm{\beta}$, we first apply the identity loss $\mathcal{L}^{\beta}_{\textnormal{id}}(\bm{I}^{\beta},\bm{I})$ to ensure identity consistency. To promote the fine-grained appearance similarity, the two images are expected to share the same appearance for corresponding areas. To encourage this, following ~\cite{discofacegan} we compute the 2D flow of rendered 3DMM faces $\bm{R}$ and $\bm{R}^{\beta}$, and use it to warp the portrait $\bm{I}$ which results to $\mathcal{W}(\bm{I})$, where $\mathcal{W}$ denotes the warping operator.  The warped image should be similar to $\bm{I}^\beta$, so the texture loss now becomes:
\begin{equation}
    \mathcal{L}^\beta_{\textnormal{tex}} = \norm{\bm{I}^{\beta}-\mathcal{W}(\bm{I})}_2.
\end{equation} 
Note that the flow computed from the 3DMM faces fails to consider the occlusions by hair and face wearings. Hence  we zero the flow of non-face regions according to the rendered semantic mask when computing $\mathcal{L}^\beta_{\textnormal{tex}}$. 

While the above expression disentanglement losses suffice for most cases, we observe severe lip color change when animating the portraits to open-mouth expressions, as shown in Figure~\ref{fig:lip}(a). We conjecture that this arises from the training data bias that faces with open-mouth expressions are rare, and the learned model inevitably shows some appearance tendency. To solve this, we propose a lip style loss that encourages the appearance consistency of the lip region. Specifically, we obtain the lip mask $\bm{S}_{\textnormal{lip}}$ from the learned semantic field and calculate the style loss~\cite{huang2017arbitrary} with the pretrained VGG model~\cite{vgg}:
\begin{equation}
    \begin{aligned}
    &\mathcal{L}_{\textnormal{lip}} = \sum^{2}_{i=1}||\mu(\phi_i(\bm{I}^{\beta}) \odot \bm{S}^{\beta}_{\textnormal{lip}})-\mu(\phi_i(\bm{I})\odot \bm{S}_{\textnormal{lip}})||_2 + \\
    &~~~~~~~~~~~\sum^{2}_{i=1}||\nu(\phi_i(\bm{I}^{\beta}) \odot \bm{S}^{\beta}_{\textnormal{lip}})-\nu(\phi_i(\bm{I}) \odot \bm{S}_{\textnormal{lip}})||_2, \\
    \end{aligned}
\end{equation}
where $\phi_i$ denotes the activation of $i$th layer whereas $\mu(\cdot)$ and $\nu(\cdot)$ calculate the feature mean and variance respectively. As shown in Figure~\ref{fig:lip}, this style loss effectively preserves the lip appearance during animation. Therefore,  overall expression disentangle loss is:
\begin{equation}
    \begin{aligned}
    \mathcal{L}^{\beta}_{\textnormal{dis}} =  \lambda^{\beta}_{\textnormal{id}}\mathcal{L}^{\beta}_{\textnormal{id}} + \lambda^\beta_{\textnormal{tex}}\mathcal{L}^\beta_{\textnormal{tex}} + \lambda^\beta_{\textnormal{lip}}\mathcal{L}^\beta_{\textnormal{lip}},
    \end{aligned}
    \label{eq:dis}
\end{equation}
where $\lambda$ are the weights for different terms.

Once imitative learning converges, we embark on the disentanglement learning.  To achieve the disentanglement of different attributes,  we additionally enforce the following losses to the generator: 
\begin{equation}
    \mathcal{L}_{\textnormal{dis}} = \mathcal{L}^{\kappa}_{\textnormal{dis}} + \mathcal{L}^{\gamma}_{\textnormal{dis}} + \mathcal{L}^{\beta}_{\textnormal{dis}}.
\end{equation}

\subsubsection{Training Details} 
In the two-stage adversarial training, we use Adam optimizer~\cite{kingma2014adam} with $\beta_1=0, \beta_2=0.99$, and initialize the learning rate to $0.002$ for the generator and $0.0025$ for the discriminator, respectively. 
As for the volumetric  rendering, 96 points are sampled for each ray where 48 points for uniform sampling and 48 for importance sampling. 

We train and evaluate our method at $512\times 512$ resolution on the FFHQ dataset.
We utilize an exponential moving average (EMA) of model weights for inference.
The imitative learning is trained for 300,000 iterations with a batch size of 4 and the disentanglement learning takes another 100,000 iterations with a batch size of 2. 
The loss weights in Equation~\ref{eq:imi} are set to $\lambda_{\textnormal{tex}}=10, \lambda_{\textnormal{id}}=10, \lambda_{\textnormal{lmk}}=10, \lambda_{\textnormal{ill}}=1e3$. The weights in Equation~\ref{eq:dis} are set to $\lambda_{\textnormal{id}}^{\beta}=100, \lambda_{\textnormal{flow}}=50, \lambda_{\textnormal{lip}}=100$. 
The overall training process takes 14 days on 8 NVIDIA Tesla 32GB V100 GPUs.

\subsection{Explicit Volume Blending}
\label{sec:blend}

\begin{figure}[t]
    \centering
    \includegraphics[width=\linewidth]{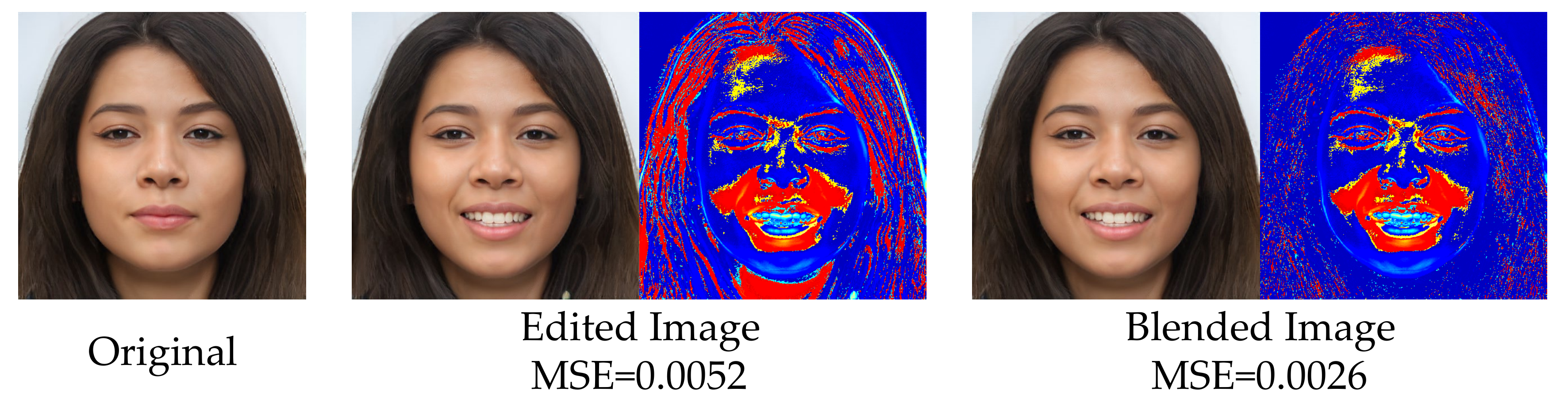}
    \caption{\textbf{The error map during the facial animation.} The error map is computed as the per-pixel error between the original pixel color and the edited image. The volume blending significantly reduces the error map in the non-face region, such as hair, shoulder and background, when changing the facial expression.}
    \label{fig:blend_error}
\end{figure}

So far, we can achieve disentangled control with the above two-stage training. Nonetheless, the disentanglement learning somehow compromises generation diversity. Reducing the strength of disentanglement learning, however, leads to obvious appearance inconsistency. As shown in Figure~\ref{fig:blend_error}, such inconsistency mainly happens in the non-face areas such as hair, clothes and backgrounds. Motivated by this, we propose a volume blending scheme during inference that allows a smaller strength of disentanglement learning while preserving consistency during the expression animation. 

\begin{figure}
    \centering
    \includegraphics[width=\linewidth]{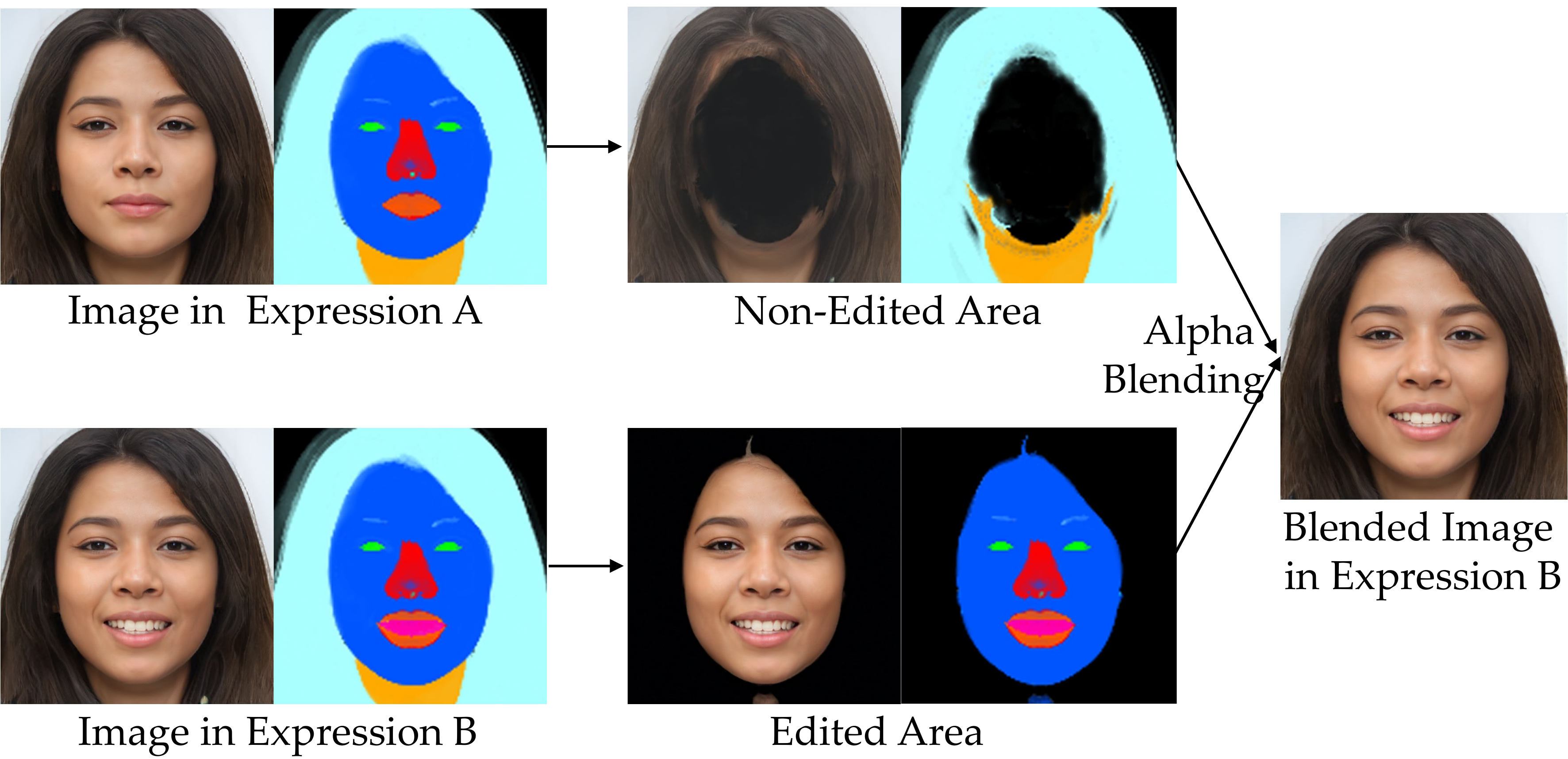}
    \caption{\textbf{The volume blending.} During the expression animation, the appearances of non-face regions such as hair and background possibly change. To amend this, we remain the radiance field of these regions the same as the original neutral inputs. }
    \label{fig:blend}
\end{figure}

Figure~\ref{fig:blend} illustrates the volume blending process. Given the latent code $\bm{z}=(\bm{\epsilon}, \bm{\kappa}, \bm{\beta}, \bm{\gamma})$ and $\bm{z}'=(\bm{\epsilon}, \bm{\kappa}, \bm{\beta}', \bm{\gamma})$ with animated expression coefficients, they yield two separate radiance fields ---  $(\bm{c}, \bm{\sigma}, \bm{s})$ and $(\bm{c}', \bm{\sigma}', \bm{s}')$, where $\bm{c}$, $\bm{\sigma}$ and $\bm{s}$ denote the view-dependent color, density and semantics of the field respectively. To remain the consistency during animation, we produce a composite radiance field $(\bm{\tilde{c}}, \bm{\tilde{\sigma}})$, which is blended from the controlled output and the original radiance field, \ie,
\begin{equation}
    (\bm{\tilde{c}},\bm{\tilde{\sigma}})= \big(\bm{w}\bm{c}+(1-\bm{w})\bm{c}', \bm{w}\bm{\sigma}+(1-\bm{w})\bm{\sigma}'\big),
\end{equation}
where $\bm{w}_p$ indicates the probability of belonging to the face part for the location $\bm{p}$, which is inferred from the learned semantic field $\bm{s}'$. We opt for volume blending over image-level blending since the former guarantees view consistency and brings fewer blending artifacts. In Figure~\ref{fig:blend_error}, we show that the temporal appearance variation of the non-edited area during animation is significantly reduced using the blending strategy. Both the qualitative and quantitative studies show much improved consistency in the disentangled control.

\section{Experiments}
\subsection{Dataset}
We adopt 70,000 wild images with $512\times 512$ resolutions in FFHQ~\cite{stylegan, stylegan2} as the training dataset. 
We utilize a face reconstruction method~\cite{deep3dface} to extract the 3DMM coefficients for each real image, and leverage these extracted coefficients as the training set to train VAE-GANs for control parameters. 

Training on FFHQ, however, gives dull expressions. To improve the expression diversity, we additionally leverage the expression coefficients collected from the emotional video dataset RAVDESS~\cite{ravdess}. 
This dataset contains 24 professional actors including 12 females and 12 males who act with exaggerated expressions. In practice, for each actor, we select 10 videos and sample 400 images from each video, resulting in 96,000 images in total. We extract expression coefficients from these images and combine these coefficients together with the expressions extracted from FFHQ to train the expression-related VAE-GAN. 

\subsection{Image Preprocessing}
\label{sec:preprocess}
\begin{figure}[t]
    \centering
    \includegraphics[width=\linewidth]{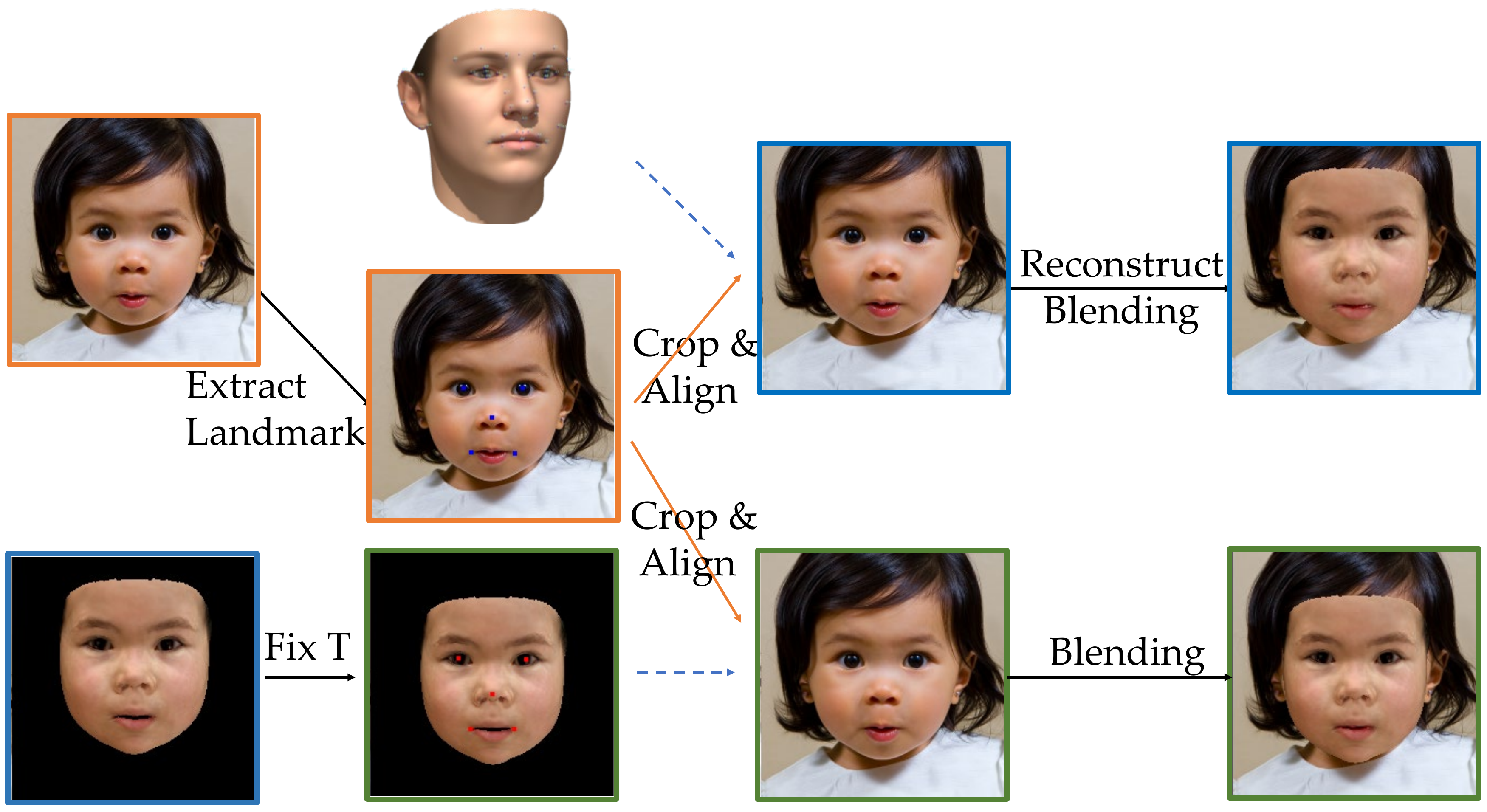}
    \caption{\textbf{The data alignment process.} The Image in orange is the original input, while the image in blue is the corresponding 3D reconstruction. ``Fix T'' refers to setting the translation to a predefined value. We use the aligned face (with green box) for training. Here, we blend the 3DMM rendering into the aligned image to validate the alignment.}
    \label{fig:preprocess}
\end{figure}

In order to align the neural radiance space and the 3DMM face, we extract 5-point landmarks using MTCNN~\cite{zhang2016joint} for the original face image and meanwhile determine the 3D landmarks of a canonical 3DMM model, which we use for face alignment. Then we reconstruct the 3DMM parameters for the face image. 
After that, for each 3DMM, we set all the translation coefficients as a pre-defined value so as to place the reconstructed face in a proper image location, which results in a new 3DMM mesh along with corresponding 3D landmarks. 
We compute the affine transformation that aligns the original five landmarks and the new 3DMM landmarks, and apply the computed affine matrix to produce aligned faces for training.
As a result, images are aligned with their corresponding 3DMM reconstruction in the canonical space. The image alignment process  is illustrated in Figure~\ref{fig:preprocess}. We show the importance of using aligned training data in Sec.~\ref{sec:ab_preprocess}.

\begin{figure*}
    \centering
    \footnotesize
    \includegraphics[width=\linewidth]{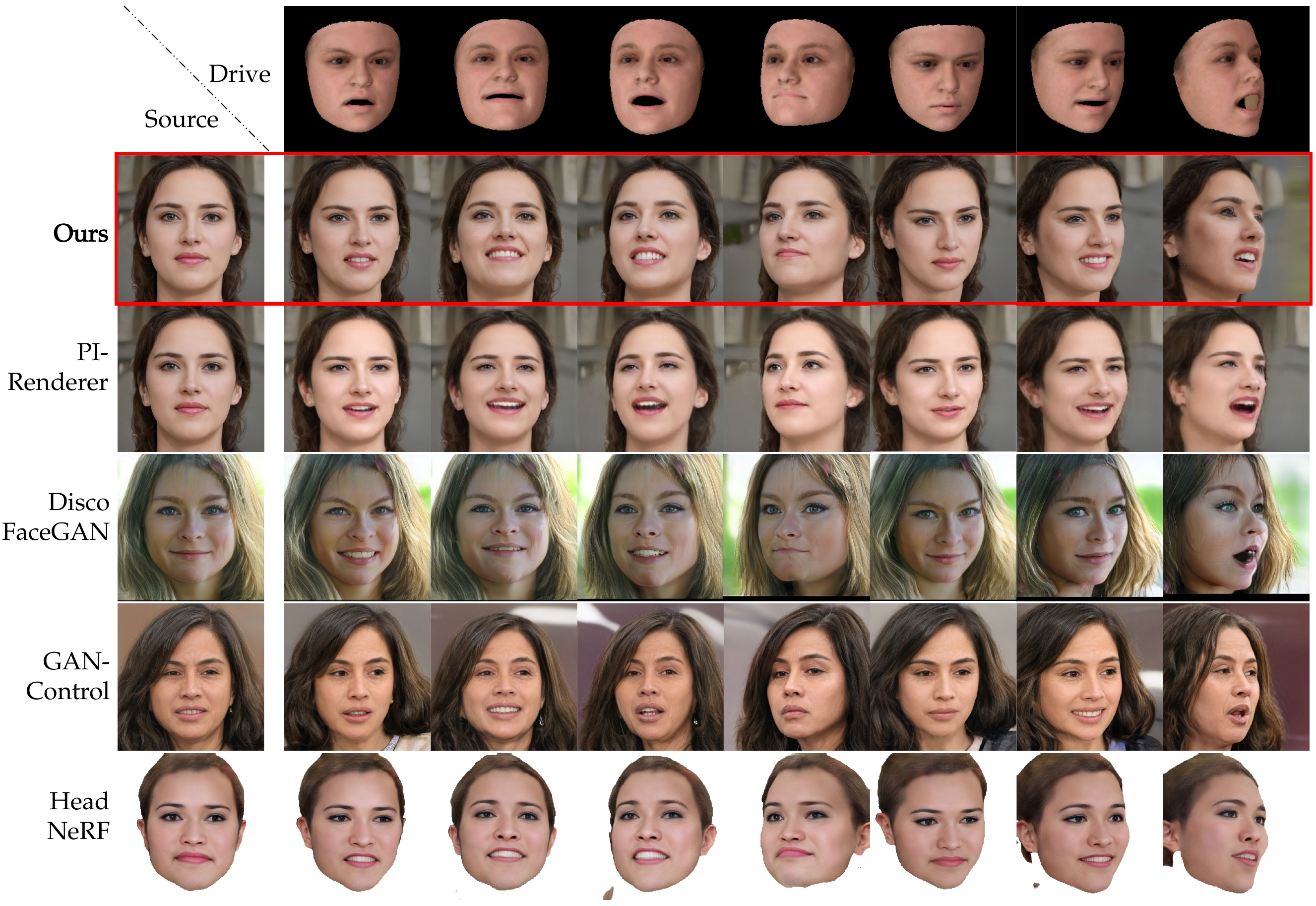}
    \caption{\textbf{Visual comparison with PIRenderer~\cite{pirenderer},  DiscoFaceGAN~\cite{discofacegan}, GAN-Control~\cite{gan-control} and HeadNeRF~\cite{gan-control}.} Given a source image and control parameters, we use different approaches to generate faces with animated expressions in different head poses. For the face reenactment method PIRenderer, its input source image is produced by our method.}
    \label{fig:compare}
\end{figure*}

\begin{figure*}
\center
\setlength\tabcolsep{0pt}
{
\renewcommand{\arraystretch}{0.0}
\footnotesize
\begin{tabular}{cc}
    \includegraphics[width=0.47\linewidth]{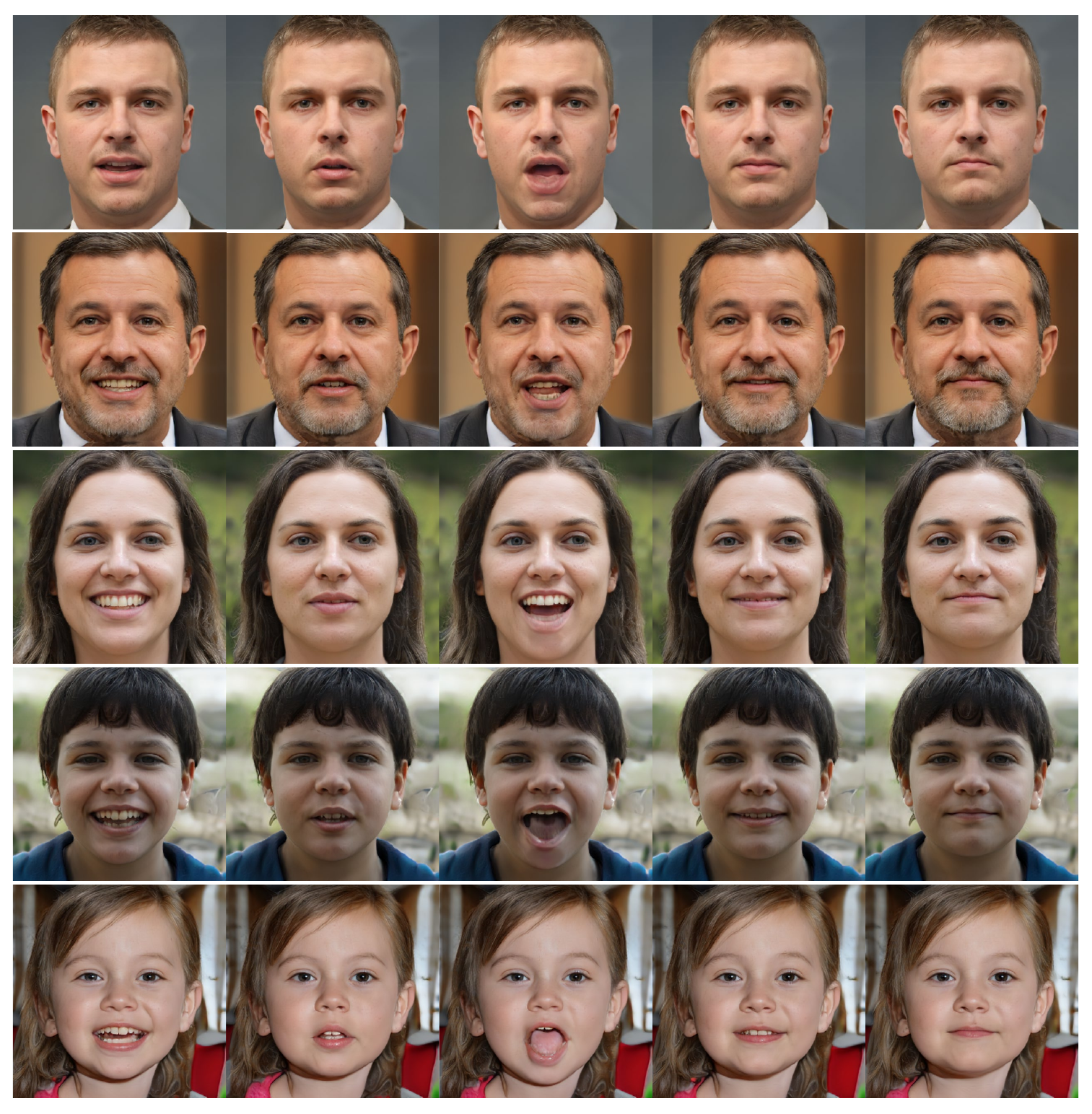} &
    \includegraphics[width=0.47\linewidth]{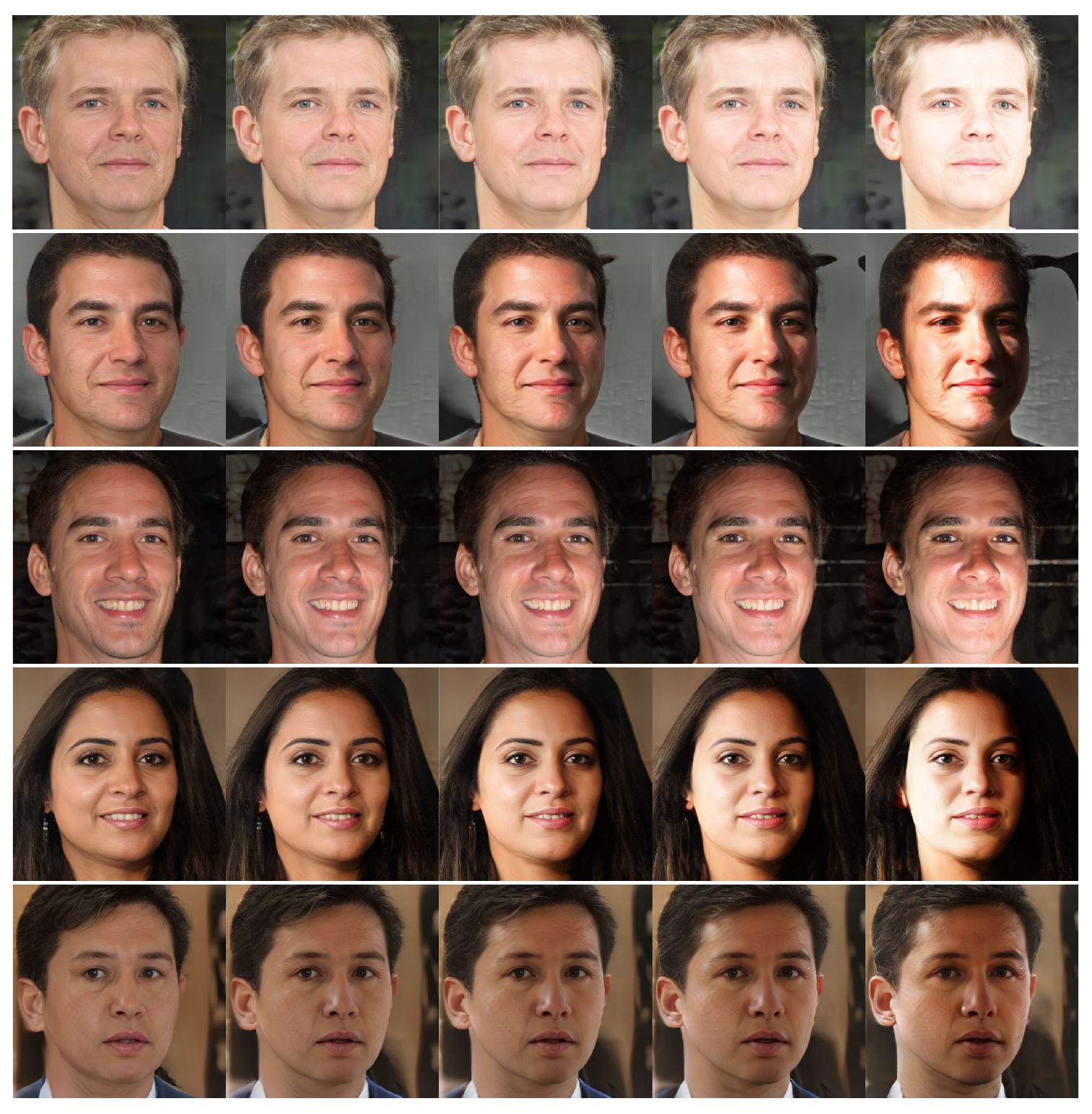} \\
    (a) Varying expressions & (b) Varying illuminations\\
    \includegraphics[width=0.47\linewidth]{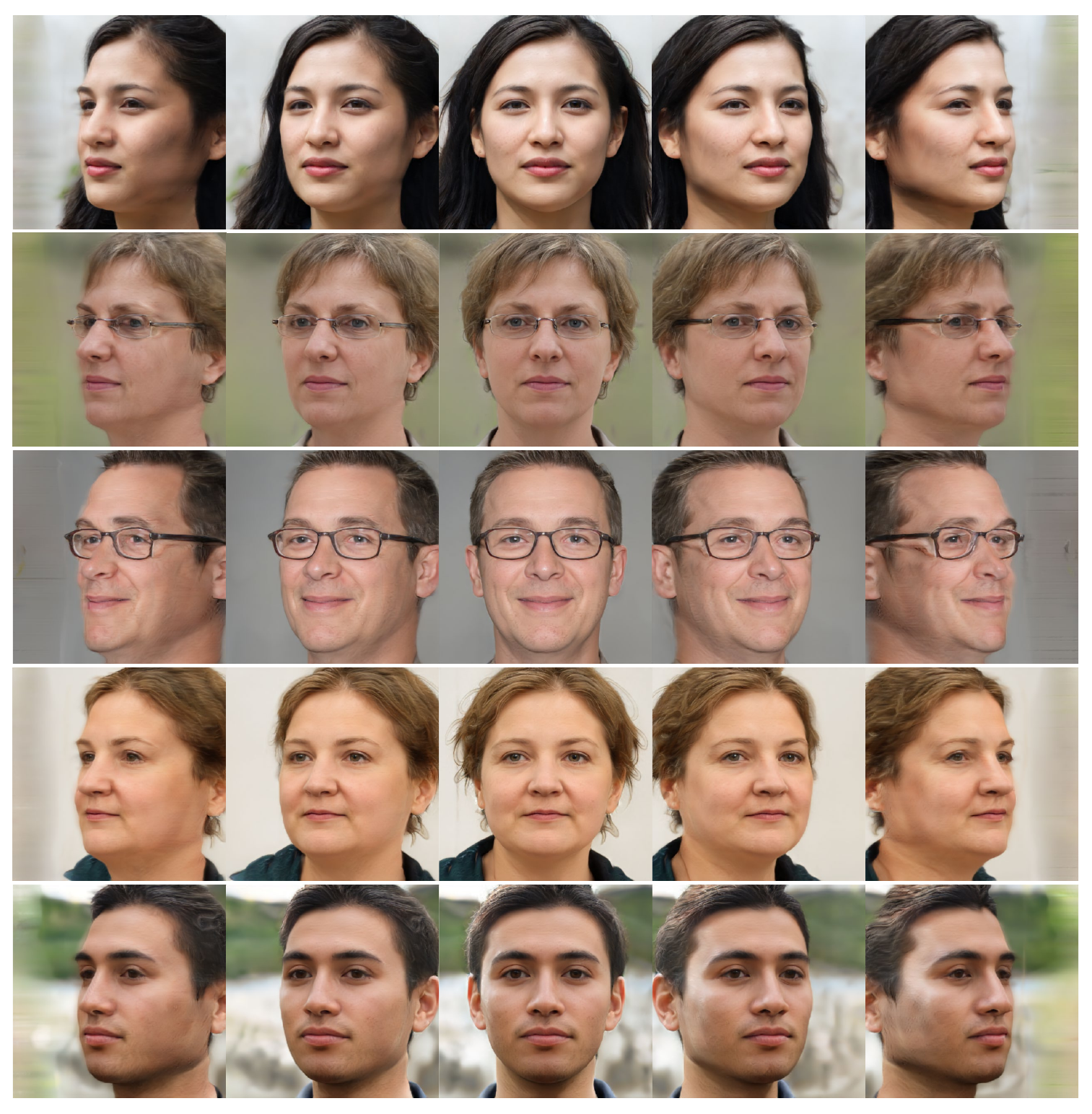} &
    \includegraphics[width=0.47\linewidth]{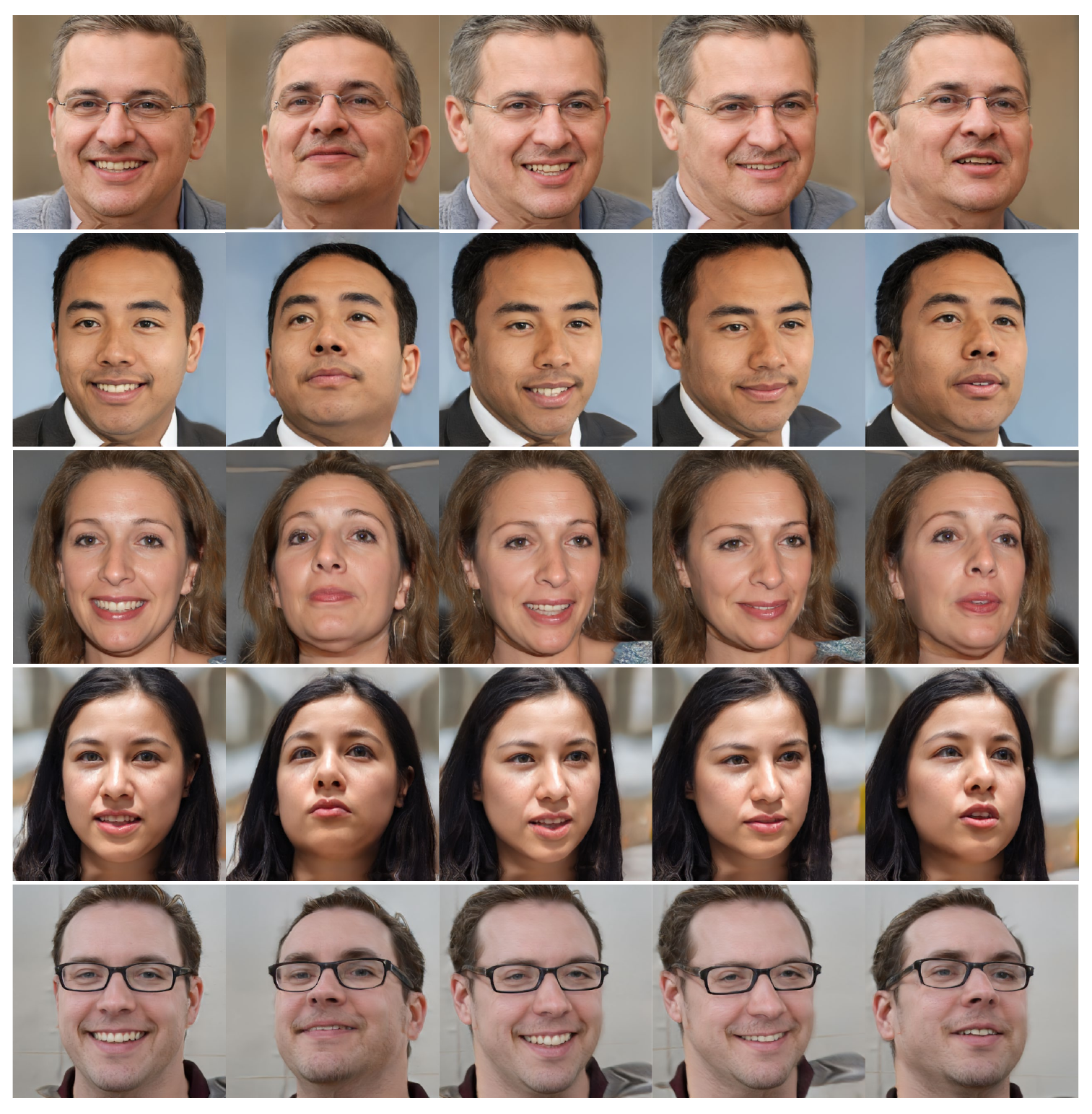} \\
    (c) Varying poses & (d) Varying expressions and poses\\
\end{tabular}
}
\caption{\textbf{The controllable generation results.} We vary expressions, illuminations and poses, independently, which are shown in (a), (b), and (c). We also vary both expressions and poses as shown in (d). 
The proposed approach can precisely control expression, illumination and pose while preserving the identity and other attributes.}
\label{fig:vary}
\end{figure*}



\subsection{Qualitative Comparison}
\subsubsection{Portrait Image Animation}

We compare our method with prior controllable portrait works including DiscoFaceGAN~\cite{discofacegan}, GAN-Control~\cite{gan-control}, HeadNeRF~\cite{HeadNeRF} and PIRenderer~\cite{pirenderer}. 
The first two methods are the 2D generative methods that also support 3DMM control. HeadNeRF~\cite{HeadNeRF} is a NeRF-based parametric head model.
PIRenderer~\cite{pirenderer} is a state-of-the-art  face reenactment method that animates the source image according to a driving video. To adapt it to controllable portrait animation, we use neutral faces generated by our method as the source images and use a pretrained PIRenderer model to animate the face into target expressions and poses.

The visual comparison is shown in Figure~\ref{fig:compare}.  PIRenderer has difficulties in producing high-quality results and preserving the identity on large poses. DiscoFaceGAN can accurately control the expression, but still fails to maintain the consistency of  non-face areas (\eg, glasses, hair and background) when varying expressions and poses. 
Similarly, GAN-Control can always generate high-quality results but may demonstrates severe inconsistency, especially during the pose control. Compared with these image-based methods, 
HeadNeRF adopts the 3D neural representation and thus ensures perfect multi-view consistency. Yet the generated images lack fine details, and hence exhibit limited perceptual quality. In contrast, our method produces the most compelling images with consistent appearance when viewed from different angles.

We further present more visual results of our method in terms of varied  expressions, illuminations and camera poses in Figure~\ref{fig:vary}, where our method animates portrait images in a disentangled manner. We also change the poses and expressions simultaneously, and our method shows impressive control capability. The non-face areas (\eg, glasses, hairs, background) that should not be affected by the face control parameters are temporally consistent during the semantic control. 

\begin{figure*}
\centering
\setlength\tabcolsep{0pt}
\footnotesize
\renewcommand{\arraystretch}{0.0}
\begin{tabular}{ccccc}
    \footnotesize
    \includegraphics[width=0.2\linewidth]{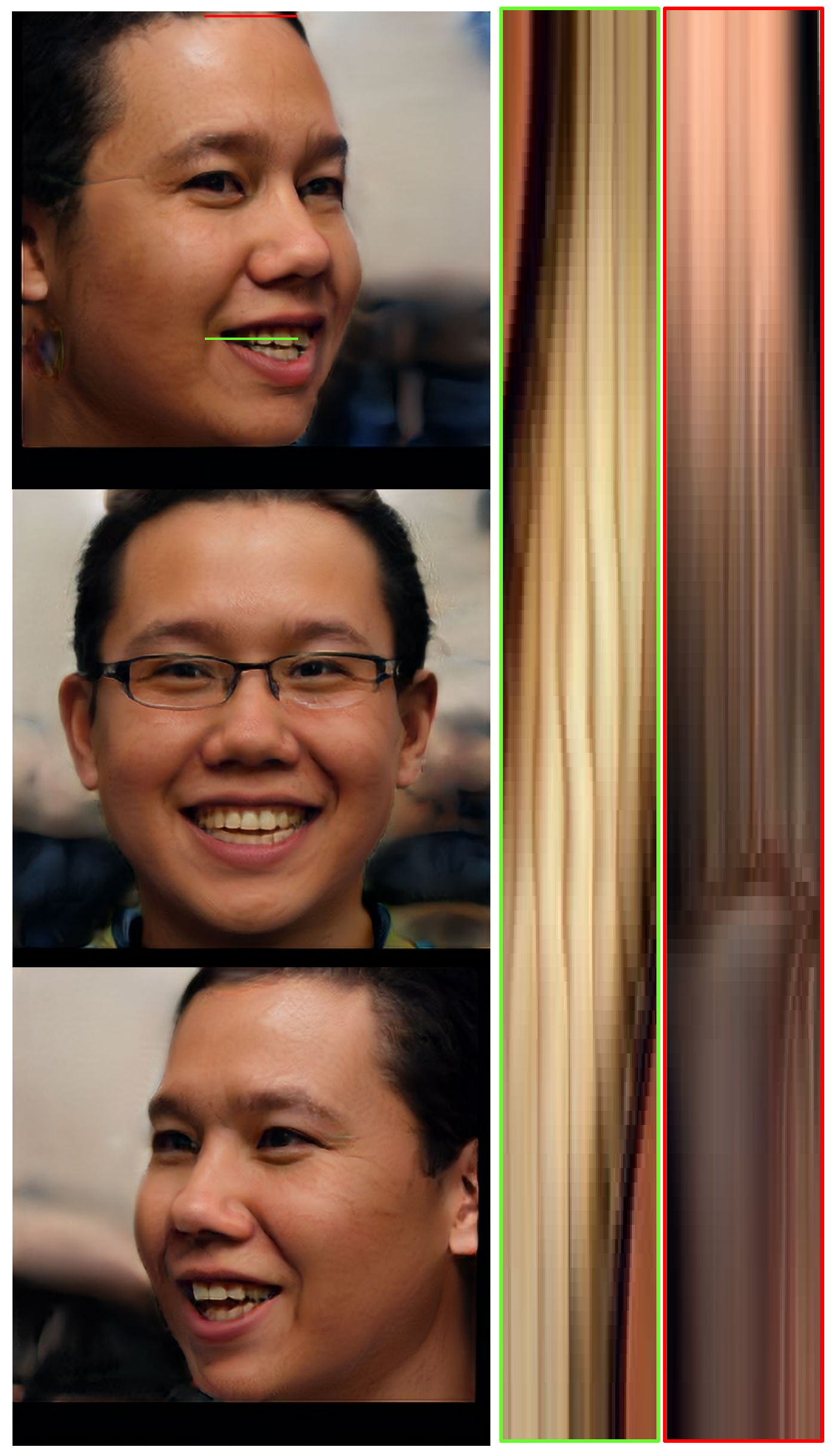} &
    \includegraphics[width=0.2\linewidth]{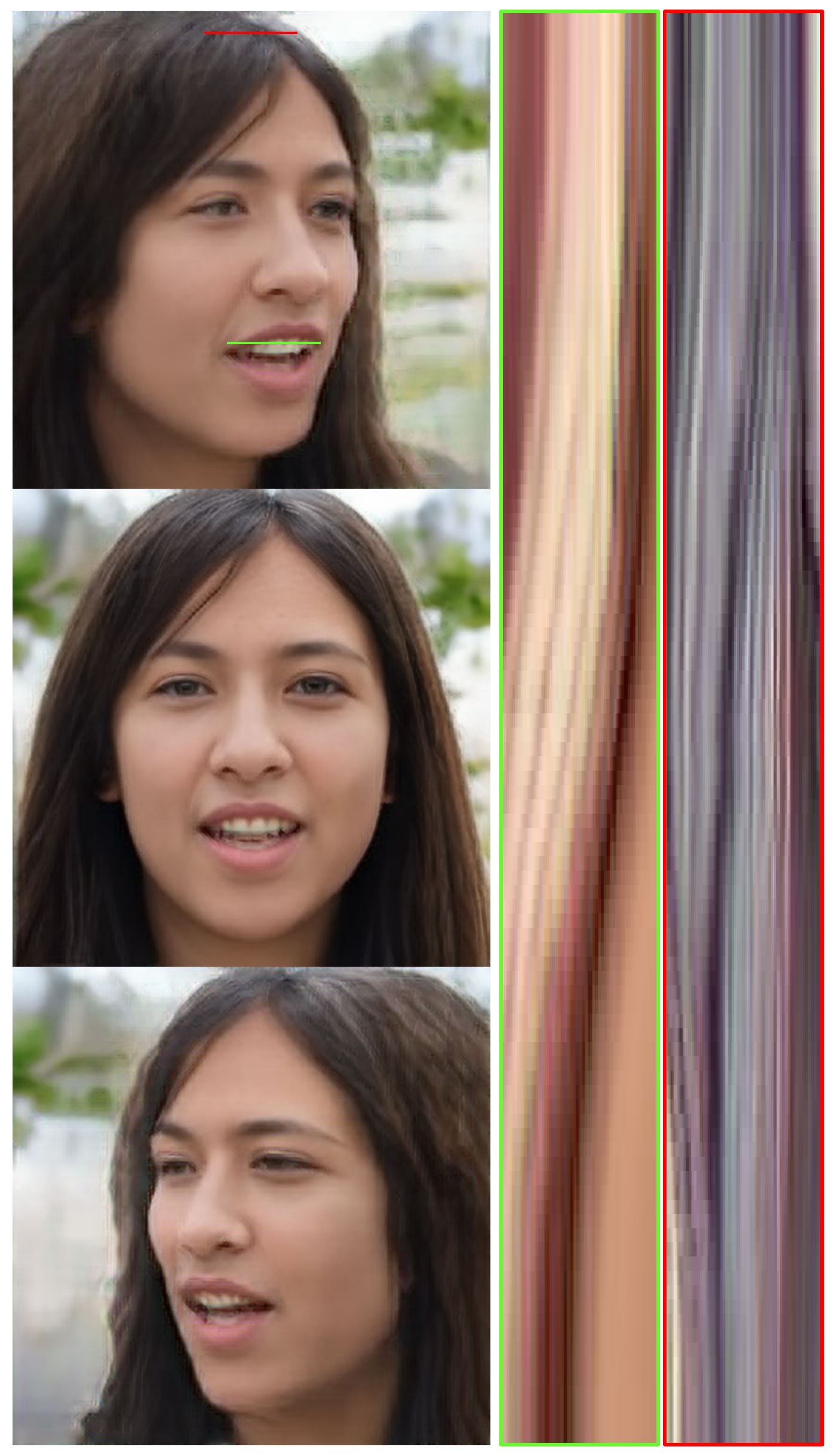}  & 
    \includegraphics[width=0.2\linewidth]{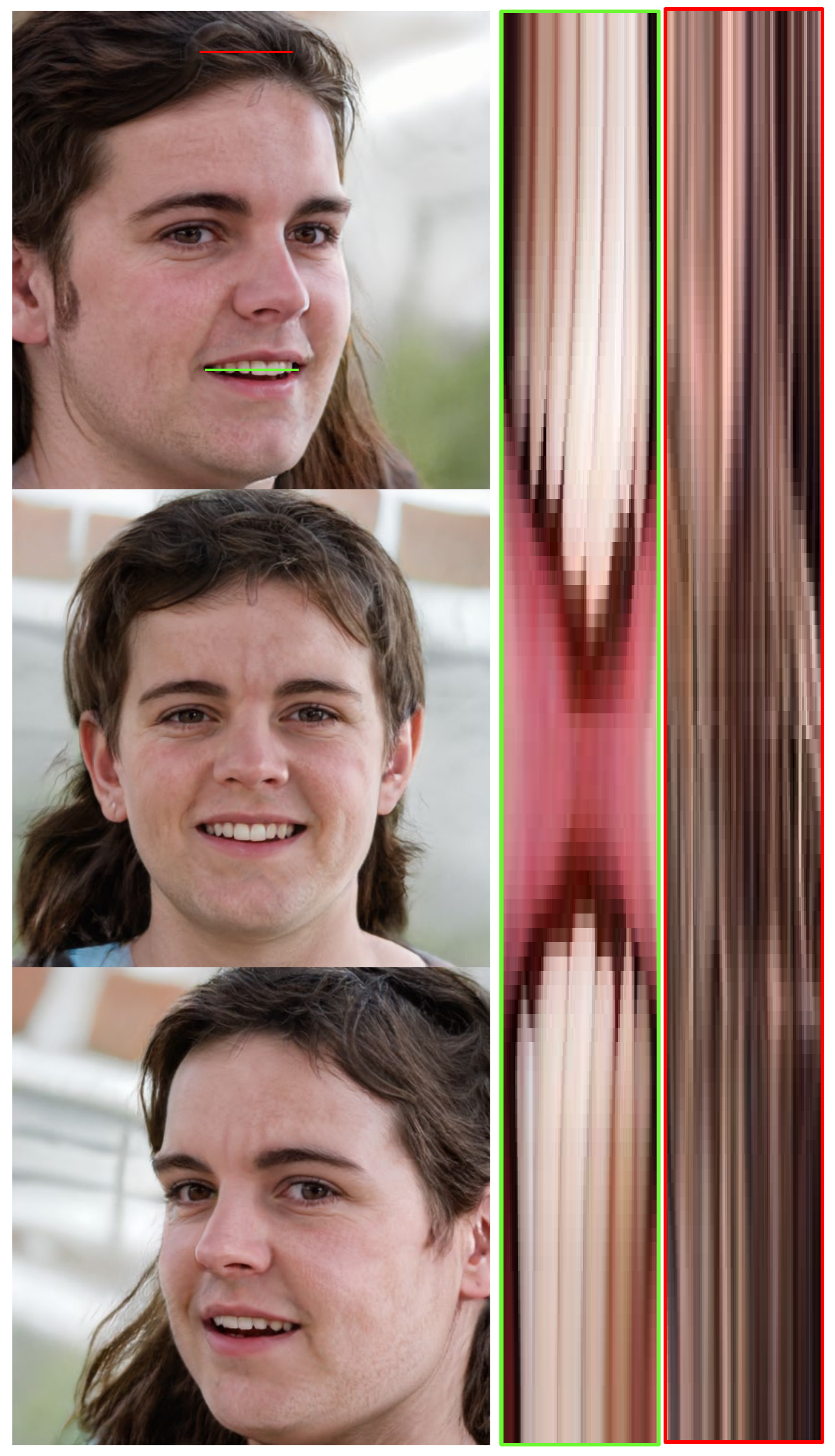}  & 
    \includegraphics[width=0.2\linewidth]{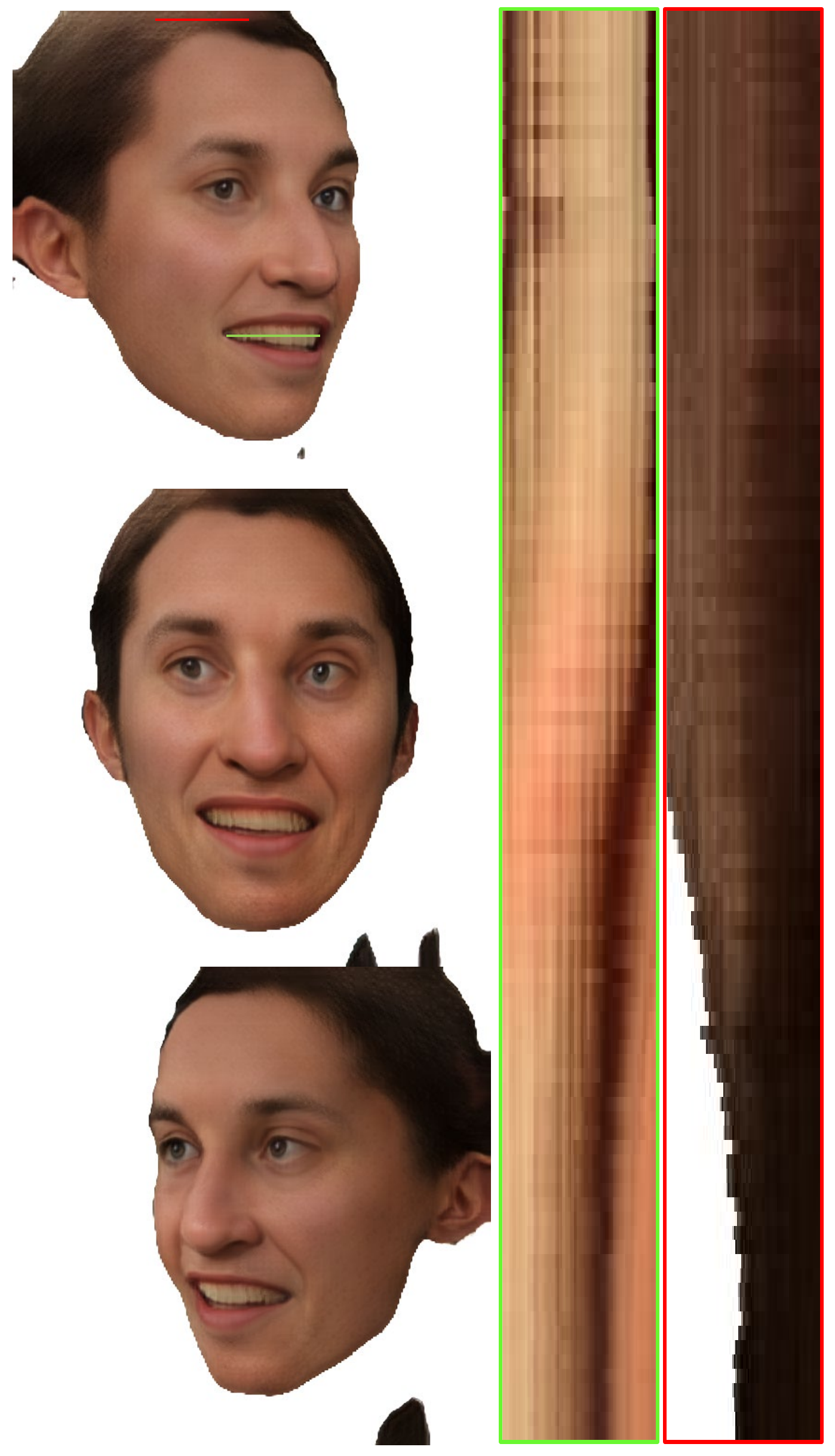} &
    \includegraphics[width=0.2\linewidth]{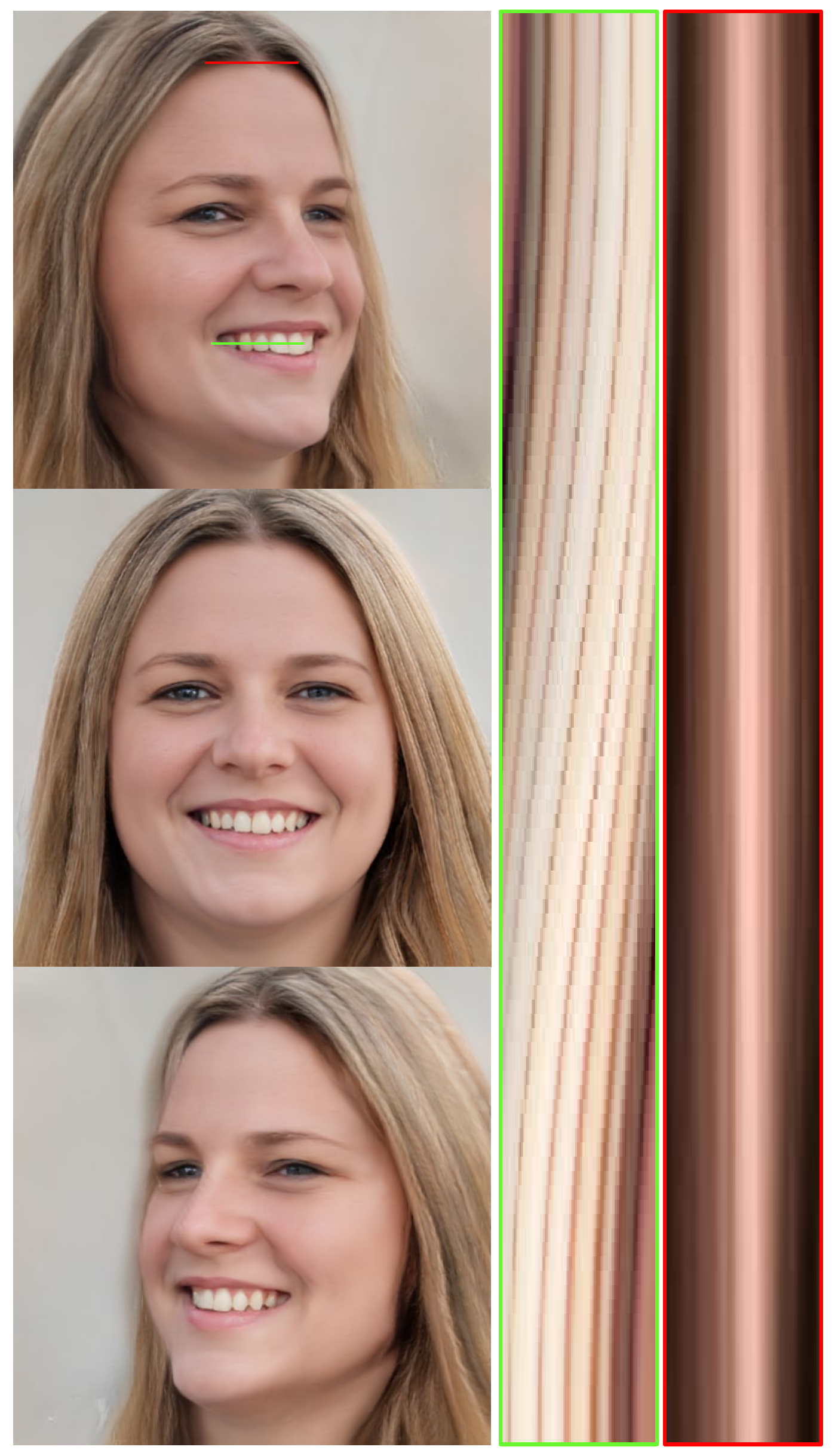} \\ 
    \rule{0pt}{2ex} 
    DiscoFaceGAN & PIRenderer & GAN-Control & HeadNeRF & Ours 
\end{tabular}
\caption{\textbf{Visualization of multi-view consistency on teeth (green) and hair (red) area.} We visualize the texture along a straight line when changing poses. The view-consistent results should demonstrate smooth pattern in the visualization.}
\label{fig:cons_pose}
\end{figure*}

\begin{figure*}
\centering
\setlength\tabcolsep{0pt}
\footnotesize
\renewcommand{\arraystretch}{0.0}
\begin{tabular}{cccccc}
   \footnotesize
    \includegraphics[width=0.167\linewidth]{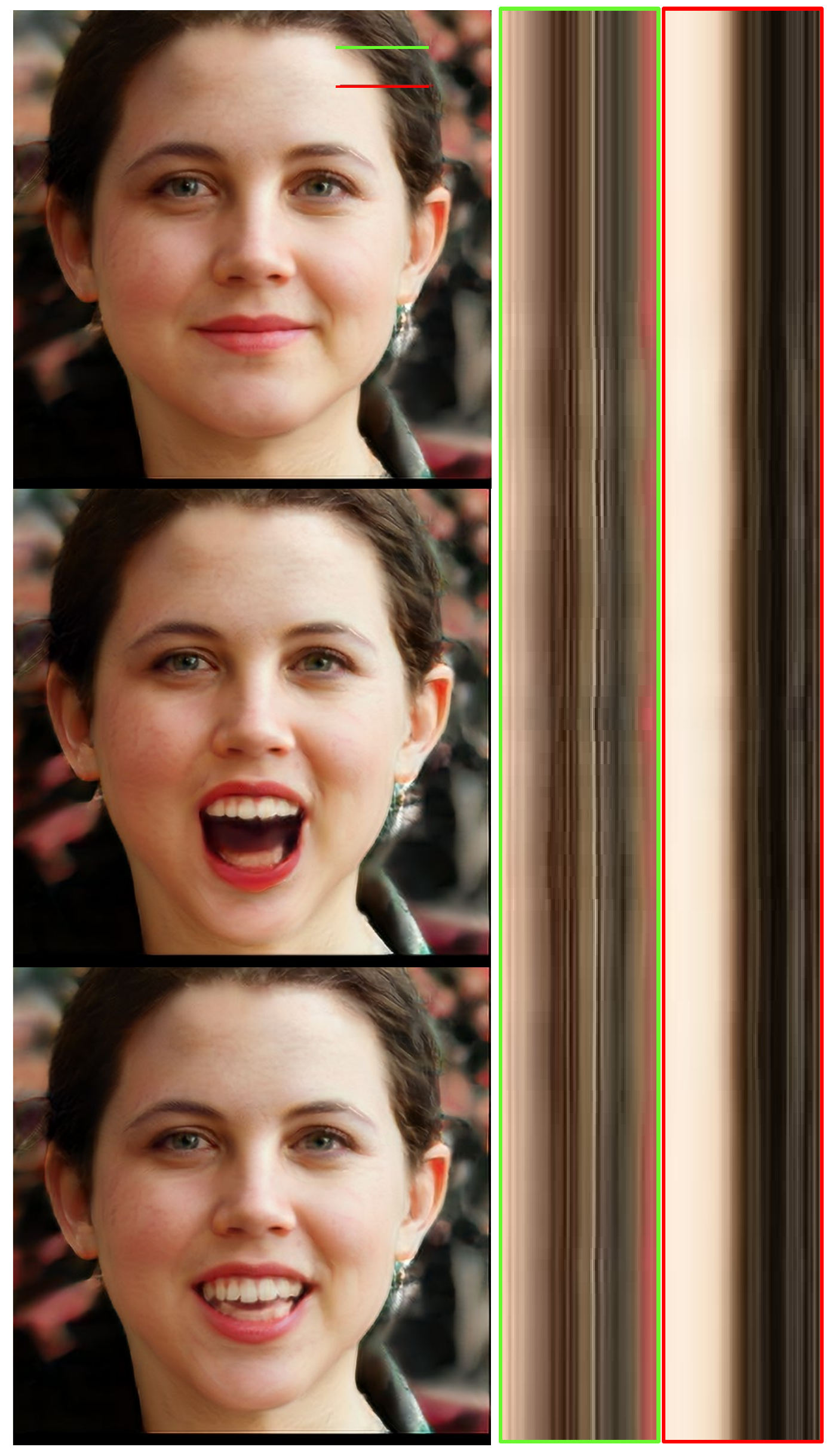} &
    \includegraphics[width=0.167\linewidth]{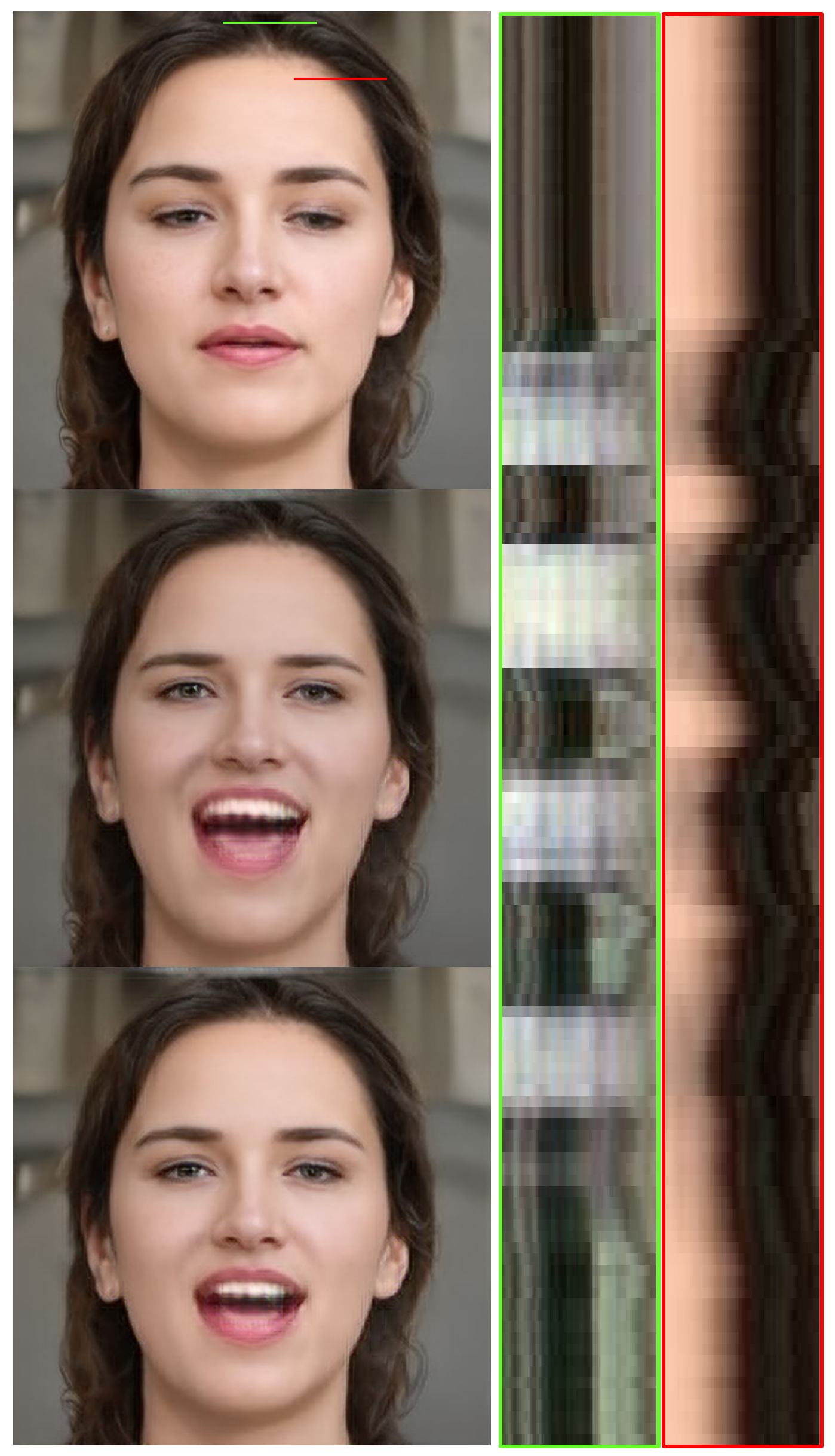}  & 
    \includegraphics[width=0.167\linewidth]{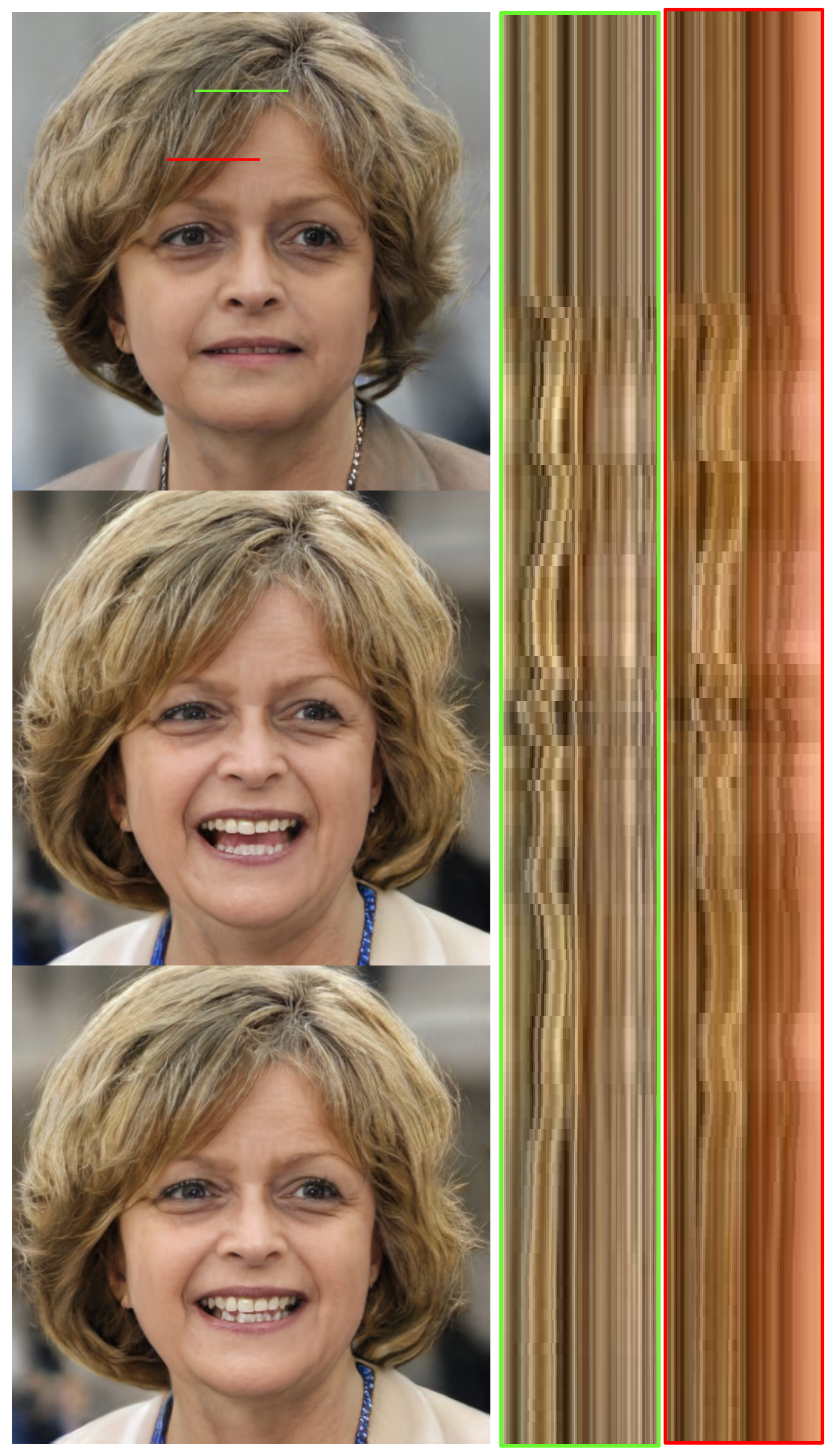}  & 
    \includegraphics[width=0.167\linewidth]{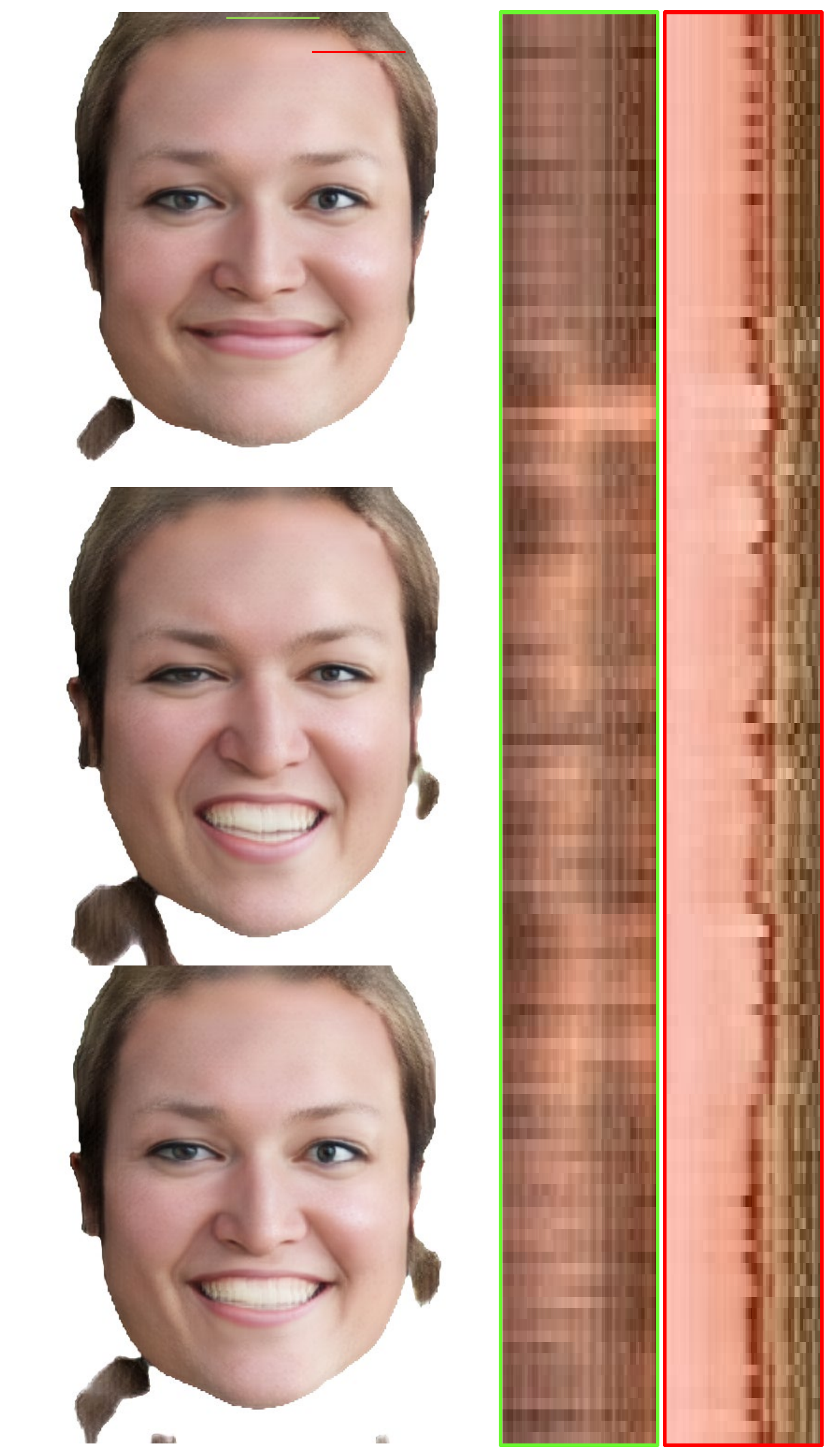} &
    \includegraphics[width=0.167\linewidth]{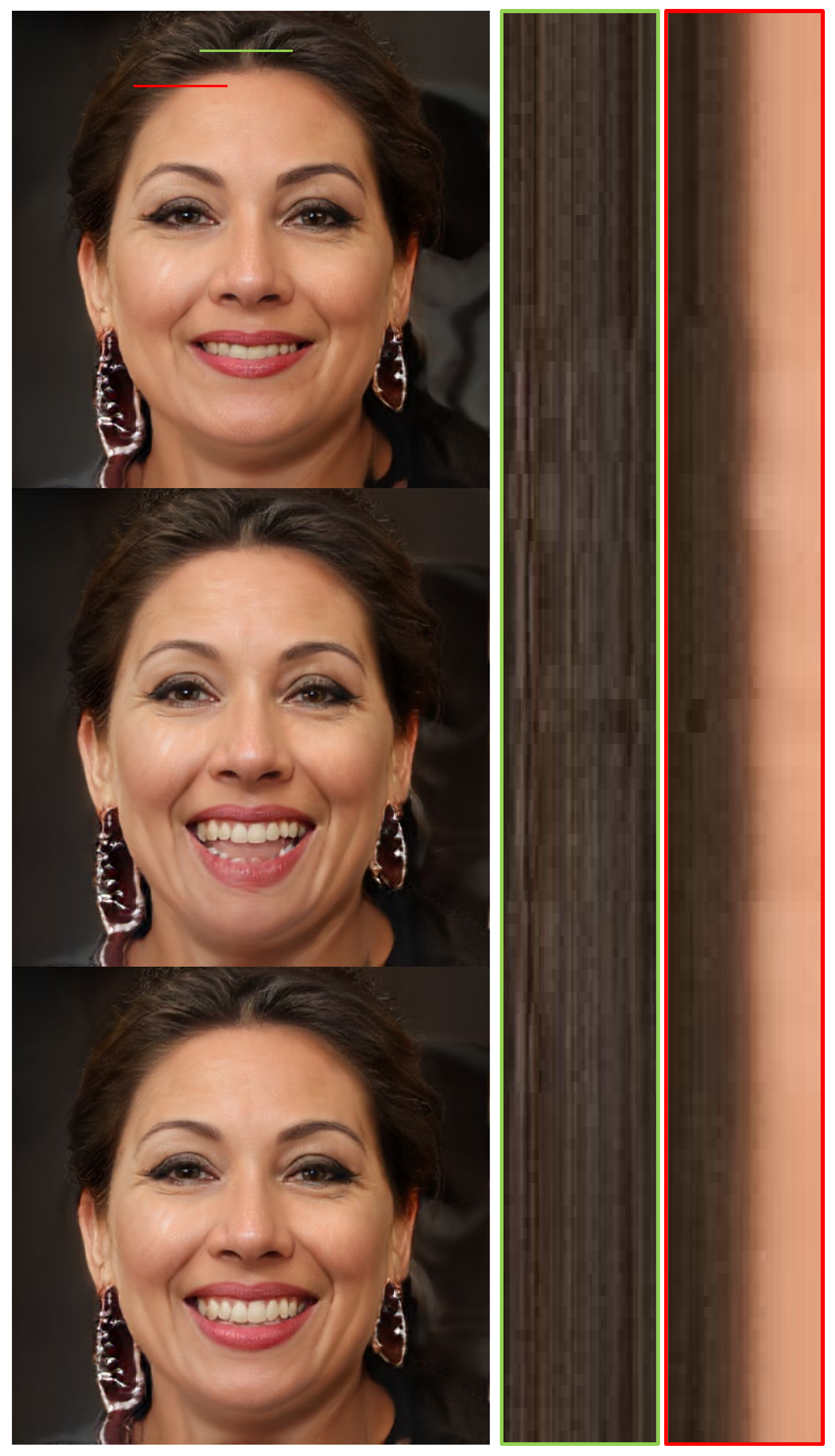} &
    \includegraphics[width=0.167\linewidth]{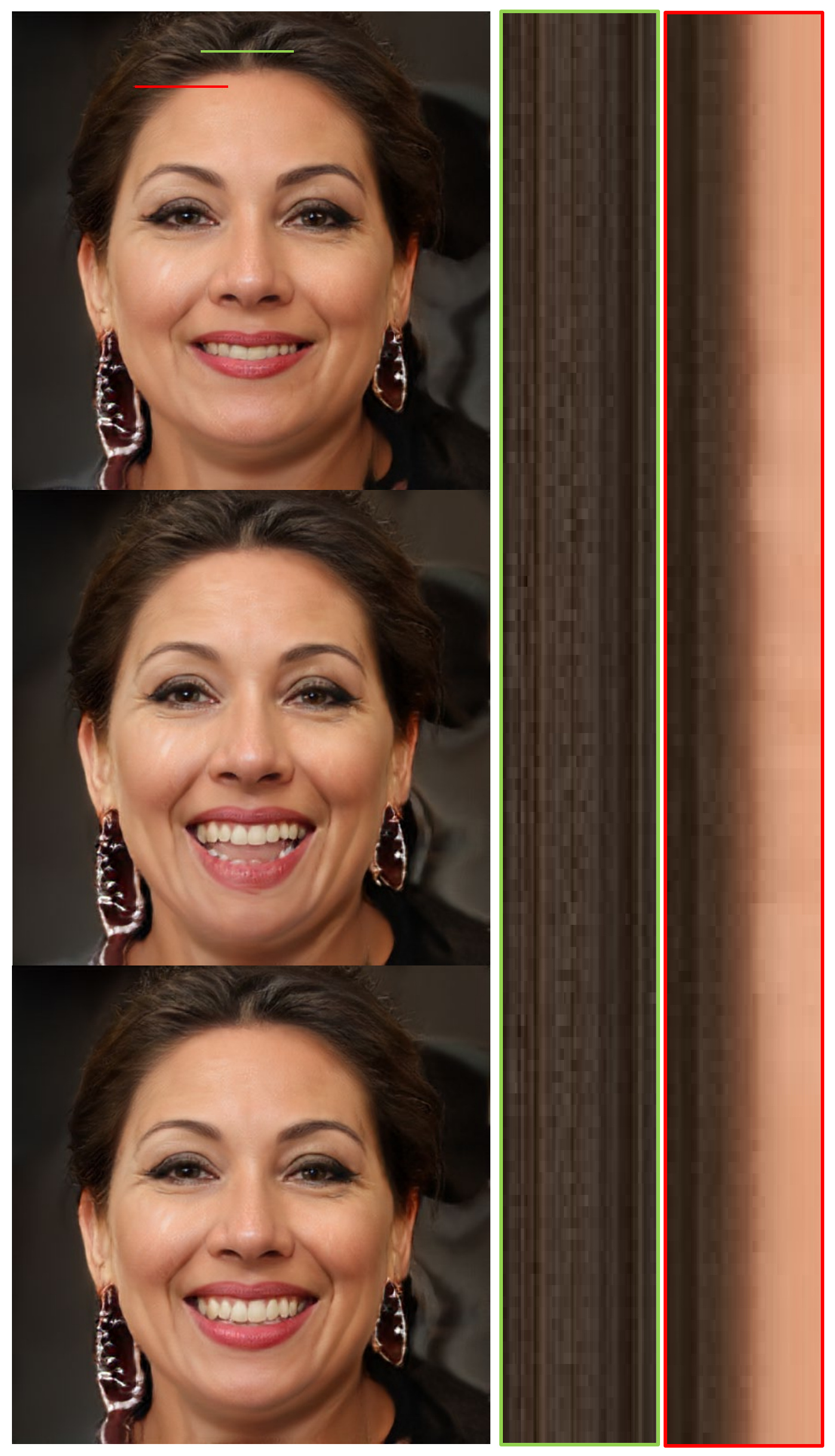}\\ 
    \rule{0pt}{2ex} 
    DiscoFaceGAN & PIRenderer & GAN-Control & HeadNeRF & Ours & Ours + Blending
\end{tabular}
\caption{\textbf{Visualization of temporal consistency on hair (green) and hairline (red) area during continuous expression animation.} We visualize the texture along a straight line during animation. The temporally consistent results should show smooth pattern in the visualization.}
\label{fig:cons_exp}
\end{figure*}

\subsubsection{Multi-view Consistency and Temporal Consistency}

Following~\cite{gram}, we visualize the consistency when varying expressions and poses using different methods. Figure~\ref{fig:cons_pose} shows the results when modifying the face poses or camera positions with other face attributes fixed. We check the texture of straight lines on the teeth and hair area.  Figure~\ref{fig:cons_pose} shows that DiscoFaceGAN, PIRenderer and GAN-Control suffer from inconsistency when smoothly changing the camera location, while HeadNeRF leads to many blurry results.  In comparison, the proposed approach produces 3D-consistent images of high fidelity.

We also vary expression continuously and the results are shown in Figure~\ref{fig:cons_exp}. For each frame, we illustrate the texture along the depicted lines to show the temporal consistency. GAN-Control clearly produces noise and distortion patterns. DiscoFaceGAN and our method produce more consistent results but still suffer tiny distortion. The volume blending, in comparison, further improves the temporal consistency.

\subsection{Quantitative Comparison with Prior Methods}

\subsubsection{Image Quality}
We measure the image quality with Frechet Inception Distance (FID) ~\cite{fid} using the ImageNet-pretrained Inception-V3 model~\cite{inception} and the vision-language  pretrained CLIP model ~\cite{clip}, and denote the latter score as ``FID-clip''.
We compute the FID between the 5,000 real images and 5,000 generated images. Since StyleRig~\cite{stylerig} and PIE~\cite{pie} do not release codes, we use images containing 168 identities with diverse expressions showcased on the project website\footnote{\url{https://vcai.mpi-inf.mpg.de/projects/PIE/}} for comparison.
When comparing DiscoFaceGAN, we run its model pretrained on FFHQ and generate 5,000 images with random poses, expressions and illuminations. Since this work is trained on $256\times 256$ resolution, we utilize a state-of-the-art super-resolution method, SwinIR~\cite{liang2021swinir}, to upsample their results to $512\times 512$ resolution for a fair comparison. PIRenderer uses  5,000 frontal face images produced by our method as inputs, and performs the face reenactment using randomly sampled expressions and poses.
We finetune the released model of HeadNeRF and generate 5,000 images with randomly sampled  3DMM coefficients. 

We align images using the same preprocessing procedure as Sec.~\ref{sec:preprocess} for all the compared methods.  As shown in Table~\ref{tab:fid}, our method significantly outperforms StyleRig, PIE, DiscofaceGAN, PIRenderer and HeadNeRF in terms of both FID and FID-clip. While GAN-Control achieves a better FID score, its disentanglement ability is much worse than our work as shown in the qualitative comparison and the following quantitative evaluations.

\begin{table}[t]
\centering
\caption{\textbf{Quantitative comparison of image quality.} The star symbol ($^{*}$) denotes we only use available data for evaluation. ``UP'' denotes the bicubic upsampling.  ``SR'' denotes the super-resolution results using~\protect\cite{liang2021swinir}. We highlight the best score and underline the second best.}
\begin{tabular}{l|cc}
\toprule
 & FID (512)$\downarrow$ & FID-clip (512)$\downarrow$ \\ \midrule
StyleRig$^{*}$\protect~\cite{stylerig} & 60.3 & 32.2 \\
PIE$^{*}$\protect~\cite{pie} & 60.9 & 24.4 \\
DiscoFaceGAN-UP\protect~\cite{discofacegan} & 56.6 & 18.6 \\
DiscoFaceGAN-SR\protect~\cite{discofacegan} & 39.1 & 17.3 \\
PIRenderer-UP\protect~\cite{pirenderer} & 77.4 & 23.8 \\ PIRenderer-SR\protect~\cite{pirenderer} & 68.9 & 24.0 \\ 
GAN-Control\protect~\cite{gan-control} & \textbf{13.5} & \textbf{8.5}\\ \midrule
HeadNeRF\protect~\cite{HeadNeRF} & 142.6 & 64.7 \\
Ours & \underline{24.1} & \underline{14.7} \\
\bottomrule
\end{tabular}
\label{tab:fid}
\end{table}

\subsubsection{Control Accuracy}
We evaluate the control accuracy of all methods using Average Expression Distance (AED), Average Pose Distance (APD) and Average Illumination Distance (AID), which have been used in \cite{pirenderer}. For StyleRig and PIE, we also use the animation results from their project webpages.
For HeadNeRF, DiscoFaceGAN, GAN-Control and our method, we randomly generate 1,000 identities with neutral expression and randomly apply 10 expressions, poses and illuminations. For PIRenderer, we use the same source images and driving coefficients as our method. Our method gets 10,000 generated images, for which we use the face reconstruction network to reconstruct 3DMM coefficients.
Finally, we calculate the average distance between the input control parameters and the reconstructed coefficients. Table~\ref{tab:aed} shows that our method achieves the best AED, APD and AID scores.

\begin{table}[t]
\centering
\caption{\textbf{Quantitative comparison of the control accuracy.} For PIRenderer we only compute AED and APD, since it cannot control illumination. We highlight the best score and underline the second best.}
\begin{tabular}{l|ccc}
\toprule
 & AED$\downarrow$ & APD$\downarrow$ & AID$\downarrow$ \\ \midrule
DiscoFaceGAN\protect~\cite{discofacegan} &\underline{0.2207} & 0.0016 & 0.0295\\
PIRenderer\protect~\cite{pirenderer} & 0.2360 & 0.0030 & -\\ 
GAN-Control\protect~\cite{gan-control} & 0.2221 & 0.0022 & 0.0217\\\midrule
HeadNeRF\protect~\cite{HeadNeRF} & 0.2732 & \underline{0.0015} & \underline{0.0129} \\
Ours & \textbf{0.2026} & \textbf{0.0010} & \textbf{0.0018}\\
\bottomrule
\end{tabular}

\label{tab:aed}
\end{table}

\begin{table}[t]
\centering
\caption{\textbf{Quantitative comparison of disentanglement.} We highlight the best score and underline the second best. $\mathrm{DS}$ denotes the disentanglement score and $\mathrm{IS}$ denotes the identity similarity score.}
\begin{tabular}{l|cccc}
\toprule
& $\mathrm{DS}_{\beta}\uparrow$ & $\mathrm{DS}_{r}\uparrow$ & $\mathrm{DS}_{\gamma}\uparrow$ & $\mathrm{IS}\uparrow$\\ \midrule
StyleRig$^{*}$\protect~\cite{stylerig} & 24.3 & 18.4 & 3.0 & 0.66\\
PIE$^{*}$\protect~\cite{pie} & 27.4 & 31.3 & 2.0 & 0.74\\
DiscoFaceGAN\protect~\cite{discofacegan} & 32.5 & 77.3 & 18.2 & 0.60\\
PIRenderer\protect~\cite{pirenderer}& 7.83  & 65.9 & -  & 0.27\\ 
GAN-Control\protect~\cite{gan-control}& \underline{37.6} & 67.5 & \underline{25.3} & 0.72\\ \midrule
HeadNeRF\protect~\cite{HeadNeRF} & 37.0 & \underline{82.6} & 3.68 & \underline{0.80} \\
Ours & \textbf{46.1} & \textbf{117.5} & \textbf{33.2} & \textbf{0.87}\\
\bottomrule
\end{tabular}

\label{tab:ds}
\end{table}

\subsubsection{Disentanglement} 
We evaluate the disentanglement using the Disentanglement Score (DS), which is first proposed in~\cite{discofacegan}. We denote the DS for  expression, pose and illumination as DS$_{\beta}$, DS$_{r}$, DS$_{\gamma}$ respectively. 
Specifically, we randomly sample images with a single modified attribute.
Then, we re-estimate the 3DMM parameters from the generated images and calculate the variance of the estimated coefficients $(\bm{\beta}, \bm{R}, \bm{\gamma})$.
To discount the influence of coefficient magnitude, we normalize the score according to the coefficient variance computed from the FFHQ dataset. Thus, $\mathrm{DS}_{i}$ is calculated as:

\begin{equation}
\begin{aligned}
    \mathrm{DS}_i = \prod_{\forall j\neq i}\frac{\sigma_{i}}{\sigma_{j}}, 
    \sigma_i=\mathrm{var}(i), 
    i,j\in\{\bm{\beta}, \bm{R}, \bm{\gamma}\}.
\end{aligned}
\end{equation}
A higher $\mathrm{DS}$ indicates that other attributes will remain the same when editing one specified attribute. Besides, we also measure identity similarity during portrait editing by calculating the cosine similarity of the face embedding using a  pretrained face recognition~\cite{deng2019arcface}. We randomly generate 1,000 source images and randomly apply two poses, expressions or illuminations to the model.
Numerical results in Table~\ref{tab:ds} show that our method achieves the best disentanglement ability in terms of all the evaluation metrics.



%

\subsection{Ablation Study}

In this section, we perform extensive ablation studies of the proposed technical components to investigate their effects on the final output.

\subsubsection{The Effect of VAE-GANs and Auxiliary Expression Data}
\label{sec:ab_cm}
\begin{figure}[]
    \centering
    \includegraphics[width=\linewidth]{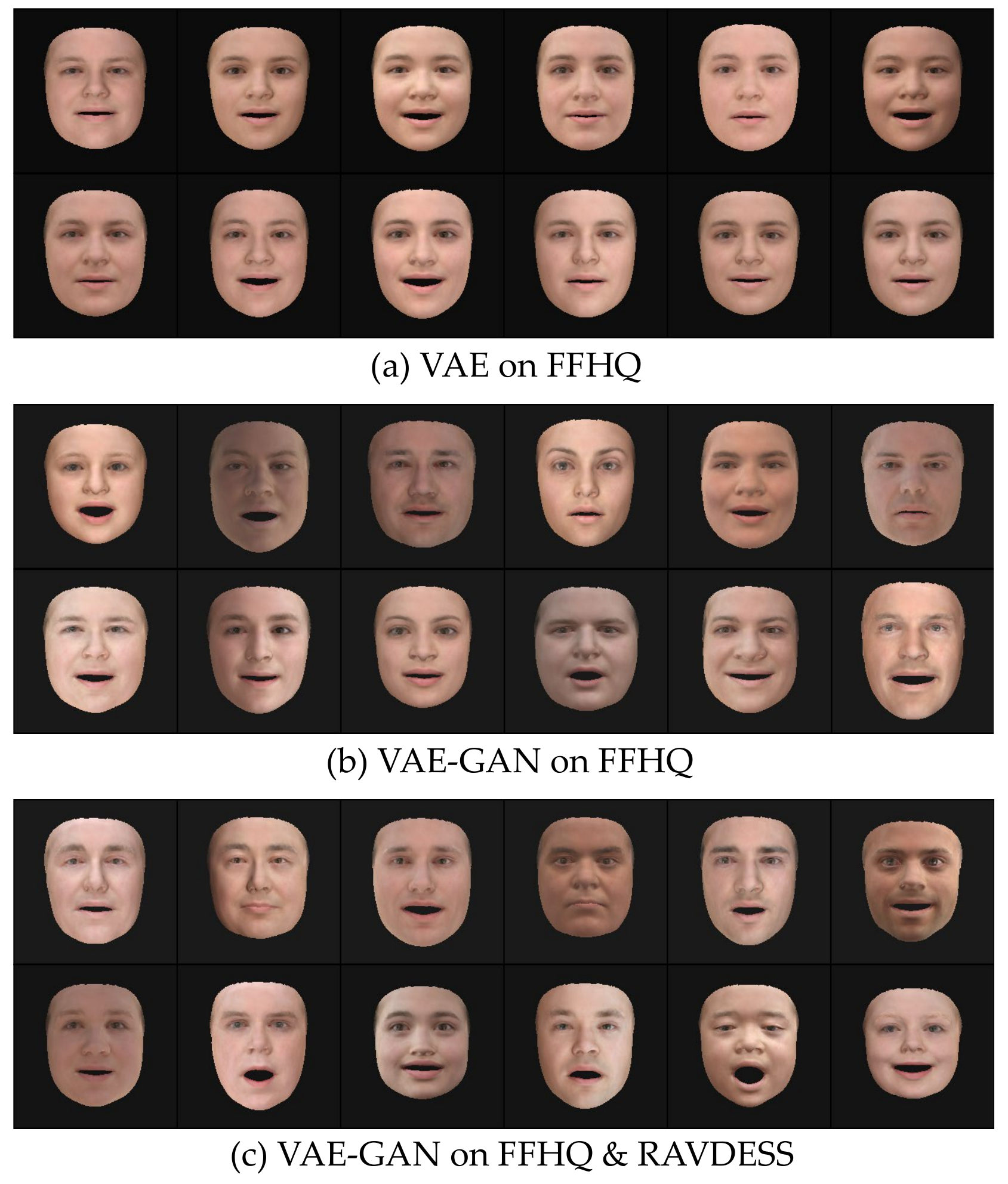}
    \caption{\textbf{Ablation study of learning VAE-GANs.} (a) VAE tends to give mean expressions. (b) VAE-GAN improves expression diversity. (c) The supplement with RAVDESS training images further helps to sample more vivid expressions. }
    \label{fig:vae}
\end{figure}

We compare the vanilla VAE model trained on FFHQ, the VAE-GAN trained on FFHQ and the VAE-GAN model trained on combined FFHQ and RAVDESS  datasets. Figure~\ref{fig:vae} shows that the vanilla VAE tends to give averaged faces with dull expressions. Adding adversarial training (VAE-GAN) improves the expression diversity, yet the expression diversity is still limited due to the data distribution of FFHQ. To address this issue, we add expression coefficients from RAVDESS, which helps to sample diverse and vivid expressions. Thus we can sample expressions like angry, surprised or fearful faces with delicate eyebrow movements.

\begin{table}[t]
\centering
\caption{\textbf{The influence of proposed components to image quality and controllability.} }
\begin{tabular}{l|ccccc}
\toprule
 & FID$\downarrow$ & FID-clip$\downarrow$ & AED$\downarrow$ & APD$\downarrow$ & AID$\downarrow$ \\ \midrule
Ours & 24.1 & \underline{14.7} & \textbf{0.2026} & \textbf{0.0010} & \textbf{0.0018} \\\midrule
w/o SF & \textbf{19.0} & \textbf{10.8} & 0.2203 & \underline{0.0013} & 0.0035\\
w/o align & 38.3 & 37.2&  0.2798 & 0.0049 & 0.0027\\
w/o $\mathcal{L}_{id}$ & 28.0 & 21.4 & 0.2146 & 0.0036 & 0.0029\\
w/o $\mathcal{L}_{tex}$ & 24.2 & 17.4 & \underline{0.2102} & 0.0058 & 0.0021\\
w/o $\mathcal{L}_{lm}$ & 41.6 & 40.0 & 0.2730 & 0.0211 & 0.0029\\\midrule
{\begin{tabular}[c]{@{}c@{}} $-\mathcal{L}_{ce}$ \\ $+\mathcal{D}_{sf}$ \end{tabular}} 
& \underline{21.0} & 16.3 & 0.2179 & 0.0026 & \underline{0.0019} \\
\bottomrule
\end{tabular}
\label{tab:ab_fid}
\end{table}

\begin{table}[t]
\centering
\caption{\textbf{The influence of proposed components to disentanglement.}}
\begin{tabular}{l|cccc}
\toprule
 & DS$_{e}\uparrow$ & DS$_{p}\uparrow$ & DS$_{il}\uparrow$ & IS$\uparrow$ \\ \midrule
Ours & \textbf{46.1} & \textbf{117.5} & \textbf{33.2} & \textbf{0.87} \\\midrule
w/o SF & 42.4 & 101.7 & 13.2 & \underline{0.85} \\
w/o $\mathcal{L}_{\textnormal{dis}}$ & 21.6 & 105.2 & 11.3 & 0.54 \\
w/o $\mathcal{L}_{\textnormal{dis}}^{\kappa}$ & 44.9 & 110.3 & 25.8 & 0.82 \\
w/o $\mathcal{L}_{\textnormal{dis}}^{\beta}$ & 30.5 & 108.5 & \underline{29.8} & 0.68\\
w/o $\mathcal{L}_{\textnormal{dis}}^{\gamma}$ & 39.4 & \underline{111.3} & 15.6 & 0.80\\
w/o blend & \underline{45.3} & \textbf{117.5} & \textbf{33.2} & \underline{0.85} \\
\bottomrule
\end{tabular}
\label{tab:ab_ds}
\end{table}

\begin{figure*}
    \centering
    \setlength\tabcolsep{0pt}
    \footnotesize
    \renewcommand{\arraystretch}{0}
    \begin{tabular}{cccccccccc}
    \footnotesize &
    \includegraphics[width=0.1\linewidth]{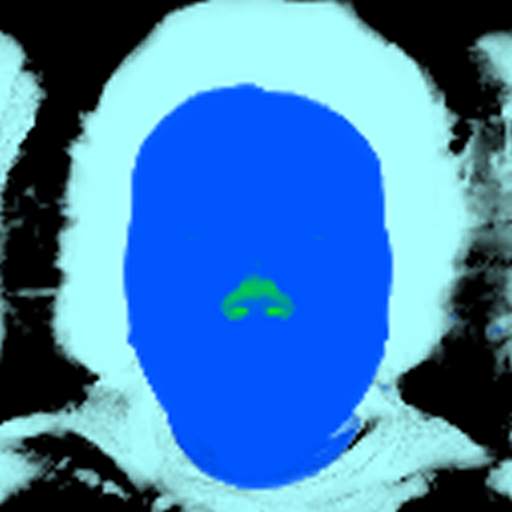} & \includegraphics[width=0.1\linewidth]{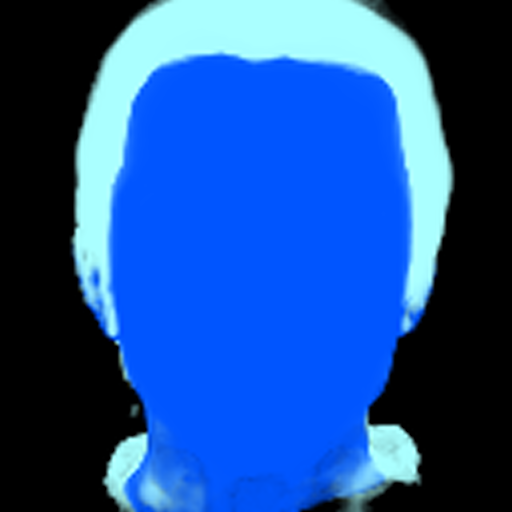} & \includegraphics[width=0.1\linewidth]{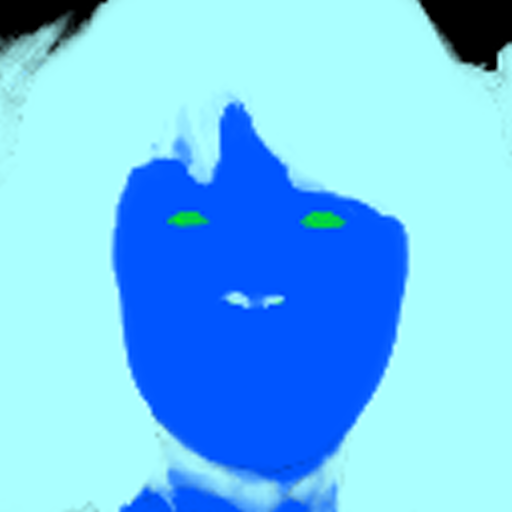} & \includegraphics[width=0.1\linewidth]{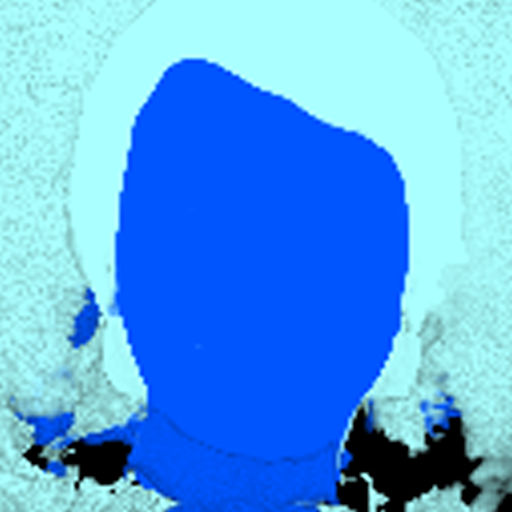} &  \includegraphics[width=0.1\linewidth]{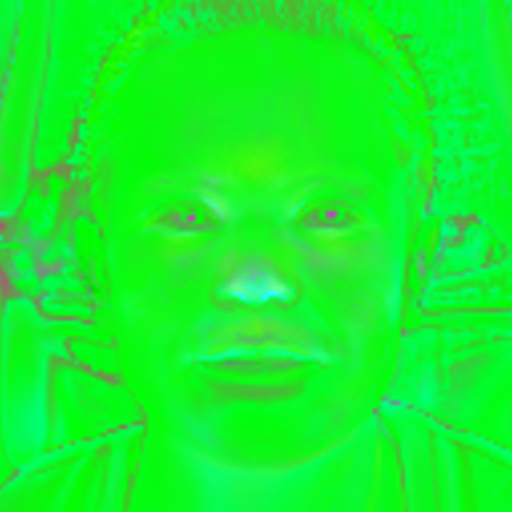} &  \includegraphics[width=0.1\linewidth]{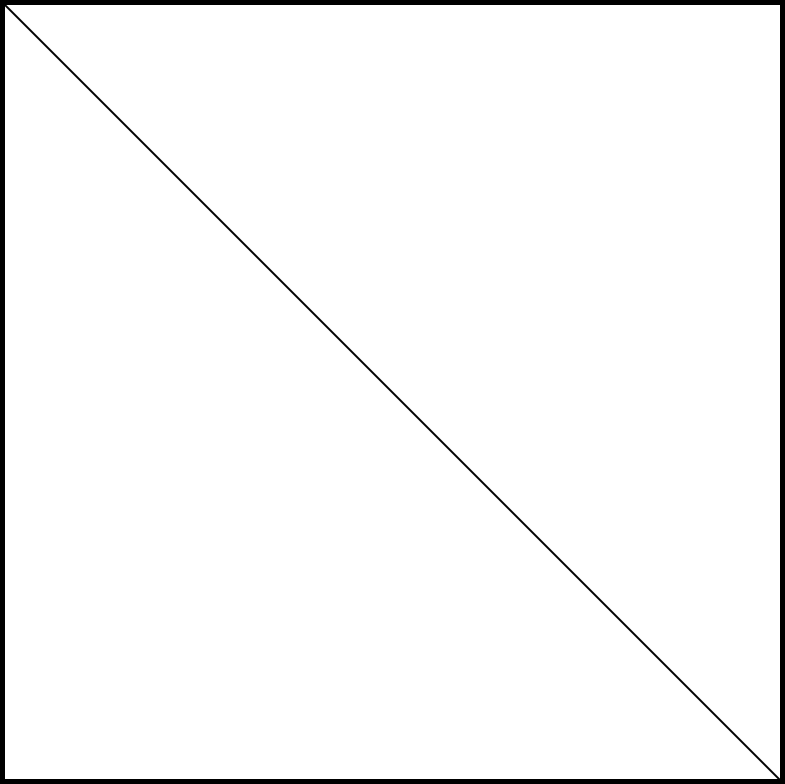} &  \includegraphics[width=0.1\linewidth]{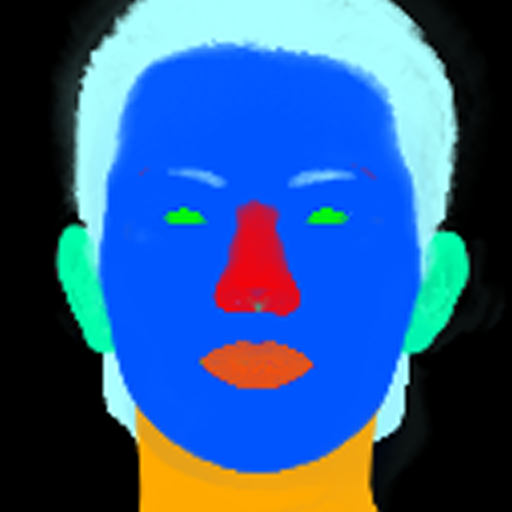} &\includegraphics[width=0.1\linewidth]{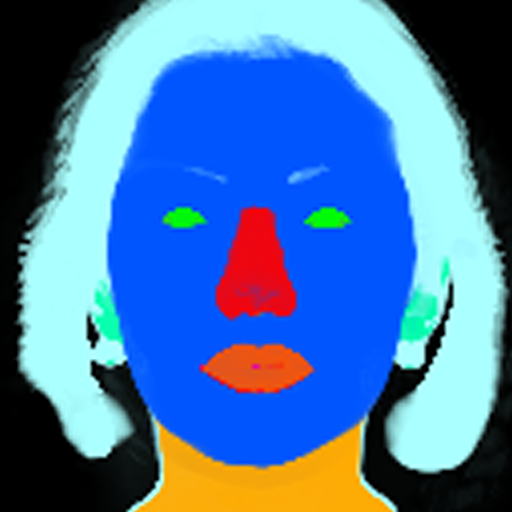} &  \includegraphics[width=0.1\linewidth]{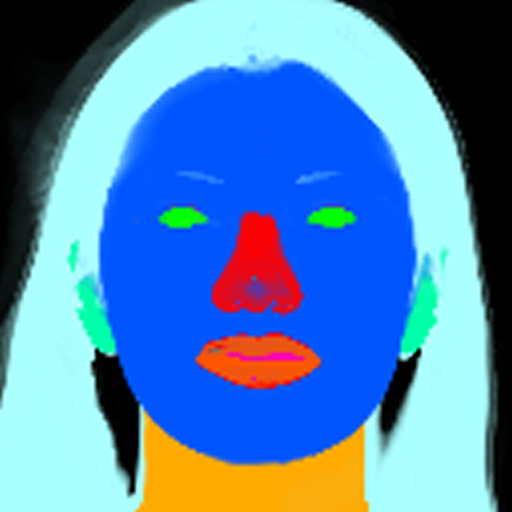} \\ 
    \includegraphics[width=0.1\linewidth]{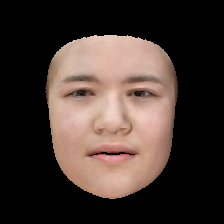} &
    \includegraphics[width=0.1\linewidth]{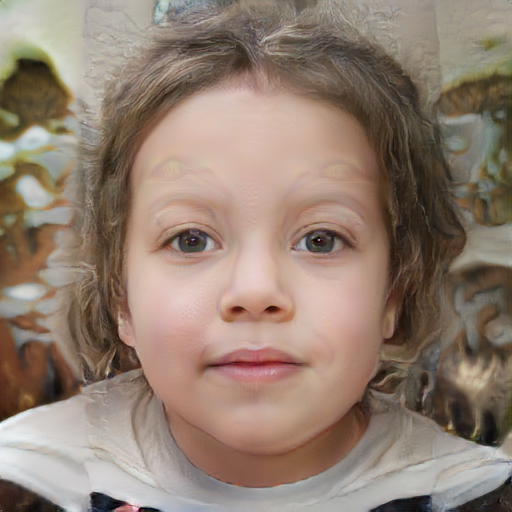} & \includegraphics[width=0.1\linewidth]{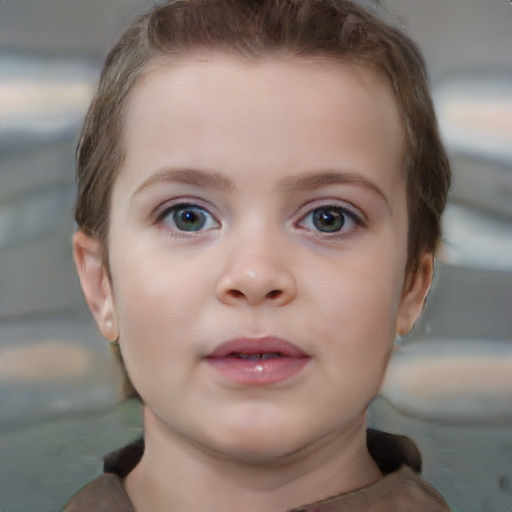} & \includegraphics[width=0.1\linewidth]{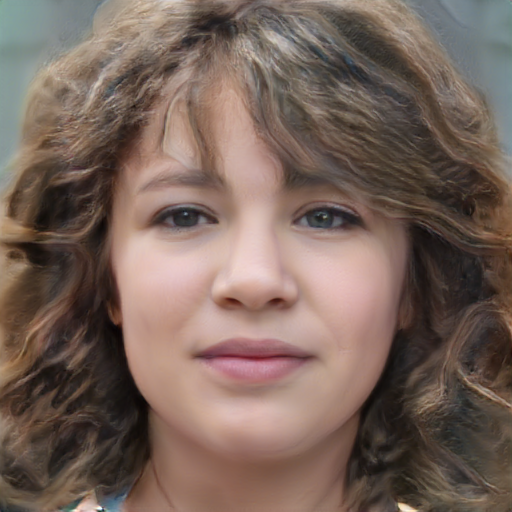} & \includegraphics[width=0.1\linewidth]{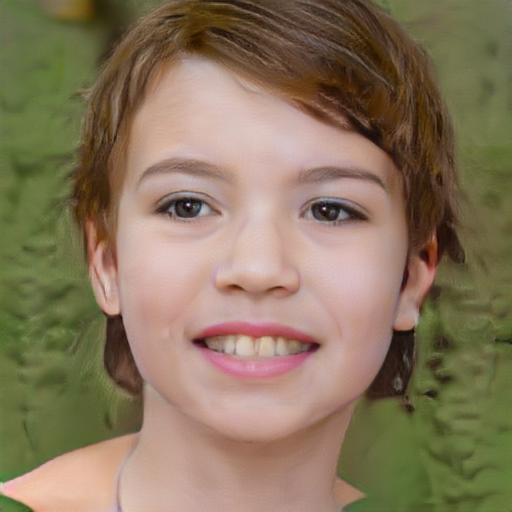} &  \includegraphics[width=0.1\linewidth]{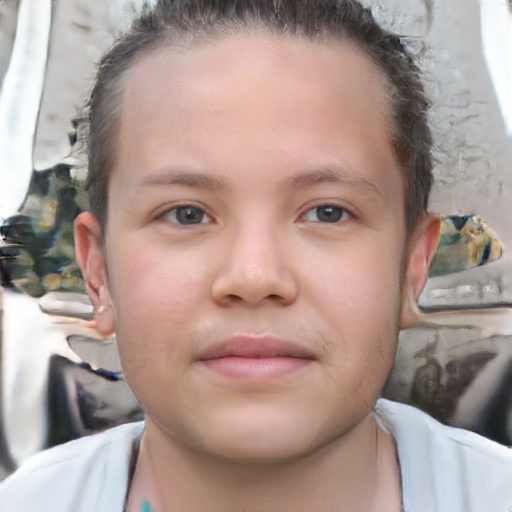} &  \includegraphics[width=0.1\linewidth]{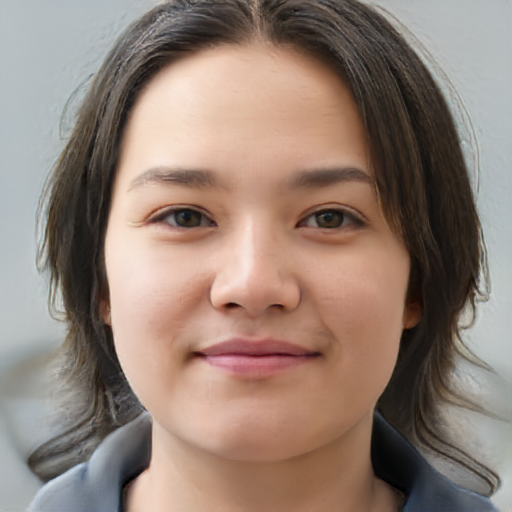} &  \includegraphics[width=0.1\linewidth]{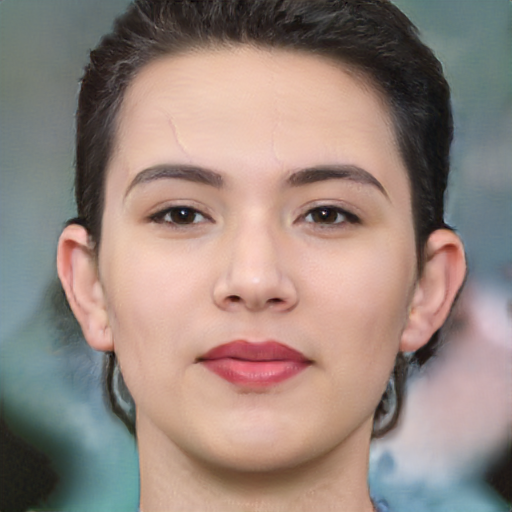} & \includegraphics[width=0.1\linewidth]{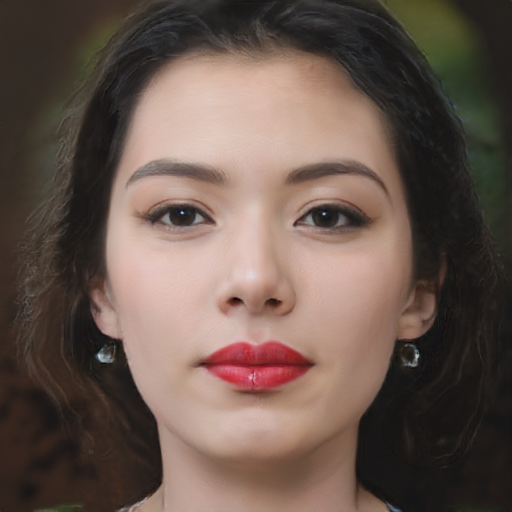} &  \includegraphics[width=0.1\linewidth]{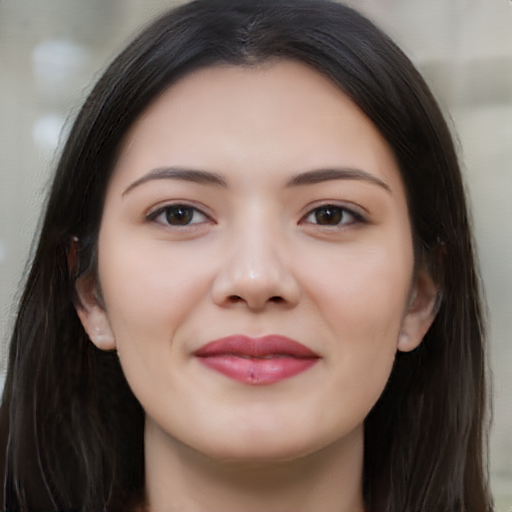} \\ 
    \includegraphics[width=0.1\linewidth]{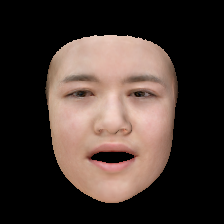} &
    \includegraphics[width=0.1\linewidth]{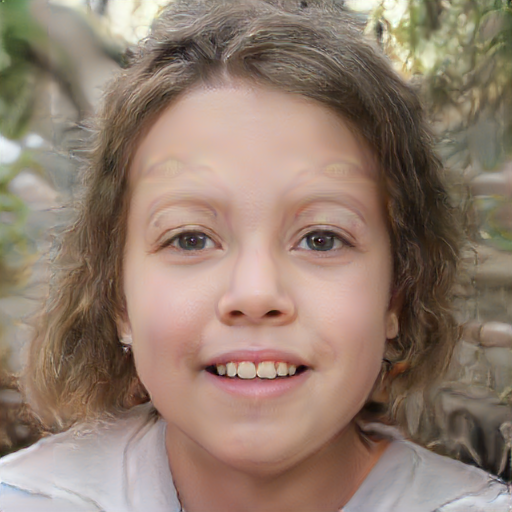} & \includegraphics[width=0.1\linewidth]{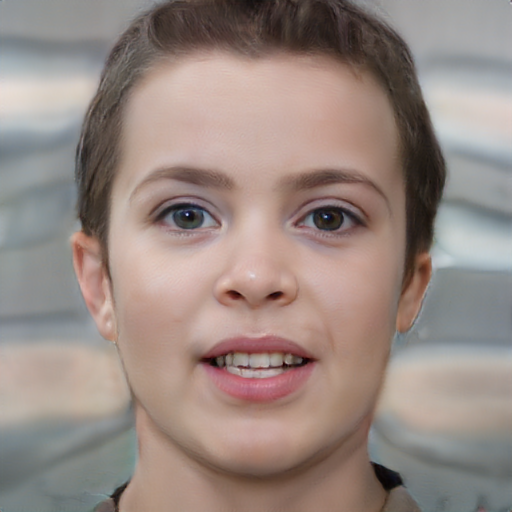} & \includegraphics[width=0.1\linewidth]{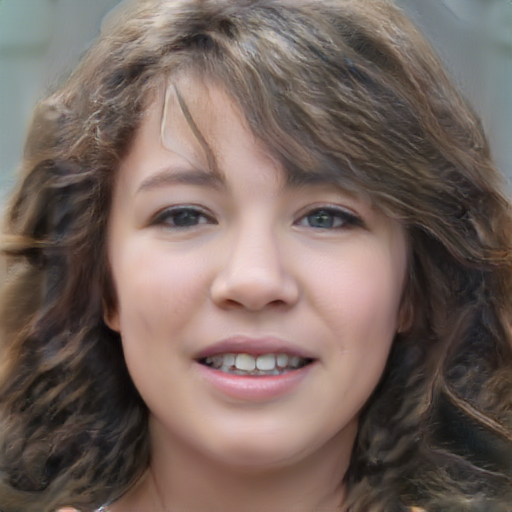} & \includegraphics[width=0.1\linewidth]{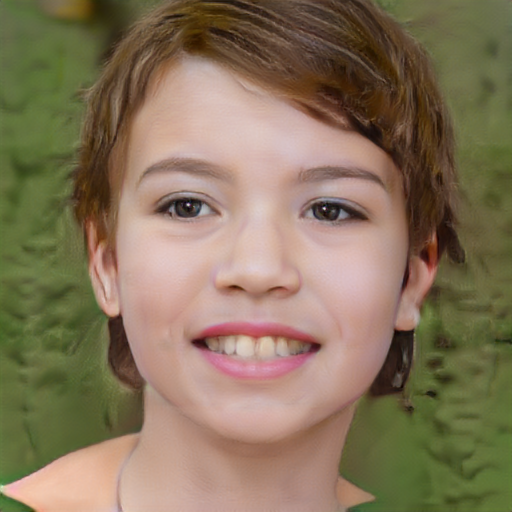} &  \includegraphics[width=0.1\linewidth]{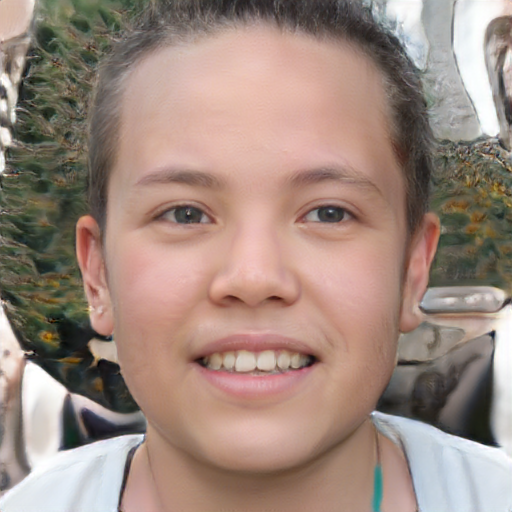} &  \includegraphics[width=0.1\linewidth]{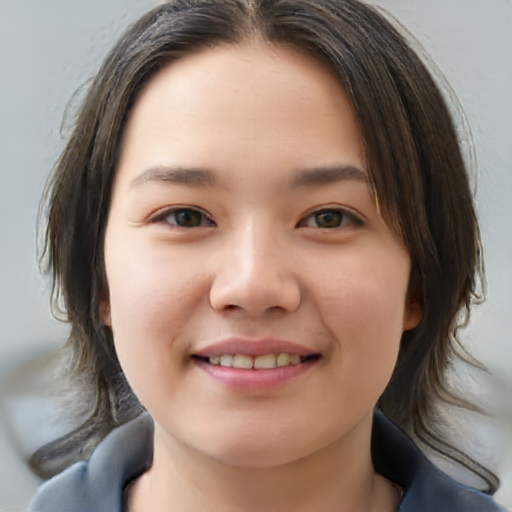} &  \includegraphics[width=0.1\linewidth]{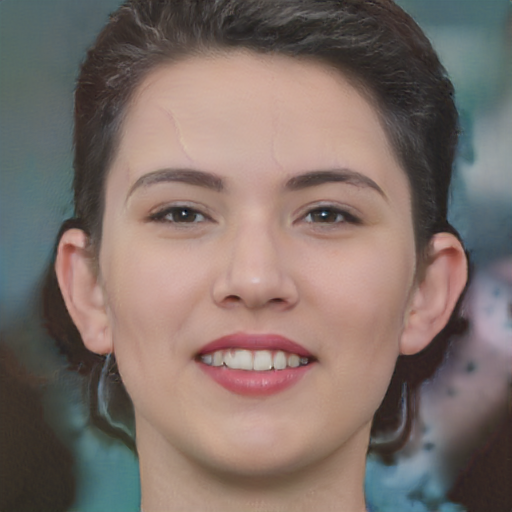} & \includegraphics[width=0.1\linewidth]{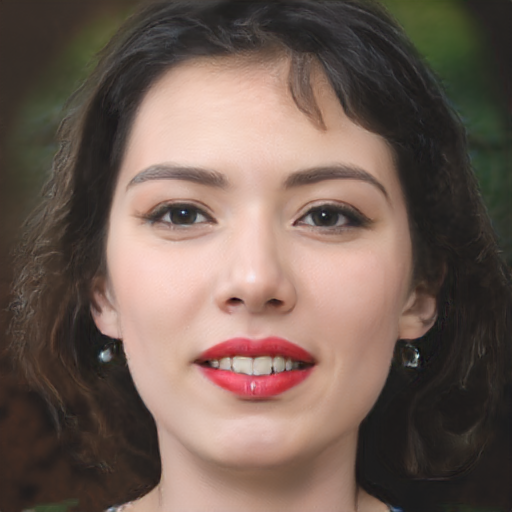} &  \includegraphics[width=0.1\linewidth]{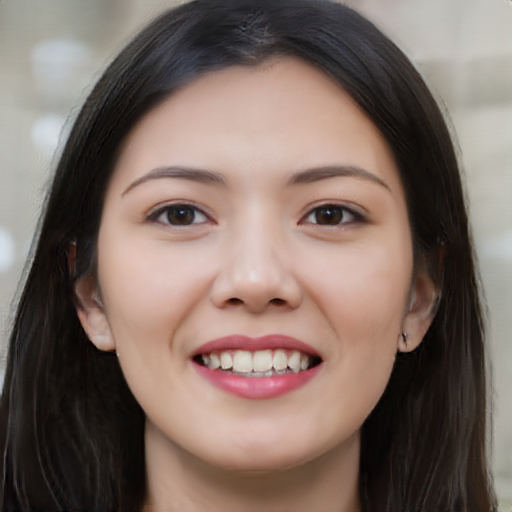} \\ 
    \rule{0pt}{2ex} 
    Guided & w/o & w/o & w/o & w/o & $-\mathcal{L}_{\textnormal{ce}}$ & w/o & w/o & w/o & Full \\ \rule{0pt}{2ex} 3DMM & align & $\mathcal{L}_{\textnormal{id}}$ & $\mathcal{L}_{\textnormal{tex}}$ & $\mathcal{L}_{\textnormal{lm}}$ & $+\mathcal{D}_{sf}$ & SF & $\mathcal{L}_{\textnormal{dis}}$ & $\mathcal{L}^{\beta}_{\textnormal{dis}}$ & model 
\end{tabular}
    \caption{\textbf{Ablation studies of loss functions.} Given two guided 3DMM coefficients, we visualize the generated image as well as  face parsing results from the models in different settings. }
    \label{fig:ab_loss}
\end{figure*}

\subsubsection{The Effect of Face Alignment}
\label{sec:ab_preprocess}
We further show the necessity of data preprocessing as mentioned in Sec.~\ref{sec:preprocess}. The experimental results shown in Table~\ref{tab:ab_fid} and Figure~\ref{fig:ab_loss} prove that the alignment in data preprocesses significantly improves the performance of the 3DMM guidance. Without this preprocessing, the generated neural radiance fields fail to align with the 3DMM guidance, leading to obvious artifacts on the eyebrows.

\subsubsection{The Effect of Loss Functions}
We conduct ablation studies to validate the benefits of different loss functions used in imitation learning and disentanglement learning, respectively. Table~\ref{tab:ab_fid} and Table~\ref{tab:ab_ds} show the quantitative ablation studies of different loss functions. Figure~\ref{fig:ab_loss} illustrates how they affect image quality and controllability. These results show that without the identity loss, the identity similarity between 3DMM renderings and the generated images drops and image quality also deteriorates. Removing the landmark loss greatly impairs the generation quality and the control accuracy. The ablation of losses in the disentanglement learning, \ie, $\mathcal{L}_{\textnormal{dis}}^{\kappa}$, $\mathcal{L}_{\textnormal{dis}}^{\beta}$, and $\mathcal{L}_{\textnormal{dis}}^{\gamma}$, degrades the control disentanglement and causes inconsistency for other facial attributes.



\begin{figure}
\centering
\setlength\tabcolsep{0pt}
\footnotesize
\renewcommand{\arraystretch}{0}
\begin{tabular}{ccccc}
   \footnotesize
    \includegraphics[width=0.2\linewidth]{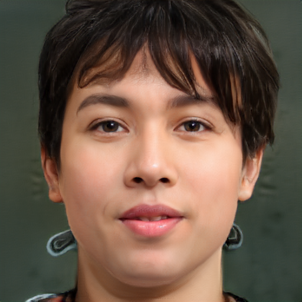} &
    \includegraphics[width=0.2\linewidth]{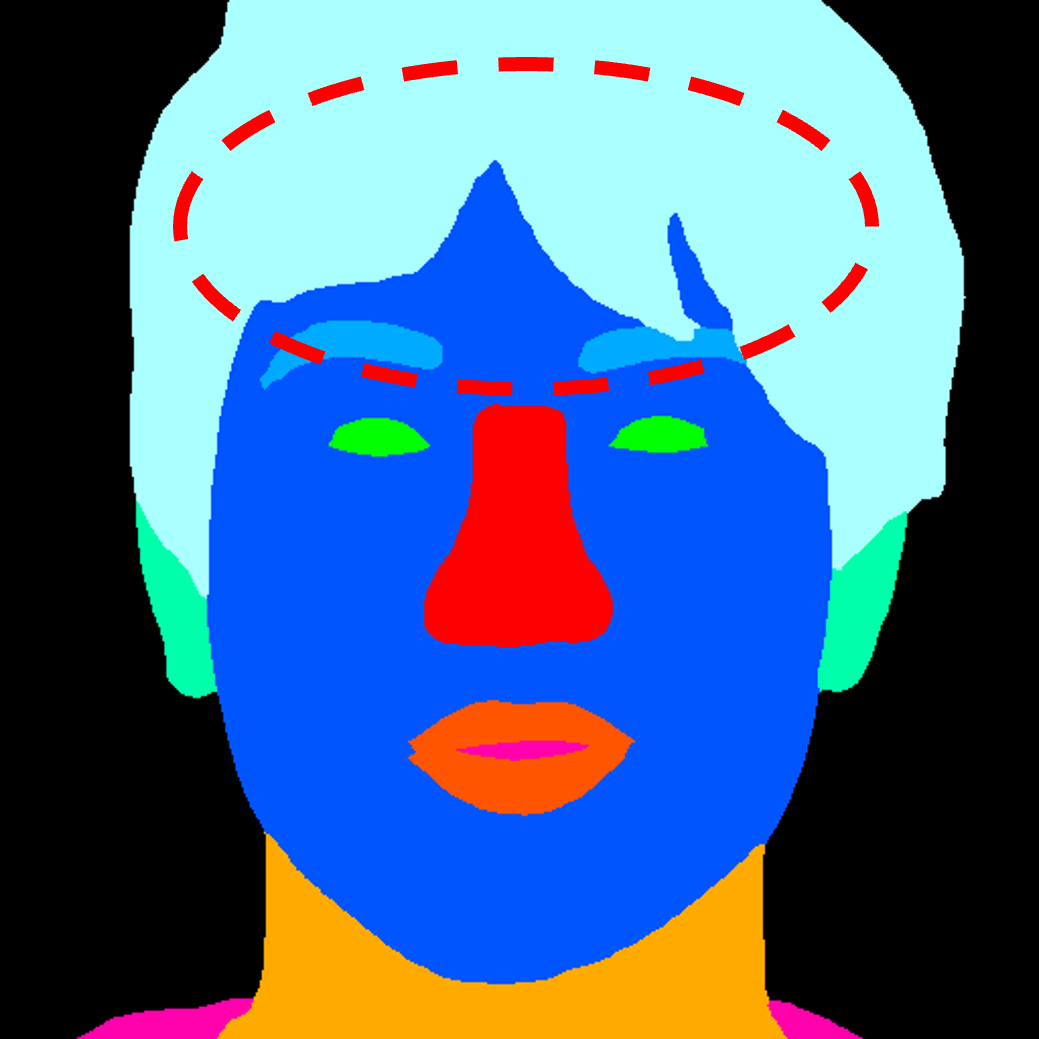}  & 
    \includegraphics[width=0.2\linewidth]{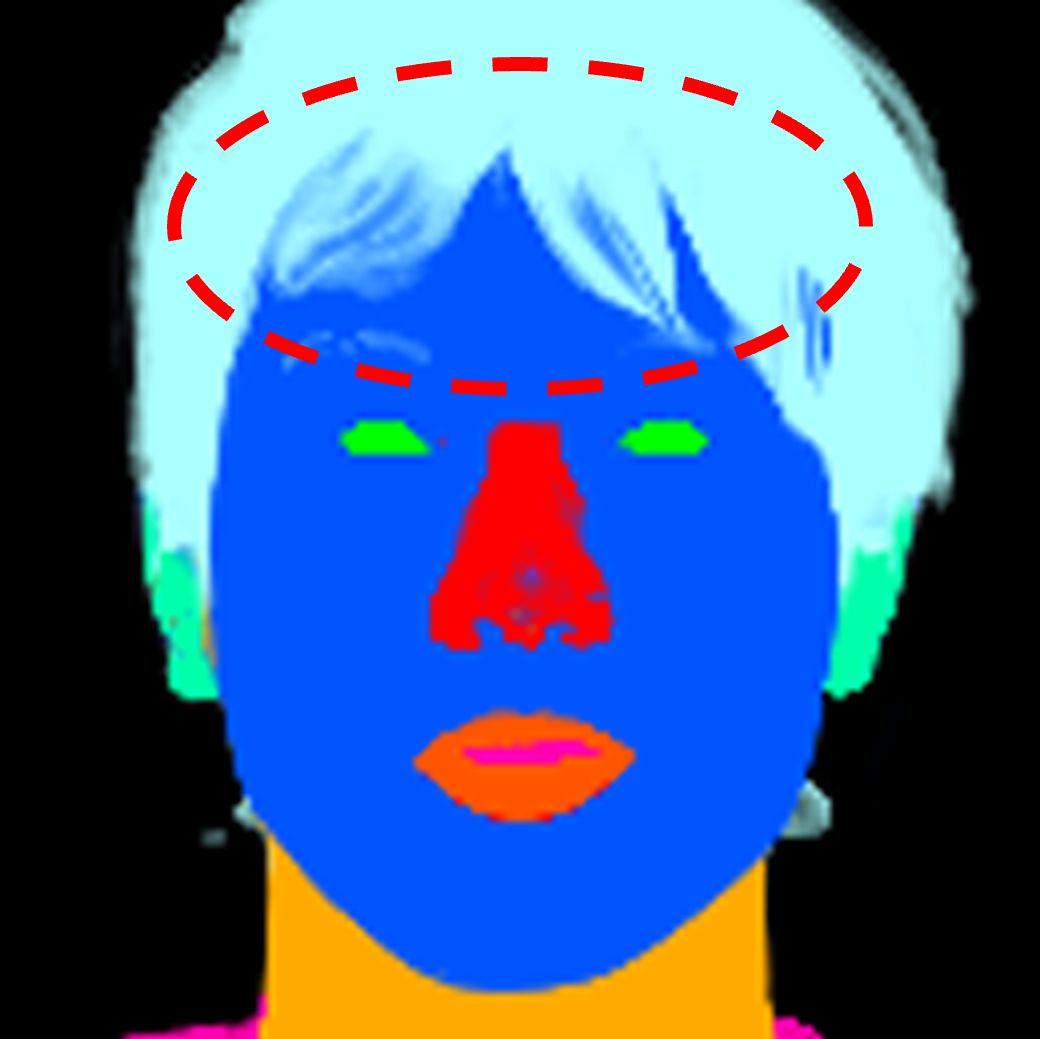}  & 
    \includegraphics[width=0.2\linewidth]{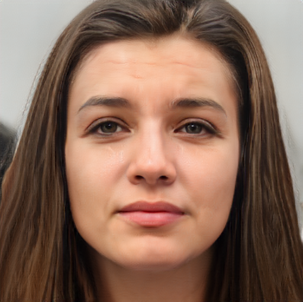} &
   \includegraphics[width=0.2\linewidth]{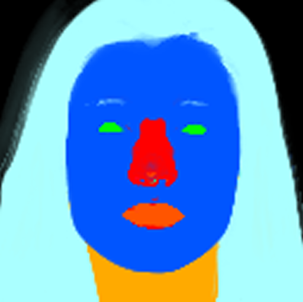} \\
    \includegraphics[width=0.2\linewidth]{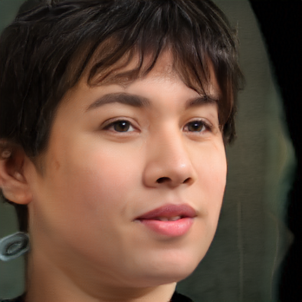} &
    \includegraphics[width=0.2\linewidth]{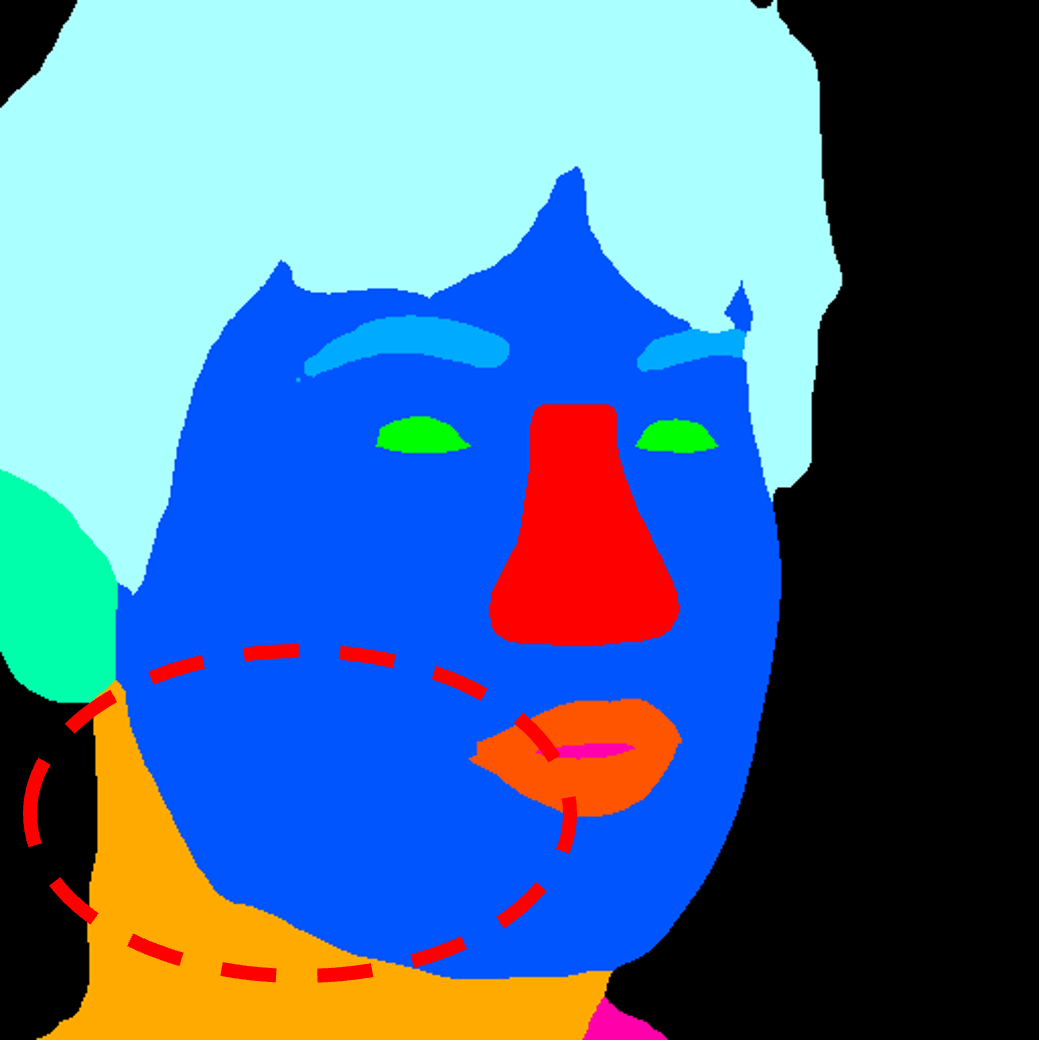}  & 
    \includegraphics[width=0.2\linewidth]{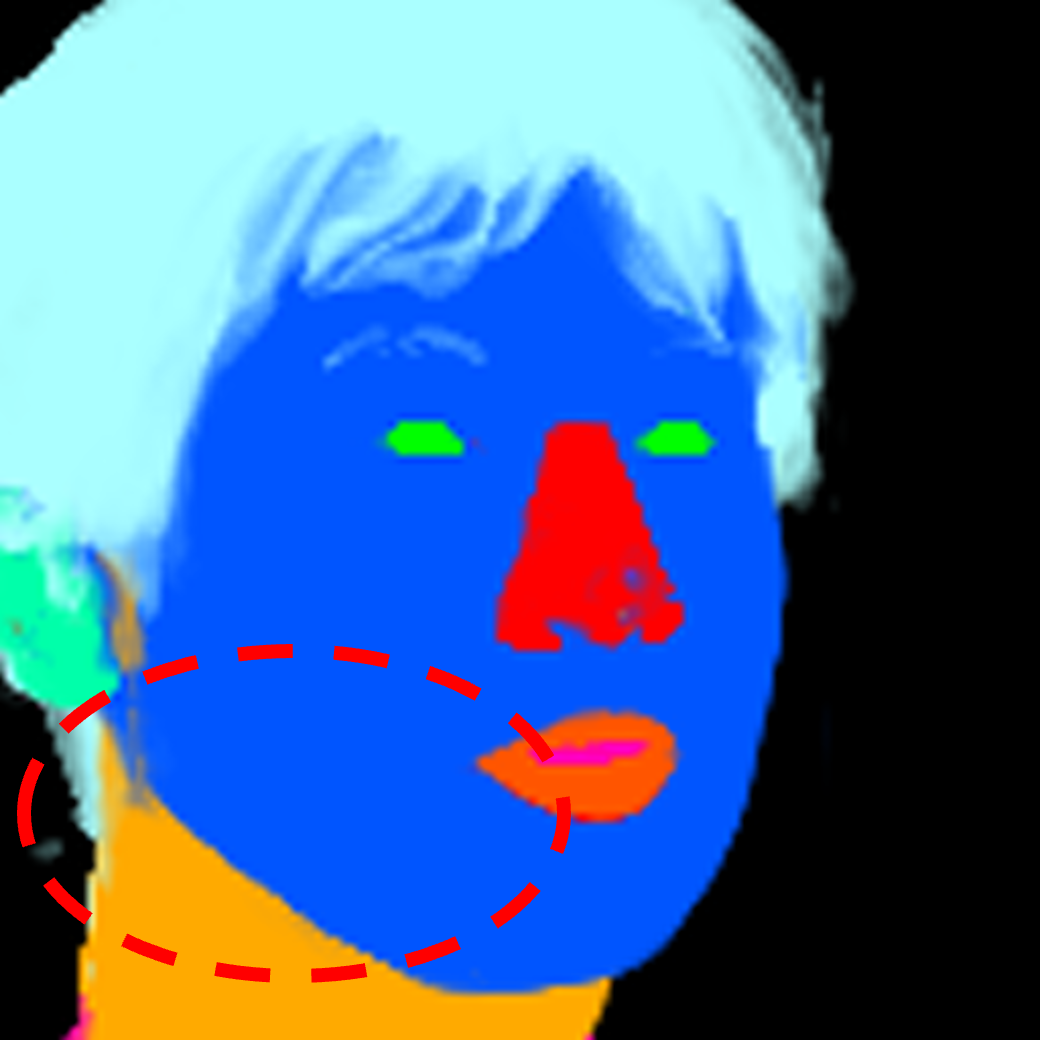}  & 
    \includegraphics[width=0.2\linewidth]{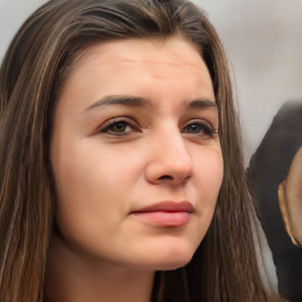} &
   \includegraphics[width=0.2\linewidth]{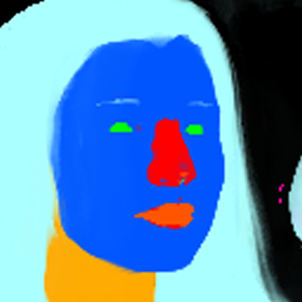} \\ 
    \rule{0pt}{2ex} 
    Synthetic & 2D Mask & Ours & \multicolumn{2}{c}{Other Results}   
\end{tabular}
\caption{\textbf{Visualization of 2D semantic mask and our rendered semantic mask.} Compared with a 2D semantic mask, the rendered semantic mask from the learned semantic field can be even more accurate than the 2D parsing results while being more view consistent. }
\label{fig:sf_com}
\end{figure}

\begin{figure*}
    \centering
    \includegraphics[width=\linewidth]{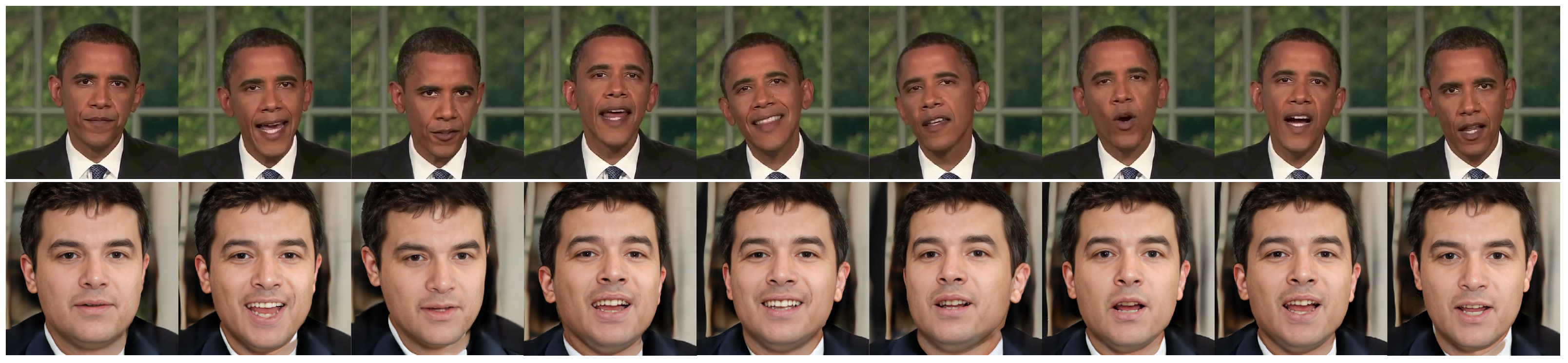}
    \caption{\textbf{Example results produced by our method on video-driven talking face generation}. Our method can generate photo-realistic portrait images (bottom) according to the drive motion (top).}
    \label{fig:drive}
\end{figure*}

\subsubsection{The Benefit of Learned Semantic Field}
While involving semantic learning may introduce additional burden to the generator and slightly worsens the image quality (see results in Table~\ref{tab:aed}), it significantly improves the control disentanglement and slightly benefits the control accuracy, as shown in Table~\ref{tab:aed} and Table~\ref{tab:ds}. This is because we can accurately parse the 3D portrait and let the generator focus on the face part for disentangled editing. Besides, the explicit volume blending also leverages the learned semantic field, and makes the non-face regions perfectly consistent.

Moreover, we compare the way to train a good semantic field. 
Rather than relying on the online parsing result as the ground truth,
we may introduce a semantic discriminator $\mathcal{D}_{sf}$ that examines the realism of rendered parsing result, which we denote as ``- $\mathcal{L}_{ce}+\mathcal{D}_{sf}$''
Figure~\ref{fig:ab_loss} shows that the discriminator cannot reliably provide the supervision for the semantic filed learning.

We also visualize more rendered results for the radiance field and the semantic field in Figure~\ref{fig:sf_com}. Both the image rendering and the face parsing result are view consistent. It is interesting to see that the learned semantic field can even more accurately parse the face than the 2D ground truth.

\begin{figure}[t]
    \centering
    \includegraphics[width=\linewidth]{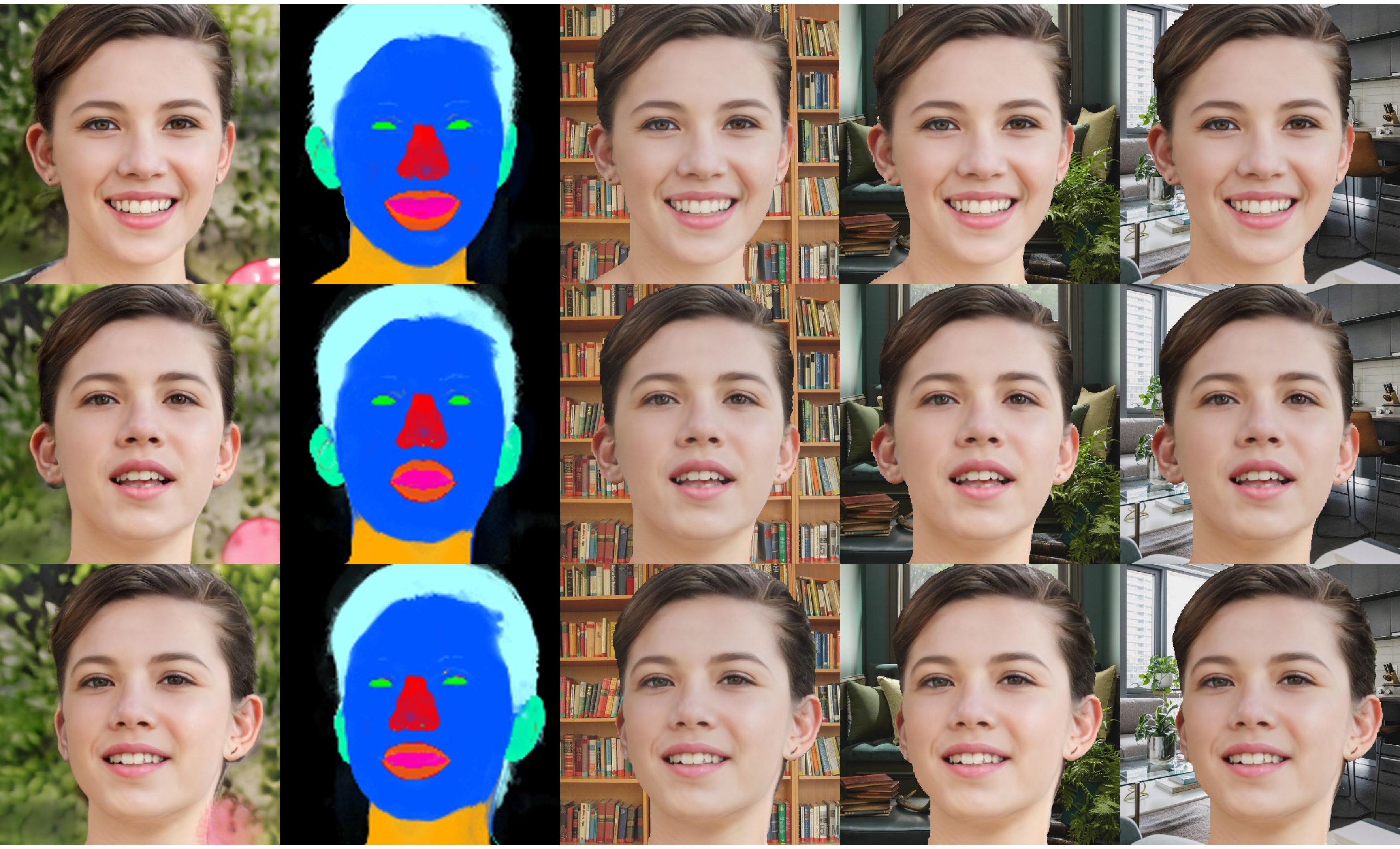}
    \caption{\textbf{Visualization of varying different background images.} From left to right, we show the generated portrait image using our method, rendered semantic mask, and images with diverse background images.}
    \label{fig:bg}
\end{figure}

\section{Applications}
\label{sec:application}
In this section, we showcase several applications of our method. Our method could enable video-driven talking head generation with the editable background. We can also achieve real portrait image editing by projecting the real image into the latent space of the network. We also investigate the out-of-domain generalization ability of our method on two stylized portrait datasets.

\subsection{Talking Face Generation}

Specifically, for an input talking face video, we can capture the expression coefficients and head pose information from a specific frame using a face reconstruction network~\cite{deep3dface}. Then we can explicitly feed each expression coefficient to our pipeline and render the image with the corresponding head pose frame by frame. Given a driving video, we can randomly generate a talking face video with the same expressions and head movements for different identities. 

We visualize the video-driven talking face results in Figure~\ref{fig:drive}.  Our method can capture expressions and mouth movements. 
We also show examples of background replacement results in Figure~\ref{fig:bg}.
To achieve this, we make all background volume transparent, and save the alpha value for each pixel to generate an alpha mask. We achieve background replacement by blending the opaque background image with a semi-transparent rendered image using the alpha mask. Thanks to the learned semantic field, we can achieve delicate portrait matting and high-quality background replacement effects.  

\begin{figure}[t]
    \centering
    \setlength\tabcolsep{0pt}
    \renewcommand{\arraystretch}{0}
    \footnotesize
    \begin{tabular}{ccccc}
    \includegraphics[width=0.2\linewidth]{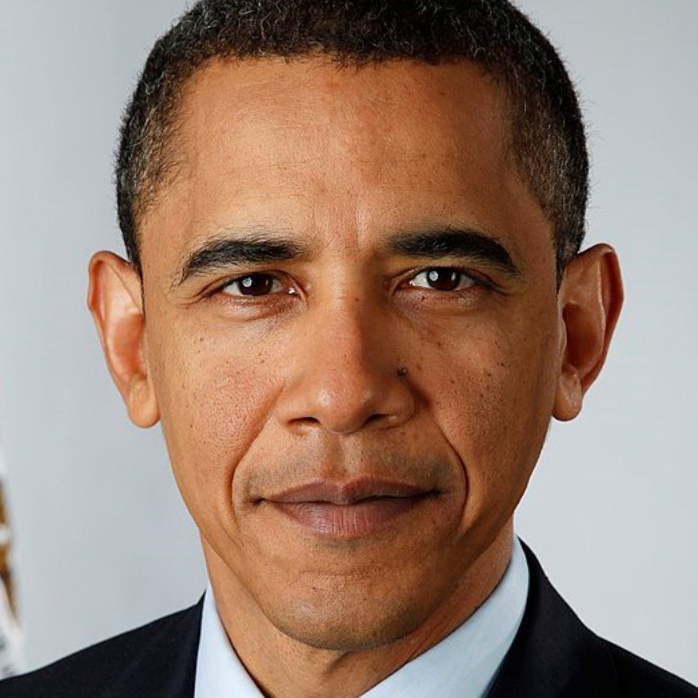} & \includegraphics[width=0.2\linewidth]{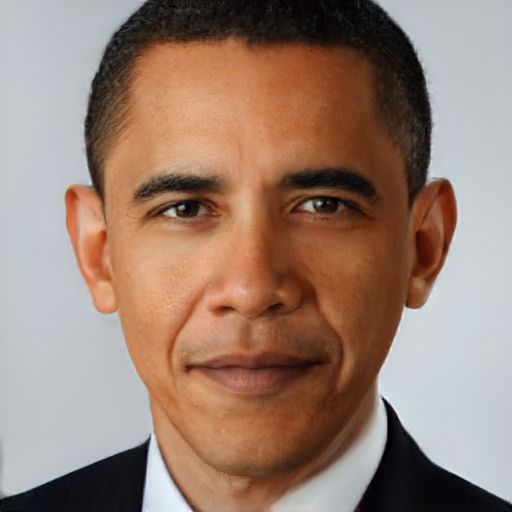} & \includegraphics[width=0.2\linewidth]{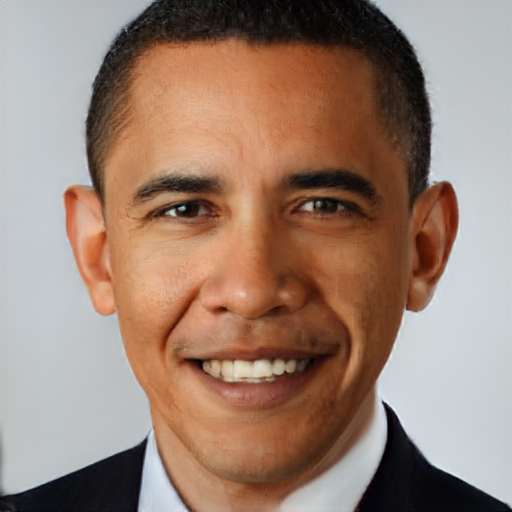} & \includegraphics[width=0.2\linewidth]{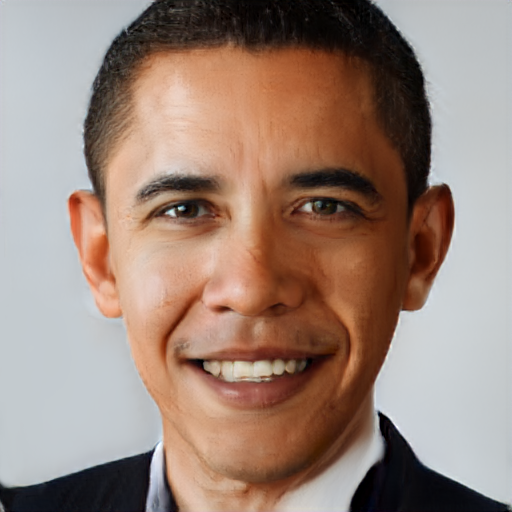} & \includegraphics[width=0.2\linewidth]{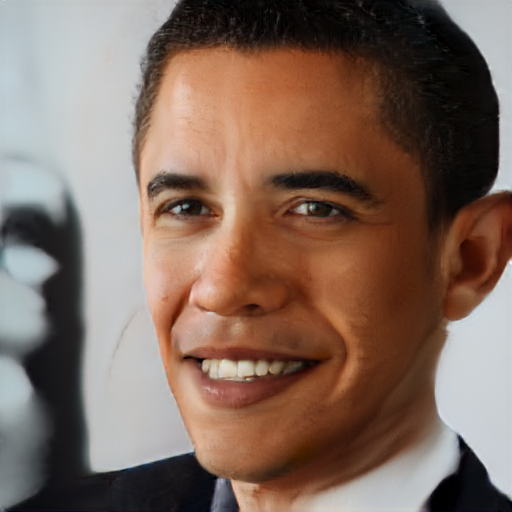} \\ 
    \includegraphics[width=0.2\linewidth]{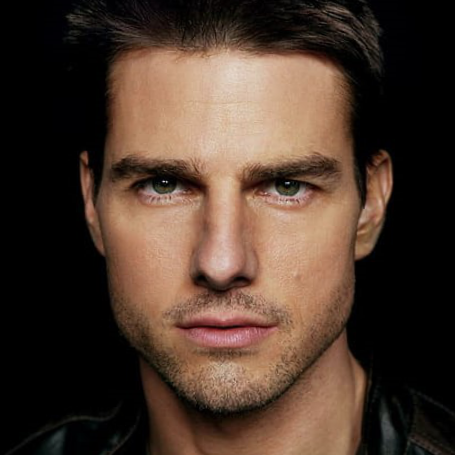} & \includegraphics[width=0.2\linewidth]{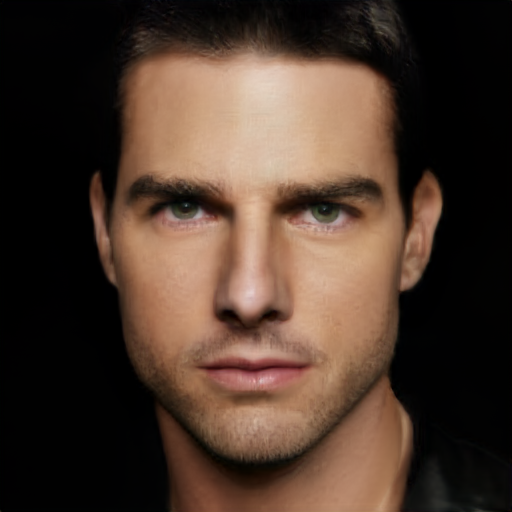} & \includegraphics[width=0.2\linewidth]{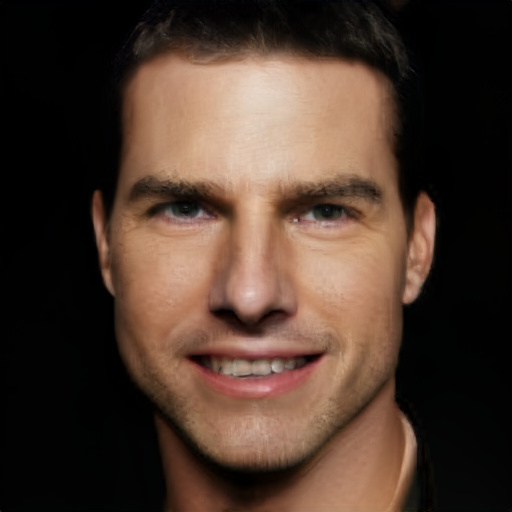} & \includegraphics[width=0.2\linewidth]{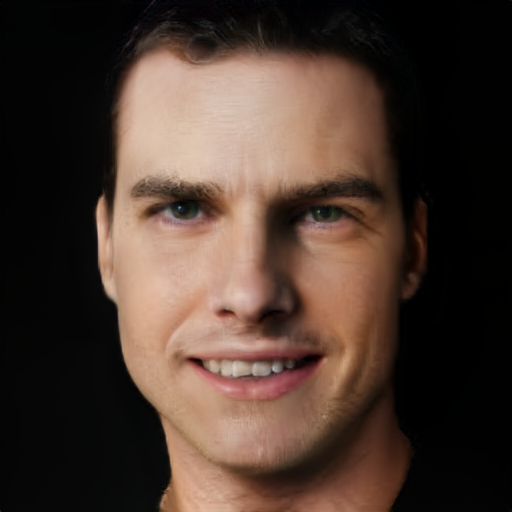} & \includegraphics[width=0.2\linewidth]{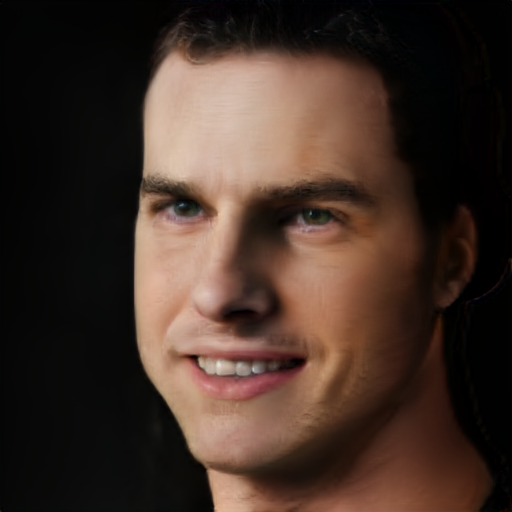} \\ 
    \includegraphics[width=0.2\linewidth]{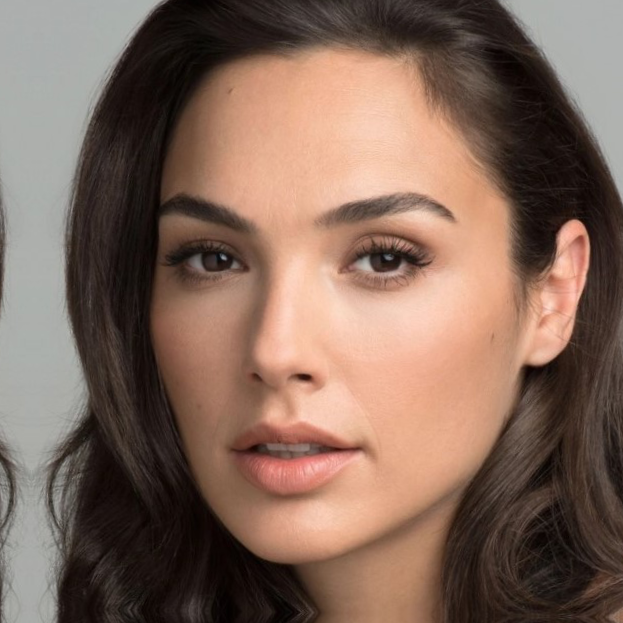} & \includegraphics[width=0.2\linewidth]{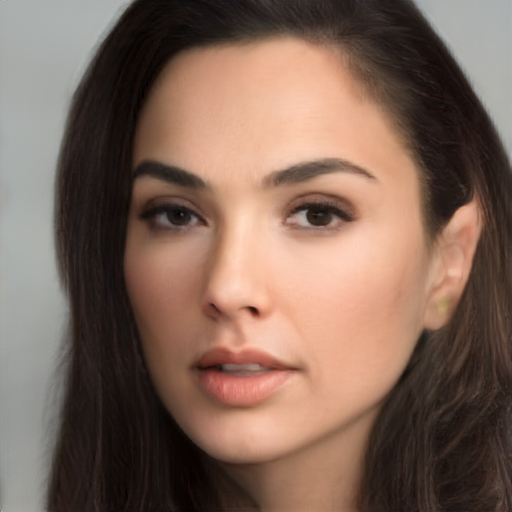} & \includegraphics[width=0.2\linewidth]{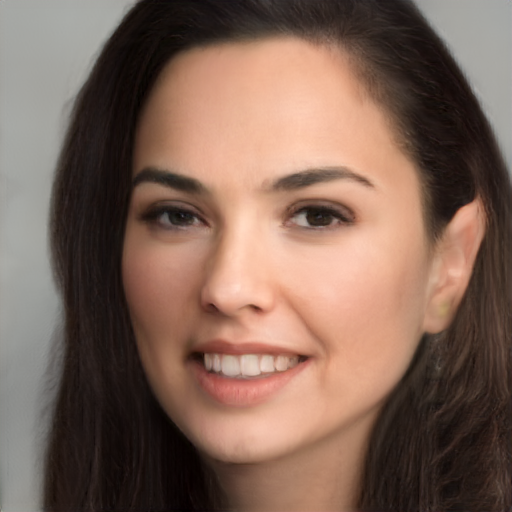} & \includegraphics[width=0.2\linewidth]{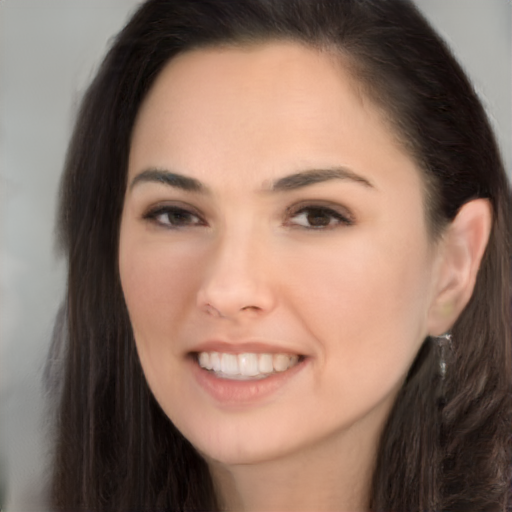} & \includegraphics[width=0.2\linewidth]{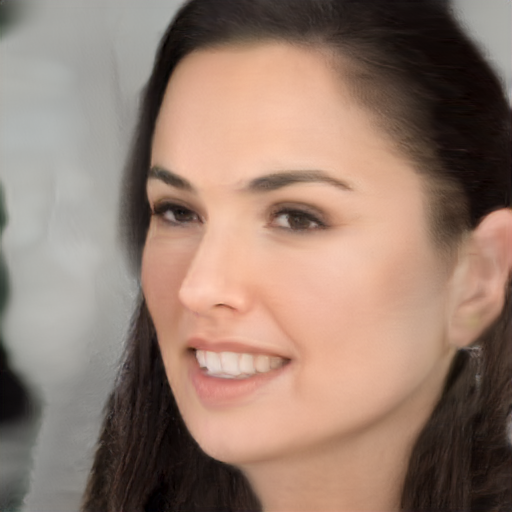}  \\ \rule{0pt}{2ex} 
    Real & Inversion & Change & Change & Change \\
    image & image & expression & illumination & pose
\end{tabular}
    \caption{\textbf{Visualization of real portrait image inversion and editing results.} Given a real image, we project the real image into the latent space and achieve portrait image editing sequentially.}
    \label{fig:inversion}
\end{figure}

\subsection{Real Portrait Image Editing}
As illustrated in Figure~\ref{fig:inversion}, our method allows precise editing of input portrait images in various expression, illumination, and pose while preserving identity.
Inspired by the disentanglement capability of StyleGAN’s latent space~\cite{stylegan}, we project the input portrait image into $\bm{Z}+$ space of the pretrained model and explicitly manipulate the latent codes for real image editing.

To obtain the corresponding latent codes, we use the off-the-shelf inversion technique introduced in StyleGAN~\cite{stylegan}.
Specifically, the latent code $\bm{z}=(\bm{\kappa}, \bm{\beta}, \bm{\gamma}, \bm{\epsilon})$ in semantic control space is initialized with the predicted identity $\bm{\kappa}$, expression $\bm{\beta}$, illumination $\bm{\gamma}$ coefficients from a pretrained face reconstruction network~\cite{deep3dface} and a random vector $\bm{\epsilon}$.
We keep the parameters of the generator fixed and directly optimize the latent codes by measuring the similarity between generated images and real images with the LPIPS~\cite{lpips} loss, a pixel-wise $\mathcal{L}_{2}$ loss and the ID loss~\cite{deng2019arcface}. 

Moreover, in StyleGAN, we can map the latent code $\bm{z}$ on each iteration to $\mathcal{W}$ space, which is an intermediate latent space after the fully connected mapping, and get the latent code $\mathbf{w}$.
To improve the image quality and photo-realism, we introduce the regularization loss $\mathcal{L}_{\textnormal{reg}}$ on $\mathcal{W}$ space as: 
\begin{equation}   \mathcal{L}_{\textnormal{reg}}(\mathbf{w},\mathbf{\overline{w}})=||\mathbf{w}-\mathbf{\overline{w}}||_2, 
\end{equation}
where $\mathbf{\overline{w}}$ is the average latent code in $\mathcal{W}$ space of the pretrained model. 
The overall training objective function can be written as:
\begin{equation}
\begin{aligned}
\mathbf{\bm{z}}^*=\mathop{\arg\min}_{\bm{z}}(\lambda_{1}\mathcal{L}_{\textnormal{LPIPS}}+\lambda_{2}\mathcal{L}_{2}+\lambda_{3}\mathcal{L}_{\textnormal{id}}+\lambda_{4}\mathcal{L}_{\textnormal{reg}}),
\end{aligned}
\end{equation}
where $\lambda_{(\cdot)}$ are the weights for different terms.

To improve the reconstruction quality, we slightly alter the parameters of generator $\mathcal{G}_{\theta}$ to $\mathcal{G}_{\theta}^*$ while keeping the optimized latent codes $\mathbf{z}^*$ fixed. The inversion process takes only 5min for one image on a single NVIDIA Tesla V100 GPU.
After obtaining the latent codes $\mathbf{z}^*$ and the corresponding generator $\mathcal{G}_{\theta}^*$, the expressions and illuminations can be edited by manipulating their corresponding latent codes, and the pose can be changed by controlling the render pose while keeping the identity. Results show that our trained disentangled semantic control space has generalization ability.


\subsection{Out-of-domain Image Editing}
\begin{figure}[t]
    \centering
    \setlength\tabcolsep{0pt}
    \renewcommand{\arraystretch}{0}
    \footnotesize
    \begin{tabular}{cccc}
   \includegraphics[width=0.25\linewidth]{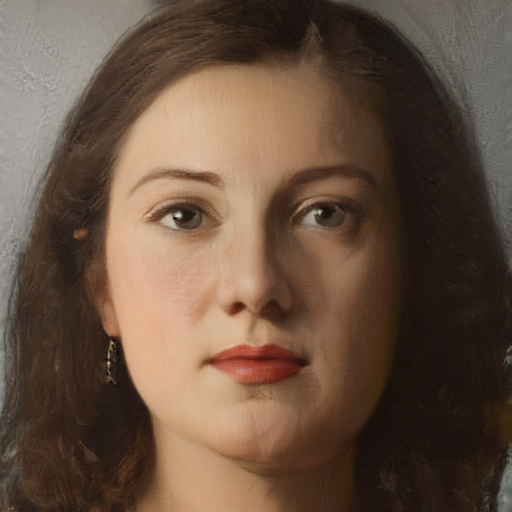} & \includegraphics[width=0.25\linewidth]{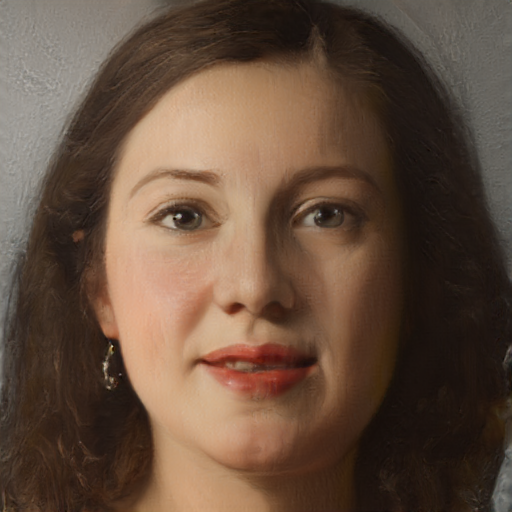} & \includegraphics[width=0.25\linewidth]{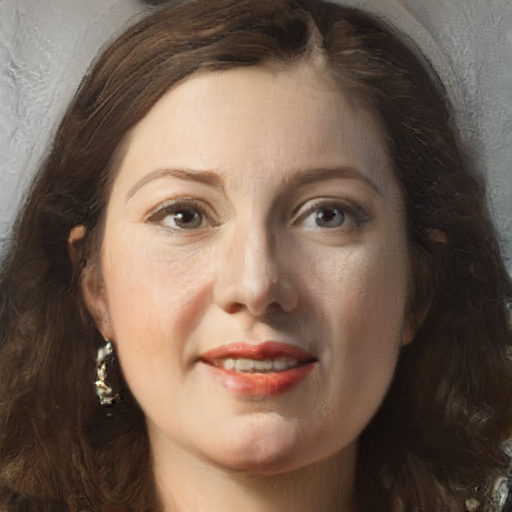} & \includegraphics[width=0.25\linewidth]{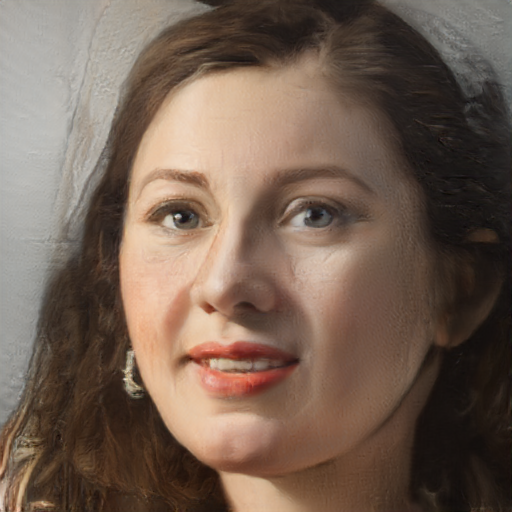}
    \\ \includegraphics[width=0.25\linewidth]{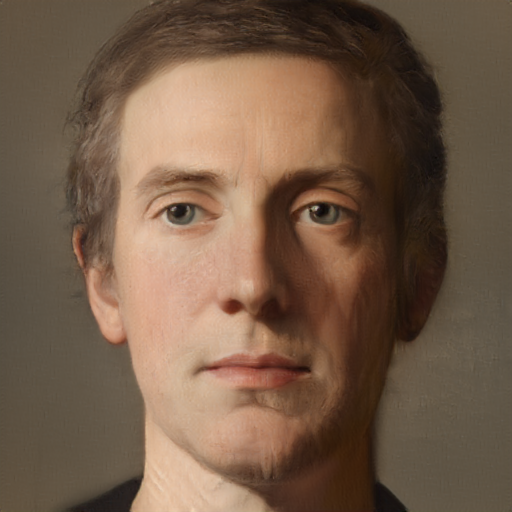} & \includegraphics[width=0.25\linewidth]{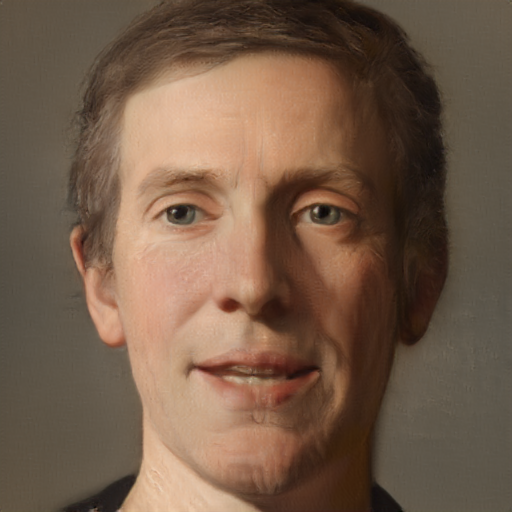} & \includegraphics[width=0.25\linewidth]{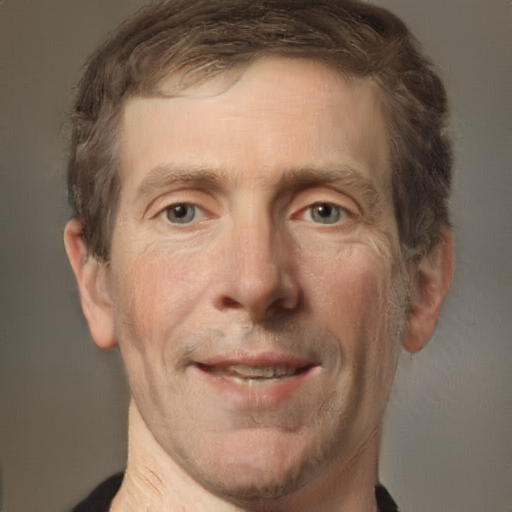} & \includegraphics[width=0.25\linewidth]{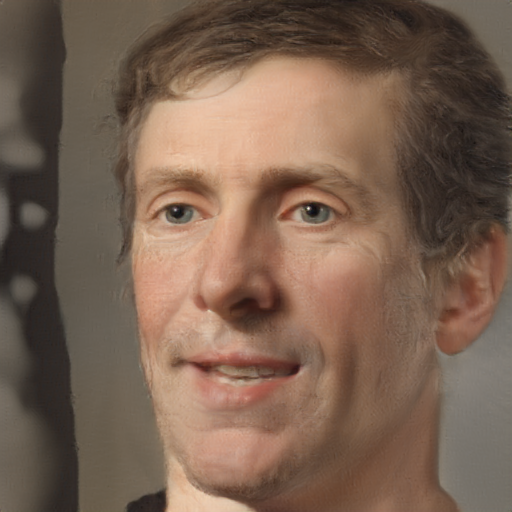}\\ \includegraphics[width=0.25\linewidth]{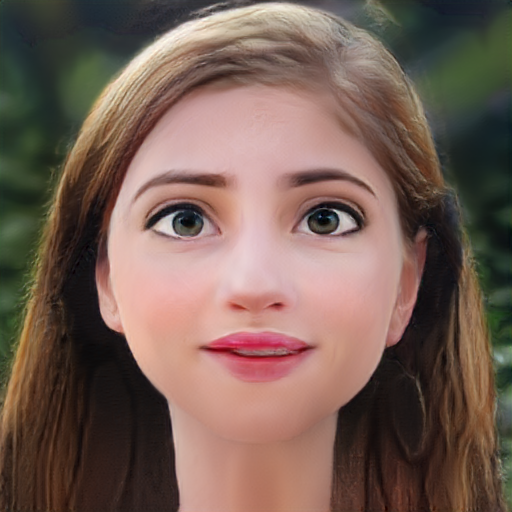} & \includegraphics[width=0.25\linewidth]{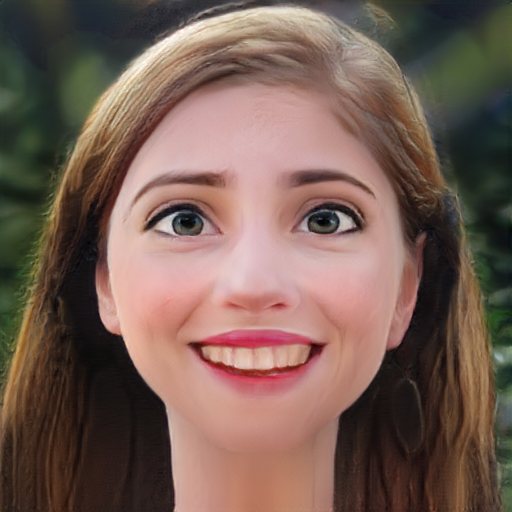} & \includegraphics[width=0.25\linewidth]{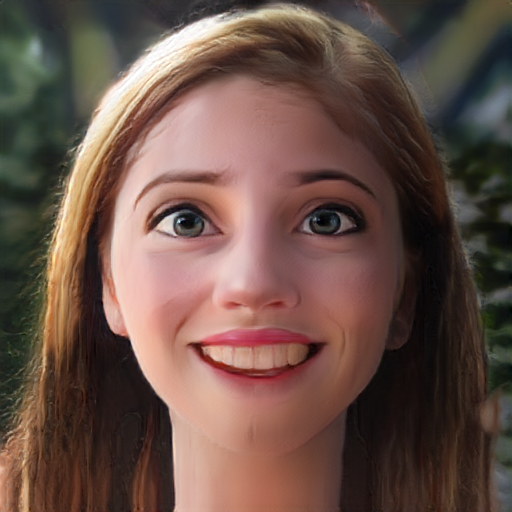} & \includegraphics[width=0.25\linewidth]{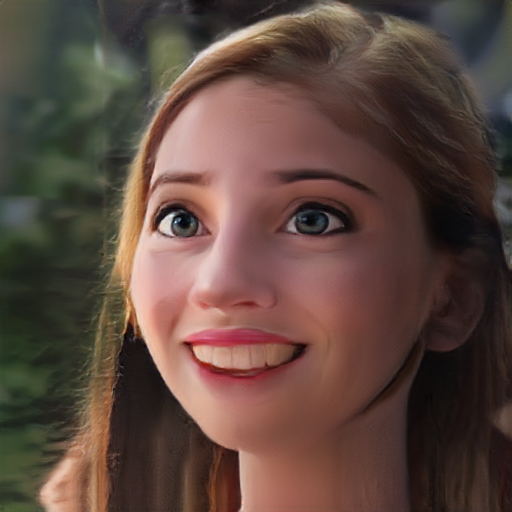} \\\includegraphics[width=0.25\linewidth]{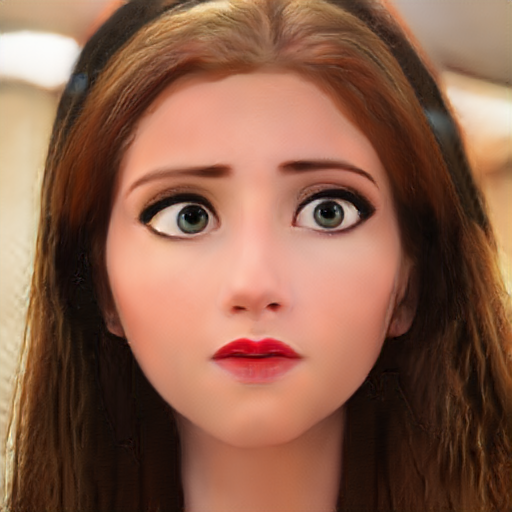} & \includegraphics[width=0.25\linewidth]{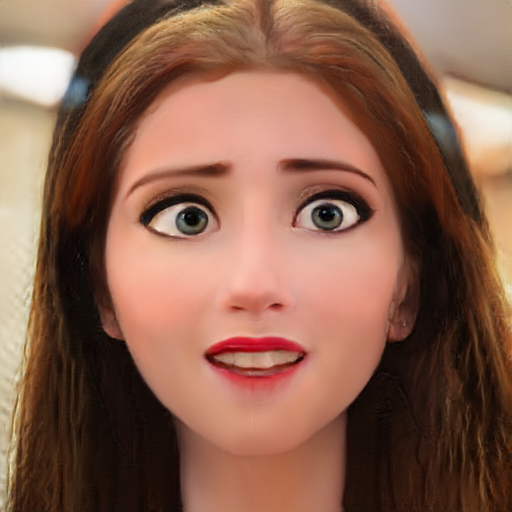} & \includegraphics[width=0.25\linewidth]{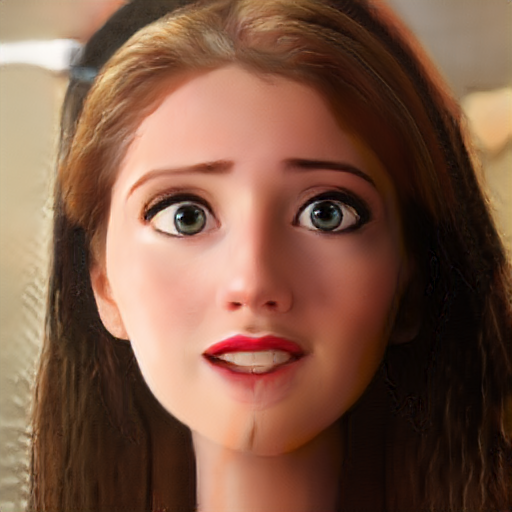} & \includegraphics[width=0.25\linewidth]{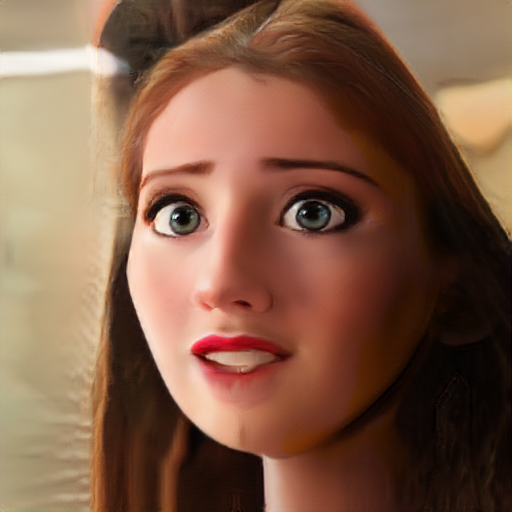} \\
    \rule{0pt}{2ex} 
    Stylized & Change & Change & Change \\
    image & expression & illumination & pose
\end{tabular}
    \caption{\textbf{Visualization of stylized portrait image editing results.} Results show we can implement progressive editing by changing one specific attribute sequentially while keeping other attributes and identities unchanged.}
    \label{fig:cartoon}
\end{figure}

In addition to realistic faces, our method also supports out-of-domain image editing, which is shown in Figure~\ref{fig:cartoon}.
We make use of two stylized portrait datasets: Metface~\cite{karras2020training}, including 1,336 images from the collection of the Metropolitan Museum of Art, and Disney cartoon face, involving 400 online images of Disney cartoon characters collected by \cite{wang2021cross}.
We preprocess the training data using the same preprocess method introduced in Sec.~\ref{sec:preprocess}. 
We freeze the first several layers of the discriminator as in Freeze-D~\cite{mo2020freeze}, and fine-tune the network with adversarial loss. The results show that our method has the out-of-domain generalization ability.

\section{Limitations}

Our method focuses on building a generative and controllable 3D-aware neural radiance field that can be rendered to a high-quality portrait image. However, there might be some limitations. First, our method leverages the 3DMM face mesh as guidance for manipulating portrait images, but 3DMM tends to represent smooth textures and limited human identities. In order to generate diverse results, the identity of the generated image might not be the same as the guided face. Second,
although our method can handle background replacement and can split background and foreground, we cannot disjoint the control for head and body, which is still a challenge in this area. Besides, when generating a talking face, our method tends to represent smiling and laughing expressions due to the data bias of the FFHQ dataset, which can possibly be addressed in the future by combining talking face data for training.

\section{Conclusion}

In this work, we propose a 3D-aware portrait generation network that produces 3D consistent portraits while being controllable according to semantic parameters regarding pose, identity, expression and illumination. 
We can explicitly control the generated neural scene representation using a parametric face model and achieve latent disentanglement.
In order to enforce consistency in non-face areas, \eg, hair and background, when animating expressions, we simultaneously train a semantic radiance field to separate dynamic and static areas. 
We propose a blending strategy in which we form a composite output by blending the dynamic and static radiance fields, with two parts segmented from the jointly learned semantic field. Experimental results show that our method outperforms prior controllable arts. We also investigate multiple applications to demonstrate the generalization ability to real images as well as out-of-domain cartoon faces. The proposed approach opens doors for various extended reality applications that demand 3D consistent avatars with explicit control.  
\ifCLASSOPTIONcaptionsoff
  \newpage
\fi

\bibliographystyle{IEEEtran}
\bibliography{reference}
\end{document}